%% file: main.tex
\documentclass{article}

\PassOptionsToPackage{numbers,compress}{natbib}

\usepackage[preprint]{neurips_2026}
\makeatletter
\renewcommand{\@noticestring}{}
\makeatother

\usepackage[utf8]{inputenc}
\usepackage[T1]{fontenc}

\usepackage{microtype}
\usepackage{float}
\usepackage{graphicx}
\graphicspath{{./figures/}{./}}
\usepackage{subcaption}
\usepackage{booktabs}
\usepackage{longtable}
\usepackage{listings}
\usepackage[table]{xcolor}
\usepackage{makecell}
\usepackage{framed}
\usepackage[framemethod=tikz]{mdframed}
\usepackage{nicefrac}

\usepackage{pifont}
\newcommand{\cmark}{\textcolor{green!70!black}{\ding{51}}}
\newcommand{\xmark}{\textcolor{red!70!black}{\ding{55}}}

\lstdefinestyle{orchestrator}{
  language=Python,
  basicstyle=\ttfamily\small,
  commentstyle=\color{gray!70}\itshape,
  keywordstyle=\color{blue!70!black},
  stringstyle=\color{green!50!black},
  showstringspaces=false,
  columns=fullflexible,
  keepspaces=true,
  frame=single,
  framerule=0.4pt,
  rulecolor=\color{black!15},
  numbers=left,
  numberstyle=\tiny\color{black!50},
  stepnumber=1,
  numbersep=8pt,
  tabsize=4,
  breaklines=true,
  captionpos=b
}

\usepackage{hyperref}
\usepackage{url}

\usepackage[ruled,vlined,linesnumbered]{algorithm2e}

\usepackage{amsmath}
\usepackage{amssymb}
\usepackage{amsfonts}
\usepackage{mathtools}
\usepackage{amsthm}

\usepackage{colortbl}
\usepackage{tcolorbox}
\usepackage{array}
\usepackage{graphicx} 
\usepackage{wrapfig}

\definecolor{gold}{RGB}{212, 175, 55}
\definecolor{silver}{RGB}{192, 192, 192}
\definecolor{bronze}{RGB}{205, 127, 50}
\definecolor{tableheader}{RGB}{26, 31, 53}
\definecolor{tableodd}{RGB}{20, 24, 39}
\definecolor{tableeven}{RGB}{26, 31, 53}
\definecolor{textlight}{RGB}{232, 233, 237}

\usepackage[capitalize,noabbrev]{cleveref}

\usepackage{etoc}

\theoremstyle{plain}

\theoremstyle{definition}

\theoremstyle{remark}

\usepackage[disable,textsize=tiny]{todonotes}

\input{commands}

\newcommand{\method}[1]{Exgentic}
\newcommand{\appworld}[1]{AppWorld}

\title{General Agent Evaluation}

\author{%
  \makebox[\textwidth][c]{\textbf{Elron Bandel$^{1}$\thanks{Correspondence: \texttt{elron.bandel@ibm.com}} \quad Asaf Yehudai$^{1}$ \quad Lilach Eden$^{1}$ \quad Yehoshua Sagron$^{1}$ \quad Yotam Perlitz$^{1}$}} \\[0.3em]
  \makebox[\textwidth][c]{\textbf{Elad Venezian$^{1}$ \quad Natalia Razinkov$^{1}$ \quad Natan Ergas$^{1}$ \quad Shlomit Shachor Ifergan$^{1}$ \quad Segev Shlomov$^{1}$}} \\[0.3em]
  \makebox[\textwidth][c]{\textbf{Michal Jacovi$^{1}$ \quad Leshem Choshen$^{1,2}$ \quad Liat Ein-Dor$^{1}$ \quad Yoav Katz$^{1}$ \quad Michal Shmueli-Scheuer$^{1}$}} \\[0.6em]
  \makebox[\textwidth][c]{\hspace*{-16pt}$^{1}$IBM Research \quad $^{2}$MIT}
}

\begin{document}

\etocsettocdepth.toc{none}

\maketitle

\begin{abstract}
General-purpose agents perform tasks in unfamiliar environments without domain-specific manual customization. Yet no study has systematically measured how agent architecture shapes performance across heterogeneous protocols and diverse unfamiliar environments. This is the first systematic study, comparing tool-calling, \MCP{}, code-generation, and CLI agents on the same benchmarks with the same models. Two gaps blocked such a study: existing harnesses require per-benchmark wiring or fixed protocol classes (web for \browsergym{}, CLI for \harbor{}), and benchmarks themselves expect human-authored prompts, context, and integration glue. To enable this study, we contribute (1) a unifying protocol that bridges existing benchmark and agent protocols; (2) an evaluation harness that surfaces any benchmark to any general-purpose agent and backbone model; and (3) the first \leaderboard{} of agent configurations, a full factorial over \nAgents{} agent architectures $\times$ \nModels{} backbone LLMs (three closed-source, two open-weight) $\times$ \nBenchmarks{} benchmarks spanning software engineering, customer service, deep research, and personal assistance. We find that (i) general agents adapt to every tested domain without per-domain customization; (ii) agent architecture choice swings results by up to 12pp within a single model, yet backbone model choice dominates overall performance; (iii) on 4 of 6 tested benchmarks, top general agents are indistinguishable from the leading heavily-customized domain-specific agents; (iv) open-weight models tested exhibit ``generality sinks'' absent from frontier closed-source models: they consistently collapse on specific agent architectures or benchmarks; (v) a behavioral failure analysis reveals architecture-distinctive error signatures that aggregate scoring cannot discriminate. Code, harness, leaderboard, and traces are at \url{www.exgentic.ai}.
\end{abstract}

\section{Introduction}

\input{sections/01-intro}

\section{\UP{} Methodology}
\label{sec:framework}
\input{sections/02-framework}

\section{Experimental Setup}
\input{sections/03-experimental_setup}

\section{Results}
\input{sections/04-results}

\section{Related Work}
\input{sections/01b-related_work}

\section{Discussion}
\input{sections/05-discussion}

\bibliographystyle{plainnat}
\bibliography{example_paper}

\newpage
\appendix

\etocsettocdepth.toc{section}
\etocsetnexttocdepth{1}
\renewcommand{\contentsname}{Appendix Contents}
\tableofcontents
\newpage

\section{Framework and Adaptation Details}
\label{app:framework_details}
\input{appendices/framework_details}

\section{Agent Adaptation}
\label{appendix:agent-adaptation}
\input{appendices/agent_adaptation}

\section{Benchmark Adaptation}
\label{sec:benchmark-examples}
\input{appendices/benchmark_adaptation}

\section{Detailed Benchmark Agent Interaction Example}
\label{sec:detailed_interaction}
\input{appendices/detailed_interaction}

\section{Detailed Results}
\label{appendix:detailed-results}
\input{appendices/detailed_results}

\section{Statistical Significance}
\label{appendix:statistical}
\input{appendices/significance}

\section{Reproducibility Details}
\label{appendix:reproducibility}
\input{appendices/reproducibility}

\section{Behavioral Error Analysis on Agentic Trajectories}
\label{sec:agentic-errormap}
\input{appendices/agentic_errormap}

\section{Limitations}\label{appendix:limitations}
\input{sections/08-limitations}

\end{document}

%% file: commands.tex
\usepackage{pifont}
\usepackage{xspace}

\newcommand\sa[1]{}
\newcommand\mss[1]{}
\newcommand{\temp}[1]{}

\newcommand{\yk}[1]{}
\newcommand{\eb}[1]{}
\newcommand{\ssi}[1]{}

\newcommand{\sh}[1]{}
\newcommand\led[1]{}
\newcommand\lc[1]{}
\newcommand{\lil}[1]{}
\newcommand{\mj}[1]{}

\newcommand{\Tbench}{$\tau^{2}$-Bench\xspace}
\newcommand{\AppWorld}{AppWorld\xspace}
\newcommand{\BrowseComp}{BrowseComp+\xspace}
\newcommand{\SWEBench}{SWE-Bench Verified\xspace}

\newcommand{\UP}{Unified Protocol\xspace}
\newcommand{\leaderboard}{Open General Agent Leaderboard\xspace}

\newcommand{\pmark}{%
  \textcolor{orange!80!brown}{\ensuremath{\overset{\times}{\checkmark}}}%
}

\newcommand{\solo}{OpenAI Solo\xspace}
\newcommand{\react}{ReAct\xspace}
\newcommand{\short}{ReAct  Short\xspace}
\newcommand{\smol}{Smolagent\xspace}
\newcommand{\cc}{Claude Code\xspace}
\newcommand{\opus}{Claude Opus 4.5\xspace}
\newcommand{\gpt}{GPT 5.2\xspace}
\newcommand{\gemini}{Gemini 3\xspace}
\newcommand{\deepseek}{DeepSeek-V3.2\xspace}
\newcommand{\kimi}{Kimi-K2.5\xspace}

\newcommand{\litellm}{LiteLLM\xspace}
\newcommand{\HAL}{HAL\xspace}
\newcommand{\browsergym}{BrowserGym\xspace}
\newcommand{\harbor}{Harbor\xspace}
\newcommand{\MCP}{MCP\xspace}

\newcommand{\nAgents}{5}
\newcommand{\nModels}{5}        
\newcommand{\nBenchmarks}{6}

%% file: sections/01-intro.tex
Recent AI agents tackle software engineering, navigate web interfaces, and handle multi-application workflows with growing proficiency~\citep{zhang2024autocoderover,deng2023mind2web}.
However, current progress largely relies on domain specialization and manual tuning, whereas heterogeneous real-world settings demand general-purpose agents capable of scalable deployment without such manual customization
~\citep[c.f.,][]{marreed2025enterprisereadycomputerusinggeneralist,bandel2026agentic}. 

Despite their importance, current evaluation practices cannot adequately assess general-purpose agent capabilities. Existing agentic 
benchmarks like \SWEBench{}~\citep{jimenez2023swebench} and \Tbench~\citep{yao2024taubench} provide valuable assessments of domain-specific agents. Yet, they impose two constraints preventing general-agent evaluation: they use bespoke communication protocols~\citep{lacoste2026cubestandardunifyingagent}, and they implicitly assume agents have prior knowledge of benchmark-specific goals and environment semantics~\citep{bandel2026readyforgeneral}.
Recent consolidation efforts like \browsergym{}~\citep{browsergym} and \harbor{}~\citep{harbor} have integrated multiple benchmarks within single domains
by exposing the agent to current goals and environment semantics. While a step forward, these frameworks still enforce a single integration interface (web-based for \browsergym{}, CLI-based for \harbor{}), preventing agents from using their native integration mechanisms and effectively evaluating a diminished version of the agent~\citep{yehudai2025surveyevaluationllmbasedagents}. 

Recent works call for evaluating general agents as a research target~\citep{bandel2026agentic}, but no such evaluation exists today. This paper addresses this gap: no prior study evaluates the same unmodified agent across multiple benchmarks, so the contribution of the agent itself cannot be isolated. Two technical problems blocked such a study: \textit{(i) A framework gap:} no evaluation harness allows any agent to be plugged into a new benchmark unchanged (e.g., \harbor{}~\citep{harbor} requires agents to integrate with its terminal API); even the most mature ones (e.g., Inspect~\citep{inspect_ai_2024}) require manual per-benchmark wiring of tools, sandbox, and solver. \textit{(ii) A benchmark gap:} contemporary benchmarks encode task information, affordances, and interaction, assuming manual integration, and do not expose the means needed for a general-agent interface (Table~\ref{tab:eval_methods_comparison} contrasts existing systems along these axes).

This paper closes the two technical gaps above and presents the first systematic study of diverse general agents across diverse environments (Table~\ref{tab:eval_methods_comparison} positions this work relative to prior evaluation systems). In this work, a \emph{general agent} performs tasks in unfamiliar environments without domain-specific customization; we test each agent unmodified on benchmarks it was not customized for and measure its cross-benchmark success. We release three artifacts: (1) the \textit{\UP{}}, a benchmark-agent mediation protocol that bridges agent interfaces (CLI, tool-calling APIs, \MCP{}) and benchmarks through a canonical task/context/actions representation; (2) \textit{\method{}}, an evaluation harness implementing the \UP{} and surfacing benchmarks through an interface accessible to any general-purpose agent with any backbone model; (3) the first public \leaderboard{}, at a total evaluation cost of \$20K (Table~\ref{tab:leaderboard}).

\input{tables/eval_methods_comparison}

\input{tables/auto_leaderboard_short}

Our analysis of the \leaderboard{} yields five findings. \textbf{(1) General agents adapt without manual tuning:} we run each unmodified across software engineering, customer service, technical support, deep research, and personal assistance benchmarks, with no per-domain engineering, producing non-trivial performance on every domain. \textbf{(2) Model choice dominates aggregate variance, but architecture is conditionally decisive:} model choice explains 27.8\% of variance versus 0.5\% for architecture, a 58$\times$ aggregate gap across closed-source configurations, yet within a single model architecture choice swings results up to 12pp (Fig.~\ref{fig:cost_performance}). \textbf{(3) Generalists are competitive with heavily-customized specialists:} on 4 of 6 tested benchmarks, the best general-agent configuration is statistically indistinguishable from the top published domain-specific score; \BrowseComp{} and \Tbench-Telecom remain specialist-led (Tab.~\ref{tab:best_configs}). \textbf{(4) The two open-weight models tested (\deepseek{}, \kimi{}) exhibit ``generality sinks'' absent from frontier closed-source:} on \Tbench{} the same backbone swings 0.83 to 0.00 across architectures (architecture sink, traceable to first-turn protocol violations), and on \AppWorld{} every architecture collapses (benchmark sink). \textbf{(5) Failure modes discriminate architectures that aggregate scoring cannot:} although architecture explains only 0.5\% of success-rate variance, agents fail in characteristic ways: \cc{} and \solo{} tend to stop too early, while \react{} and \short{} tend to skip evidence gathering (\S\ref{sec:failure_taxonomy}).

Ultimately, advancing general-purpose agents requires a collective effort. We hope the \leaderboard{} serves as a catalyst for approaches that transcend individual tasks and invite the research community to expand this ecosystem by contributing benchmarks that challenge generalization and by designing novel evaluation protocols.

%% file: tables/eval_methods_comparison.tex
\begin{table}[t!]
\centering
\small
\caption{Positioning of this work relative to prior agent-evaluation studies and frameworks along five axes. \cmark~indicates the property holds; \xmark~indicates it does not; \pmark~indicates it holds within a constrained class. \emph{Multi-Protocol Benchmark}: benchmarks are supported in their native form across heterogeneous protocols, without forcing a fixed transport. \emph{Multi-Protocol Agent}: agents are supported in their native form across heterogeneous protocols. \emph{Agent Plug-and-Play}: an agent integrates without per-benchmark adaptation. \emph{Systematic Factorial Study}: The work perform a comparative study that decomposes performance across the full agent-architecture $\times$ backbone-model $\times$ benchmark factorial. \emph{General-Purpose Tasks}: evaluated tasks span domains rather than a single application class. Per-cell justifications are provided in App.~\ref{app:framework_details}.}
\label{tab:eval_methods_comparison}
\resizebox{0.95\textwidth}{!}{%
\begin{tabular}{@{}lccccc@{}}
\toprule
\textbf{System} & \makecell{\textbf{Multi-Protocol}\\\textbf{Benchmark}} & \makecell{\textbf{Multi-Protocol}\\\textbf{Agent}} & \makecell{\textbf{Agent}\\\textbf{Plug-and-Play}} & \makecell{\textbf{General-Purpose}\\\textbf{Tasks}} & \makecell{\textbf{Systematic}\\\textbf{Factorial Study}} \\
\midrule
agent-eval~\citep{bragg2025astabench}       & \cmark & \cmark & \xmark & \xmark & \pmark \\
\HAL{}~\citep{kapoor2025holisticagentleaderboardmissing}      & \cmark & \cmark & \xmark & \cmark & \pmark \\
Inspect~\citep{inspect_ai_2024}             & \cmark & \cmark & \xmark & \cmark & \xmark \\
\browsergym{}~\citep{browsergym}            & \xmark & \xmark & \cmark & \pmark & \pmark \\
\harbor{}~\citep{harbor}                    & \xmark & \xmark & \cmark & \pmark & \xmark \\
AgentBeats                                  & \xmark & \xmark & \cmark & \cmark & \xmark \\
CUBE~\citep{lacoste2026cubestandardunifyingagent} & \xmark & \xmark & \cmark & \xmark & \xmark \\
\midrule
\makecell[l]{\textbf{This work}} & \cmark & \cmark & \cmark & \cmark & \cmark \\
\bottomrule
\end{tabular}%
}
\end{table}

%% file: tables/auto_leaderboard_short.tex
\definecolor{tableheader}{RGB}{248, 249, 250}
\definecolor{rowgray}{RGB}{252, 252, 253}
\begin{table}[t!]
\centering
\renewcommand{\arraystretch}{1.15}
\setlength{\tabcolsep}{5pt}
\small
\caption{The first \leaderboard{}, ranking all 25 (agent architecture $\times$ backbone model) configurations on standardized benchmarks. Each cell reports binary success rate; Avg is the benchmark-weighted average; Cost is average USD per task. Per-cell Wilson 95\% CIs span $\pm 7$--$13$pp; aggregate significance tests in App.~\ref{appendix:statistical}. Zero scores are discussed in \S\ref{sec:generality_sinks}.}
\label{tab:leaderboard}
\begin{tabular}{@{}rllcccccccc@{}}
\toprule
\rowcolor{tableheader}
\textbf{\#} & \textbf{Agent} & \textbf{Model} & \textbf{App} & \textbf{BC+} & \textbf{SWE} & \textbf{T-Air} & \textbf{T-Ret} & \textbf{T-Tel} & \textbf{Avg} & \textbf{Cost} \\
\midrule
  1 & \solo{} & \opus{} & 0.68 & 0.61 & 0.81 & 0.74 & 0.85 & 0.84 & 0.73 & \$8.5 \\
\rowcolor{rowgray}  2 & \cc{} & \opus{} & 0.66 & 0.53 & 0.74 & 0.66 & 0.83 & 0.76 & 0.67 & \$8.0 \\
  3 & \smol{} & \opus{} & 0.70 & 0.61 & 0.65 & 0.72 & 0.78 & 0.58 & 0.66 & \$4.4 \\
\rowcolor{rowgray}  4 & \short{} & \gemini{} & 0.55 & 0.48 & 0.71 & 0.70 & 0.82 & 0.73 & 0.62 & \$0.7 \\
  5 & \short{} & \opus{} & 0.64 & 0.49 & 0.61 & 0.66 & 0.78 & 0.76 & 0.62 & \$3.8 \\
\rowcolor{rowgray}  6 & \react{} & \gemini{} & 0.50 & 0.48 & 0.71 & 0.70 & 0.82 & 0.73 & 0.61 & \$0.8 \\
  7 & \react{} & \opus{} & 0.61 & 0.49 & 0.61 & 0.66 & 0.78 & 0.76 & 0.61 & \$5.8 \\
\rowcolor{rowgray}  8 & \solo{} & \gemini{} & 0.57 & 0.33 & 0.72 & 0.62 & 0.73 & 0.79 & 0.59 & \$2.8 \\
  9 & \cc{} & \gemini{} & 0.36 & 0.51 & 0.67 & 0.70 & 0.71 & 0.71 & 0.56 & \$2.5 \\
\rowcolor{rowgray} 10 & \smol{} & \gemini{} & 0.13 & 0.57 & 0.76 & 0.68 & 0.75 & 0.88 & 0.56 & \$1.8 \\
 11 & \short{} & \gpt{} & 0.22 & 0.46 & 0.57 & 0.54 & 0.73 & 0.53 & 0.46 & \$0.3 \\
\rowcolor{rowgray} 12 & \react{} & \deepseek{} & 0.09 & 0.36 & 0.69 & 0.56 & 0.82 & 0.71 & 0.46 & \$0.4 \\
 13 & \short{} & \deepseek{} & 0.04 & 0.36 & 0.69 & 0.56 & 0.82 & 0.71 & 0.45 & \$0.2 \\
\rowcolor{rowgray} 14 & \short{} & \kimi{} & 0.10 & 0.34 & 0.57 & 0.62 & 0.65 & 0.83 & 0.43 & \$0.4 \\
 15 & \react{} & \kimi{} & 0.09 & 0.34 & 0.57 & 0.62 & 0.65 & 0.83 & 0.43 & \$0.5 \\
\rowcolor{rowgray} 16 & \smol{} & \kimi{} & 0.11 & 0.33 & 0.58 & 0.56 & 0.72 & 0.71 & 0.42 & \$0.7 \\
 17 & \cc{} & \deepseek{} & 0.03 & 0.48 & 0.64 & 0.28 & 0.65 & 0.61 & 0.42 & \$0.2 \\
\rowcolor{rowgray} 18 & \smol{} & \deepseek{} & 0.13 & 0.21 & 0.56 & 0.60 & 0.77 & 0.84 & 0.41 & \$0.2 \\
 19 & \react{} & \gpt{} & 0.00 & 0.46 & 0.57 & 0.54 & 0.73 & 0.53 & 0.41 & \$0.2 \\
\rowcolor{rowgray} 20 & \cc{} & \gpt{} & 0.00 & 0.43 & 0.58 & 0.48 & 0.64 & 0.55 & 0.39 & \$0.4 \\
 21 & \solo{} & \gpt{} & 0.00 & 0.48 & 0.55 & 0.50 & 0.53 & 0.53 & 0.39 & \$0.2 \\
\rowcolor{rowgray} 22 & \smol{} & \gpt{} & 0.07 & 0.26 & 0.53 & 0.60 & 0.68 & 0.71 & 0.38 & \$0.4 \\
 23 & \solo{} & \deepseek{} & 0.06 & 0.30 & 0.74 & 0.20 & 0.19 & 0.18 & 0.32 & \$0.1 \\
\rowcolor{rowgray} 24 & \cc{} & \kimi{} & 0.08 & 0.56 & 0.52 & 0.12 & 0.03 & 0.00 & 0.30 & \$0.6 \\
 25 & \solo{} & \kimi{} & 0.08 & 0.35 & 0.57 & 0.00 & 0.01 & 0.00 & 0.25 & \$0.2 \\
\bottomrule
\end{tabular}
\end{table}

%% file: sections/02-framework.tex
To enable rigorous study of general agents, this work provides a solution for evaluating diverse agents on diverse agentic benchmarks, even when they natively communicate with different protocols. Existing approaches block wide agent evaluation by either restricting to single-protocol harnesses (e.g., \browsergym{}, \harbor{}) or requiring costly per-benchmark integration. We introduce a \UP{} that serves as a faithful mediation layer between agents and benchmarks.

The \UP{} serves as a ``narrow waist'': adding a new agent (or benchmark) only requires adhering to the \UP{}, not to every benchmark (or agent). Thus, it significantly reduces integration complexity, development effort and learning curve.

The \UP{} is not an imposed standard; we derived it by surveying existing agent and benchmark communication patterns and extracting their common structure. By construction, every semantic these protocols express is faithfully representable in the \UP{}, so nothing is lost when translating between them. 

\subsection{Agent Benchmark \UP{}}

The protocol defines instances that are passed between the benchmark and the agent. Each instance has three fields: task, context, and actions. Here we demonstrate them with \Tbench{} as our running example (see other benchmark examples in Appendix~\ref{sec:benchmark-examples}).

\begin{wrapfigure}{r}{0.5\textwidth}
\centering
\includegraphics[width=0.5\textwidth]{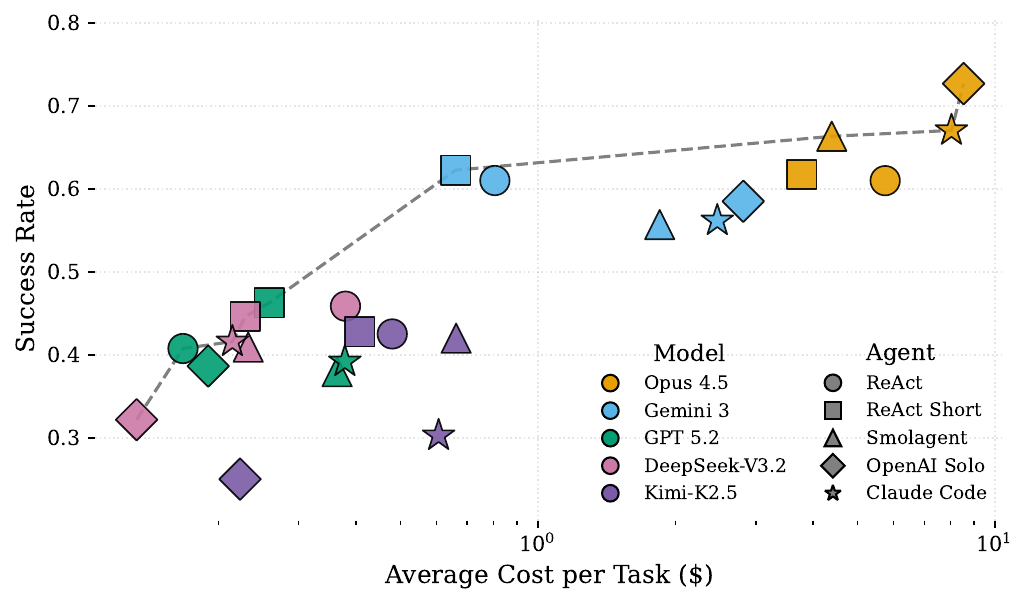}
  \caption{Cost-performance tradeoffs across all 25 agent-model configurations (cost on log scale). The Pareto frontier (dashed line) shows optimal tradeoffs: \deepseek{}- and \gpt{}-backed configurations occupy the cost-efficient end, while \opus{}-backed configurations (notably \solo{}+\opus{}) achieve the highest performance at 1--2 orders of magnitude higher cost. Each point aggregates $n=550$ tasks; model-axis contrasts aggregate $n=2{,}750$ per backbone model, giving tight CIs of $\sim\pm 1.8$pp (App.~\ref{appendix:statistical}).}
  \label{fig:cost_performance}
\end{wrapfigure}
\textbf{Task} (\textit{what the agent should do}): a textual description of the task. In \Tbench{}, it is \textit{``You are a customer service agent that helps the user according to the policy provided below. Try to be helpful and always follow the policy.''}; the first user utterance, such as \textit{``Cancel my flight reservation AH3BDS''}, is passed to the agent separately as the first observation from the environment. \textbf{Context} (\textit{what the agent should know}): additional information provided to the agent to accomplish the task. In \Tbench{}, the context contains the \textit{policy}; the agent can use it in different ways (naively append to the task, store in a dedicated memory or document store for conditional retrieval). \textbf{Actions} (\textit{what the agent can do}): a set of environment actions constituting the complete operations available for performing the task. Each action specifies a typed set of parameters and may return one or more observations of arbitrary types. In \Tbench{} (airline domain), example actions are $\operatorname{cancel\_reservation}(reservation\_id)$ and $\operatorname{search\_direct\_flight}(origin, destination, date)$.

Reviewing existing protocols and agents, we observed that many introduce special handling for two specific types of interactions with the environment: (1) sending a message to a user, and (2) submitting a final answer to the benchmark, signaling that the agent has completed the task. To support these common interaction patterns, the \UP{} allows implementers to optionally designate one action as the message action and one as the final-answer action.

The three-field task/context/actions representation is minimal by design, accommodating any agent protocol decomposable into discrete actions.

\subsection{Running the Evaluation}

We manually adapted six benchmarks and five agents into \method{}, a unified evaluation harness. 
These benchmarks and agents use heterogeneous interaction protocols:
tool-calling APIs (\react{} agent, \Tbench{} benchmark), 
\MCP{} (\solo{}, \cc{} agents), 
Python code generation (\smol{} agent, \AppWorld{} benchmark), 
bash/CLI (\cc{} agent, \SWEBench{} benchmark), 
and conversational messaging (\Tbench{} benchmark).

For each benchmark, we derived a \UP{} interface from a reference agent implementation, preserving the benchmark's intended semantics. 
We then implemented wrapper code that adapts the original benchmark and agent implementations to this \UP{} interface.

The \method{} evaluation harness executes the original agents and benchmarks as black boxes. 
The \method{} orchestrator instantiates the corresponding \UP{} wrappers and executes 150 agent $\times$ model $\times$ benchmark configurations in isolated, reproducible sessions.

Appendix~\ref{app:framework_details} details one complete adaptation example (Mini-SWE agent$\rightarrow$\SWEBench) and describes the orchestrator design.

%% file: sections/03-experimental_setup.tex
We evaluate 5 agent architectures across 5 LLMs (3 frontier closed-source: \gpt{}~\citep{openai2025gpt52}, \opus{}~\citep{anthropic2025opus45}, \gemini{} Pro~\citep{google2025gemini3}; 2 open-weight: \deepseek{}~\citep{deepseekai2025deepseekv32}, \kimi{}~\citep{moonshot2026kimik25}) on 6 benchmarks, with 100 tasks per benchmark (50 for \Tbench{} Airline) and a 100-turn cap per task, following~\citet{perlitz2024efficient}, yielding 150 configurations. By \emph{agent architecture} we mean the full bundle shipped with the agent: scaffold, available tools, memory, schema guards, and any auxiliary components --- everything the agent author commits to except the backbone model. LLM sampling uses each provider's documented defaults (temperature, top-$p$, reasoning mode), chosen to avoid confounding the model-versus-agent comparison with hyperparameter tuning. Full run configurations are released with the code.

\subsection{Benchmarks}\label{sec:benchmarks}
Appendix~\ref{sec:benchmark-examples} provides detailed descriptions of the benchmark adaptations to the \UP{}.

\textbf{\BrowseComp}
~\citep{chen2025browsecompplusfairtransparentevaluation} is a deep research benchmark to assess an agent's ability to handle complex information-search tasks involving iterative search planning and multi-step reasoning. 
While the original benchmark jointly evaluates LLMs and retrieval components, we fix the retriever to isolate agent reasoning and decision-making. We use the authors' provided retriever with either BM25~\citep{robertson1994okapi} or Qwen3 Embedder-based dense retrieval~\citep{zhang2025qwen3embeddingadvancingtext}, and report results using the latter.

\textbf{\Tbench}
evaluates customer-service agents across retail, airline, and telecom domains via LLM-simulated users, measuring both policy-compliant task completion and violation rejection. 
\Tbench{} has a bespoke Python API, where the agent receives a simulated user message and returns either a message reply or calls to one or more predefined tools. We map these into a \textit{message} action and \method{} actions respectively. 

\textbf{\SWEBench}
 A human-validated subset of real-world software engineering tasks from popular Python repositories. Each provides a GitHub issue and repository snapshot; agents produce patches that are evaluated against hidden test suites. Following mini-swe-agent~\citep{miniswe2024,yang2024sweagent} (Tab.~\ref{tab:benchmark-reference-agents}), we expose a single \texttt{bash} action for repository interaction in a sandboxed environment, generating patches via \texttt{git diff} for evaluation. This ensures uniform agent interaction.

\textbf{\AppWorld}
is a benchmark for evaluating user-assistance agents on realistic day-to-day digital tasks. 
In the original protocol, the agent interacts with the environment by writing Python code that is executed in a dedicated interpreter with access to the \AppWorld APIs. 
In our setup, we adopt this native interpreter-based interaction protocol and use the official task definitions and evaluation harness, ensuring consistent API access and evaluation conditions across all agent configurations.

\subsection{Agents}\label{sec:agents}

The five architectures evaluated are: \react{}, \short{} (\react{} with tool shortlisting), \smol{}, \solo{}, and \cc{}. We treat \react{} and \short{} as distinct architectures because tool shortlisting is a substantive component change with measurable impact (\S\ref{sec:leaderboard}), not a hyperparameter sweep.

\textbf{ReAct}
We implement two ReAct-style \citep{yao2023react} agents: a vanilla ReAct baseline using \litellm{}'s~\citep{berriai2024litellm} tool-calling interface (\react{}), and an extended version with tool shortlisting (\short{}). Both are integrated with \method{} by exposing benchmark actions as tool specifications, while the shortlisting variant is designed to handle large action spaces efficiently.

\textbf{\smol{} CodeAgent}
A code-generation agent that produces Python code to invoke tools rather than calling them directly. We integrate \smol{} v1.24.0 \citep{smolagents} with \method{} by exposing benchmark actions as Python functions and adapting its termination behavior to use the benchmark-defined finish action.

\textbf{\solo{} + \MCP{}}
An agent built on the OpenAI Agents SDK~\citep{openai2025agentssdk} v0.7.0 in solo mode with Model Context Protocol~\citep{anthropic2024mcp} integration (\solo for short). The agent operates in solo mode, interacting with environments exclusively through \MCP{} tool calls. We integrate it with \method{} by implementing an adapter that translates benchmark actions into \MCP{} tool specifications.

\textbf{\cc{}}
A feature-rich command-line agent originally designed for software engineering tasks and recently claimed to be generally effective beyond coding\footnote{\smaller \href{https://www.anthropic.com/engineering/building-agents-with-the-claude-agent-sdk}{Building agents with the Claude Agent SDK.}}. We evaluate \cc{} v2.1.7 without modifying its internal logic, integrating it with \method{} via \MCP{}-exposed benchmark actions. The agent runs in a Docker container to ensure isolation and reproducibility.

\subsubsection{Agent Architectural Components}
\label{sec:agent-components}
Agents differ in implementation but share common conceptual components. To gain insight into agents' internal behavior and its impact on performance, we adopt a component-level view covering execution runtime, tool shortlisting, schema guards, communication protocols, memory, and planning. Appendix~\ref{appendix:agent-components} details their presence across agents.

\subsection{Metrics}
To enable consistent comparison across agents and tasks, we adopt the following general metrics. \\
\textbf{Success Rate.} The proportion of runs deemed successful according to the original success definition and evaluation procedure of the benchmark. \\
\textbf{Cost per Task.} The average monetary cost of completing a task, enabling comparison of agent efficiency in addition to performance. In our experiments, costs
are reported using \litellm{}'s pricing data\footnote{\smaller \href{https://github.com/BerriAI/litellm/blob/main/model_prices_and_context_window.json}{Model prices and context window.}}. \\
\textbf{Average Steps.} The mean number of steps taken by an agent to reach task completion. \\
\textbf{Bench-Weighted Mean.} A weighted mean of success across six benchmarks, with the three \Tbench subdomains aggregated and given equal total weight to each of the other benchmarks.
\textbf{Cost-Efficiency.} The ratio of bench-weighted success to mean cost-per-task in dollars (score/\$); component ablations report paired cost deltas at fixed (model, benchmark). \\

\textbf{Step-Based Comparisons.} For step-level analyses (\S\ref{sec:failure_patterns}), zero-step sessions (run-level orchestrator failures) are excluded (7.0\% of runs) and step counts on the remaining runs are capped at 50 to limit outlier influence (10.2\% of remaining runs).

\paragraph{Statistical Analysis.}
Variance decomposition uses $\eta^2 = \mathrm{Var}(\mathbb{E}[Y|X])/\mathrm{Var}(Y)$~\citep{cohen1988statistical} with $Y$ the cell success rate, computed across the 15 closed-source configurations. Pairwise model and architecture rankings use paired $t$-tests on shared (benchmark, task) outcomes; leaderboard top-vs-rest claims use a pooled McNemar test~\citep{mcnemar1947note}. Step-zero rates use two-proportion $z$-tests; cross-benchmark agreement uses Spearman rank correlations~\citep{spearman1904proof}. Multiplicity is controlled via Benjamini--Hochberg~\citep{benjamini1995controlling} and Benjamini--Yekutieli~\citep{benjamini2001control} corrections at $\alpha=0.05$. Full procedures, $p$-values, confidence intervals, and bootstrap settings are in App.~\ref{appendix:statistical}.

%% file: sections/04-results.tex
\subsection{The Open General Agent Leaderboard}\label{sec:leaderboard}

The agent-configuration leaderboard (Table~\ref{tab:leaderboard}) is led by \solo{}+\opus{}, with the top three (all \opus{}-backed) clustering within 6pp; all top-ten configurations are closed-source pairings, with no \gpt{}- or open-weight-backed configuration appearing. The gap from this top cluster to configurations using other backbone models is significant under cross-model paired $t$-tests ($p<0.001$, App.~\ref{appendix:statistical}).

Backbone-model rankings are consistent and significant. Bench-weighted mean success rate of each model averaged across the five architectures: \opus{} 0.66, \gemini{} 0.59, \gpt{} and \deepseek{} tied at 0.41, \kimi{} 0.37 ($p<0.001$, App.~\ref{appendix:statistical}). \opus{}-backed configurations excel broadly; the \gpt{}-backed aggregate is dragged down by configuration failures in tool-rich environments; open-weight backbones trail the frontier on average but vary widely across architectures (\S\ref{sec:generality_sinks}).

Architecture rankings are not significant in aggregate (range $\sim$7pp; $p>0.1$, App.~\ref{appendix:statistical}). But architecture-model pairings matter: \solo{}+\opus{} reaches 0.73 while \solo{}+\gpt{} drops to 0.39, whereas \short{}-based configurations perform more consistently across backbone models.

Two outliers anchor later analysis. \smol{}+\gemini{} achieves the highest single-benchmark cell (0.88 on \Tbench-Telecom). Three \gpt{}-backed configurations score 0.00 on \AppWorld{} without tool shortlisting because \gpt{}'s API is limited to 128 tools while \AppWorld{} exposes $\sim$468 (App.~\ref{app:framework_details}); \opus{}-backed configurations score 0.61--0.70 on the same environment with no shortlisting, indicating the failure is a model-API limitation rather than an integration artifact (analyzed further in \S\ref{sec:generality_sinks}). Within-model architectural spread reaches 12pp for closed-source backbones and 14--18pp for open-weight backbones, with the most dramatic cell-level swing being \kimi{}-backed \Tbench-Telecom 0.83 vs.\ 0.00 across architectures. This motivates both the per-tier sensitivity analysis in \S\ref{sec:sensitivity} and the generality-sinks analysis in \S\ref{sec:generality_sinks}.

\subsection{Generalists Match Heavily-Customized Specialists}\label{sec:specialists}

On 4 of 6 benchmarks, the top general-agent configuration is statistically indistinguishable from the top reported domain-specific agent score~\citep{miniswe2024} under the $n=100$ Wilson half-width~\citep{wilson1927probable} of $\sim$8--10pp (Table~\ref{tab:best_configs}; details in App.~\ref{app:wilson_specialist}, sources in App.~\ref{app:ref_leaderboard}). On \SWEBench{}, \Tbench-Airline, \Tbench-Retail, and \AppWorld{}, the gap is within sampling noise or favors the generalist; on \BrowseComp{} and \Tbench-Telecom, the generalist trails. No single architecture dominates: \solo{} wins on \SWEBench{} and all three \Tbench{} subdomains; \smol{} wins on \AppWorld{}; both tie on \BrowseComp{}. Optimal architecture is benchmark-dependent.

\begin{wraptable}{r}{0.5\textwidth}
\scriptsize
\centering
\caption{Best-performing general-agent configuration per benchmark compared against the top reported domain-specific agent performance on the original benchmark leaderboards (links: App.~\ref{app:ref_leaderboard}). Scores are success rates on 100 randomly sampled instances per benchmark; original leaderboards use full benchmarks.}
\resizebox{0.5\textwidth}{!}{%
\label{tab:best_configs}
\begin{tabular}{@{}llcc@{}}
\toprule
\textbf{Benchmark} & \textbf{Best Configuration} & \textbf{General Agent Score} & \textbf{Domain-Specific Agent Score} \\
\midrule
\SWEBench & \solo{} + \opus{} & 0.81 & 0.79 \\
\BrowseComp & \smol{} + \opus{} & 0.61 & 0.80 \\
\Tbench-Airline & \solo{} + \opus{} & 0.74 & 0.73 \\
\Tbench-Retail & \solo{} + \opus{} & 0.85 & 0.86 \\
\Tbench-Telecom & \smol{} + \gemini{} & 0.88 & 0.98 \\
\AppWorld & \smol{} + \opus{} & 0.70 & 0.73 \\
\bottomrule
\end{tabular}
}
\end{wraptable}

\subsection{Model Choice Dominates Aggregate Variance; Architecture Decides Within Cells}\label{sec:variance}

Within a single backbone model, architecture choice swings results by up to 12 percentage points (\opus{}-backed configurations: best 0.73, worst 0.61). Aggregated across architectures, models, and benchmarks, model effects dwarf architecture effects by an order of magnitude~\citep{kapoor2024aiagents,kapoor2025holisticagentleaderboardmissing}. Variance explained, $\eta^2 = \text{Var}(\mathbb{E}[Y|X]) / \text{Var}(Y)$ where $Y$ is the cell success rate and $X$ is the grouping variable, attributes 27.8\% of total variance to model choice across the 15 closed-source configurations ($p<10^{-10}$, App.~\ref{app:variance-decomposition}); architecture explains only 0.5\%. A complementary additive decomposition over (model, architecture) cell means estimates the interaction effect at 5.4\%, an order of magnitude larger than architecture's main effect, indicating that optimal architecture choice depends on the model. The 58$\times$ aggregate dominance is a closed-source statistic; once the two open-weight models are included, the architecture main effect becomes detectable ($F(4,136)=3.82$, $p=0.006$), reflecting the conditional architectural sensitivity examined in \S\ref{sec:generality_sinks}.

These two views are complementary, not contradictory. The aggregate $\eta^2$ averages out architecture's main effect when pooled across all dimensions; the remaining subsections show where this aggregate view conceals decisive architectural effects in specific cells.

\subsection{Architectural Sensitivity Scales with Model Tier}\label{sec:sensitivity}

Architecture choice matters dramatically more for open-weight than for frontier closed-source backbones~\citep{sclar2023quantifying}. Architectural spread, the difference between best and worst bench-weighted success across the five architectures within a fixed backbone model, separates the tiers: spread is 7--12pp for closed-source backbones and 14--18pp for open-weight backbones ($p<0.001$ best-vs-worst, App.~\ref{app:within-model}). With a strong closed-source model, developers can iterate on agent design without extensive model-specific tuning; with an open-weight model, agent-architecture co-design becomes necessary, examined further in \S\ref{sec:generality_sinks}.

\subsection{Cross-Benchmark Consistency}

Pairwise Spearman rank correlations between benchmark scores across the 15 closed-source configurations are predominantly positive: median $+0.67$, range $[+0.44, +0.81]$ (full matrix in App.~\ref{app:spearman}). \BrowseComp{} shows the lowest pairwise correlations (0.44--0.75), suggesting it captures somewhat distinct capabilities. The positive correlation structure is consistent with the variance decomposition: when a configuration ranks high on one benchmark it tends to rank high on others, leaving model identity as the dominant axis of cross-benchmark consistency (\S\ref{sec:variance}).

\subsection{Schema Guards and Tool Shortlisting Yield Cross-Model Gains}\label{sec:components}

Two architectural components appear to correlate with stronger performance across architectures and models. The three top-performing architectures (\solo{}, \cc{}, and \smol{}) all employ a schema guard (Table~\ref{tab:agent-comparison}): a mechanism that detects when an action with an invalid schema is invoked and allows the agent to self-correct. This is an observational pattern, since these architectures also differ on other components (App.~\ref{appendix:agent-components}); the shortlisting result below is a controlled ablation. Tool shortlisting~\citep{qin2023toolllm}, when added to a simple \react{} architecture in tool-rich environments, improves performance for four of the five backbone models tested: \react{}+\gpt{} gains 5.5 percentage points overall, while \react{}+\opus{} shows a smaller gain but exhibits a \$1.97 cost reduction per task. \deepseek{} is the lone exception (App.~\ref{app:shortlisting_ablation}).

\subsection{Generality Sinks}\label{sec:generality_sinks}

The two open-weight models tested (\deepseek{}, \kimi{}) exhibit two distinct failure modes that frontier closed models do not~\citep{froger2025arescalingagentenvironments}: an architecture sink, where the same backbone model scores 0.83 with one architecture and 0.00 with another, and a benchmark sink, where every architecture collapses on a specific environment. Whether these patterns generalize beyond the two tested checkpoints is an open question.

The architecture sink concentrates on \Tbench{}. Failures cluster on autonomous architectures (\cc{}, \solo{}), whose system prompt instructs the agent to communicate exclusively through tool calls (App.~\ref{appendix:agent-adaptation}). Open-weight models violate this instruction on the first turn, emitting a direct user message instead of a tool call, and the orchestrator terminates the run immediately. The pattern is stark: 94\% of \kimi{}+autonomous \Tbench{} sessions take zero steps, 31\% for \deepseek{}, 1.7\% for closed-source ($p<10^{-15}$, App.~\ref{app:zero-step}). Structured architectures (\react{}, \smol{}) route output directly into tool invocations and bypass the failure, leaving open-weight models competitive: \kimi{}+\react{} reaches 0.83 on \Tbench-Telecom; \deepseek{}+\react{} reaches 0.82 on \Tbench-Retail.

The benchmark sink appears on \AppWorld{}: every open-weight-backed configuration collapses regardless of architecture, with the best \deepseek{}+\smol{} reaching only 0.13 versus 0.70 for \opus{}+\smol{}, reflecting a capability ceiling for tool-rich autonomous interaction over $\sim$468 actions. \gpt{}-backed configurations show a related collapse with a different mechanism: \gpt{}'s 128-tool API limit on \AppWorld{}'s $\sim$468 actions drives three of four \gpt{} configurations to 0.00, with tool shortlisting recovering only to 0.22 (\S\ref{sec:leaderboard}).

\Tbench{} exhibits the architecture sink, \AppWorld{} the benchmark sink; \SWEBench{} and \BrowseComp{} exhibit neither (gap 95\% bootstrap CIs in App.~\ref{app:sinks-decomposition}).

\subsection{Cost-Efficiency Spans 30$\times$ Without a Universal Best}\label{sec:cost_efficiency}

Cost-efficiency (success per dollar of inference cost; per-task cost methodology in App.~\ref{app:repro-data}) varies by approximately 30$\times$ across agent configurations (Fig.~\ref{fig:cost_performance}; full breakdown in App.~\ref{app:cost_efficiency}). \gpt{}- and \deepseek{}-backed configurations share the efficiency frontier (\react{}+\gpt{} and \solo{}+\deepseek{} tie at 2.43 score/\$), while the highest-scoring configuration (\solo{}+\opus{}, 0.73 score) operates at 0.09 score/\$ and the least efficient (\cc{}+\opus{}) at 0.08 score/\$.

\subsection{Failure Patterns Expose Architectural Differences in Resource Use}\label{sec:failure_patterns}

Failed runs are systematically more expensive than successful ones, and architectures differ in how they fail. Comparing successful and failed runs at the task level (across the three closed-source backbones --- open-weight autonomous failures terminate by protocol violation rather than resource exhaustion, \S\ref{sec:generality_sinks} --- after excluding zero-step sessions and capping step counts at 50), bench-weighted averages are positive for every architecture (\cc{} $+39\%$, \solo{} $+20\%$, \smol{} $+26\%$, \react{} $+54\%$, \short{} $+45\%$). The largest overheads appear on interaction-heavy benchmarks (\react{} on \AppWorld{} shows $+111\%$, \cc{} on \BrowseComp{} shows $+70\%$); a few \Tbench{} cells are near zero or negative, indicating early-termination failure modes. The step-count data describes how much each architecture spends on failed runs but does not isolate the underlying mechanism. Detailed per-cell breakdowns are in App.~\ref{app:steps_counts} and App.~\ref{app:per_model_steps}.

\subsection{Behavioral Failure Analysis}\label{sec:failure_taxonomy}

\begin{figure}[t!]
\centering
\includegraphics[width=0.7\textwidth]{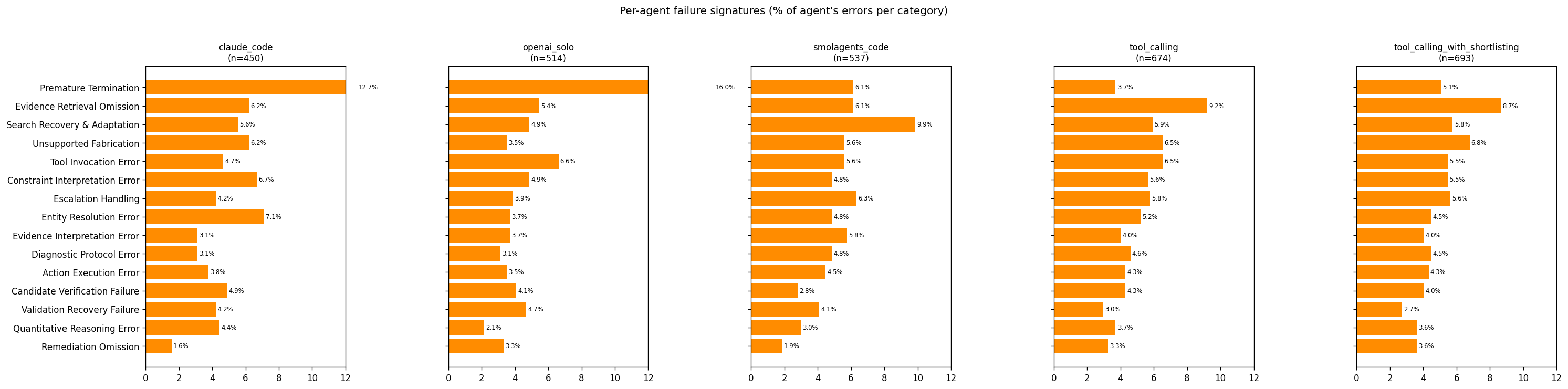}
\caption{Per-architecture failure-category shares (top-15 categories, column-normalized). The architecture axis produces structured spread on this dependent variable, in contrast to its $0.5\%$ contribution to success-rate variance (\S\ref{sec:variance}); per-architecture findings are discussed in \S\ref{sec:failure_taxonomy}.}
\label{fig:agentic-per-agent-body}
\end{figure}

Step counts describe \emph{how} architectures fail; failure modes describe \emph{why}. We adapt \textsc{ErrorMap}~\citep{ashury2026errormap} to the $2{,}868$ failed sessions with full trajectories, producing 27 failure categories (App.~\ref{sec:agentic-errormap}). Although architecture contributes only $0.5\%$ to success-rate variance (\S\ref{sec:variance}), each architecture has a distinctive failure profile: \cc{} and \solo{} over-represent \emph{Premature Termination}; \react{} and \short{} over-represent \emph{Evidence Retrieval Omission}; and \smol{} is dominated by \emph{Search Recovery \& Adaptation} (Fig.~\ref{fig:agentic-per-agent-body}). A chi-squared test of the architecture~$\times$~failure-category contingency table rejects the architecture-uniform null at $\chi^2(196) = 328.5$, $p<10^{-10}$ (Cram\'er's $V = 0.17$); failure modes thus discriminate architectures that aggregate scoring cannot. Per-model, per-architecture, and per-benchmark breakdowns and judge validation are in App.~\ref{sec:agentic-errormap}.

%% file: sections/01b-related_work.tex
\textbf{Domain-Specific Agent Benchmarks.}
The rapid advancement of AI agents has led to a proliferation of benchmarks~\citep{zhou2023webarena,deng2023mind2web,xie2024osworld,liu2023agentbench,mialon2023gaia}, each targeting specific domains such as software engineering~\citep{jimenez2023swebench,yang2024sweagent,merrill2026terminal}, customer service~\citep{yao2024taubench}, and deep scientific research~\citep{bragg2025astabench}.
Each benchmark defines domain-specific protocols and task specifications.

\textbf{Attempts at Consolidation.}
\HAL{}~\citep{kapoor2025holisticagentleaderboardmissing} unifies infrastructure across benchmarks but requires per-benchmark agent adaptation. \browsergym{}~\citep{browsergym} and \harbor{}~\citep{harbor} standardize interaction via fixed protocols (web/CLI) but restrict evaluation to single environment classes. Inspect~\citep{inspect_ai_2024} consolidates the \emph{infrastructure} layer (sandboxing, logging, scoring, and native agent execution, with log-analysis tools such as Scout built on top of it), but its \texttt{Task(solver=[use\_tools([...]), agent()])} pattern still requires the evaluation author to manually choose a tool-set, sandbox, and solver per benchmark. AgentBeats\footnote{\href{https://github.com/agentbeats/agentbeats}{AgentBeats}} requires agents and benchmarks to conform to A2A+\MCP{} subsets for evaluation; CUBE~\citep{lacoste2026cubestandardunifyingagent} unifies benchmarks via an MCP+Gym schema (wrap once, consume universally). Both bind one or both sides to a specific transport; the \UP{} is instead a thin mediation layer that wraps existing agents and benchmarks in their native protocols, without rewriting either side.
The \UP{} operates at the \emph{integration} layer, mediating between agent protocols and benchmark interfaces via a canonical task/context/actions representation. \method{} implements the \UP{} and can be layered on top of infrastructure harnesses like Inspect, enabling protocol-preserving evaluation across heterogeneous benchmarks without per-benchmark agent adaptation. A side-by-side comparison of evaluation systems across five axes is in Table~\ref{tab:eval_methods_comparison}.

%% file: sections/05-discussion.tex
This work takes a step toward the systematic study of general-purpose agents, a target the community has called for but not yet pursued. We develop \method{} and the \UP{} as infrastructure that lets unmodified agents run on unmodified benchmarks, and release the first \leaderboard{} as a public reference for measuring progress.

Across the configurations we evaluate, general agents reach meaningful performance on every domain without any per-domain customization (\S\ref{sec:leaderboard}). On most benchmarks, the best general-agent configuration is indistinguishable from heavily-customized domain specialists (\S\ref{sec:specialists}), showing that general agents are a viable research direction.

The leaderboard analysis also points to clear directions for progress. Models need to be trained to work consistently across architectures (\S\ref{sec:variance}), so that model research and agent-architecture research can advance independently. Today, the same open-weight model can swing between strong performance and total failure depending on which architecture it pairs with, entangling the two research lines (\S\ref{sec:generality_sinks}). New agent architectures and components are worth pursuing: simple components such as schema guards and tool shortlisting moved performance meaningfully across models, and the behavioral failure analysis (\S\ref{sec:failure_taxonomy}) points to additional architecture-specific targets (e.g., resisting early termination), together suggesting substantial headroom (\S\ref{sec:components}). Cost per task also varies widely across configurations, leaving cost-efficiency as a third open axis (\S\ref{sec:cost_efficiency}). The \leaderboard{} provides infrastructure for tracking progress on each of these axes; we invite the community to extend it with new agents, benchmarks, and protocol adaptors as they emerge.

\textbf{Limitations.} Our evaluation is bounded by cost and scope: five LLMs, five agent implementations, and six benchmarks at $\sim$\$20K. Per-benchmark scores carry Wilson CI half-widths of $\pm 7$--$9.5$pp; aggregated comparisons remain highly significant (App.~\ref{appendix:statistical}). Multimodal and continuous-action extensions are discussed in App.~\ref{appendix:limitations}.

%% file: appendices/framework_details.tex
This appendix details how we adapt existing benchmarks and agents to the \UP{}, and the \method{} orchestrator design that runs the full factorial evaluation.

Table~\ref{tab:eval_methods_comparison_extended} reproduces the comparison table from the main paper (Table~\ref{tab:eval_methods_comparison}) with the full caption defining each of the five axes used to position this work against prior evaluation systems.
\input{tables/eval_methods_comparison_appendix}

\begin{figure*}[h!]
    \centering
    \includegraphics[width=1\textwidth]{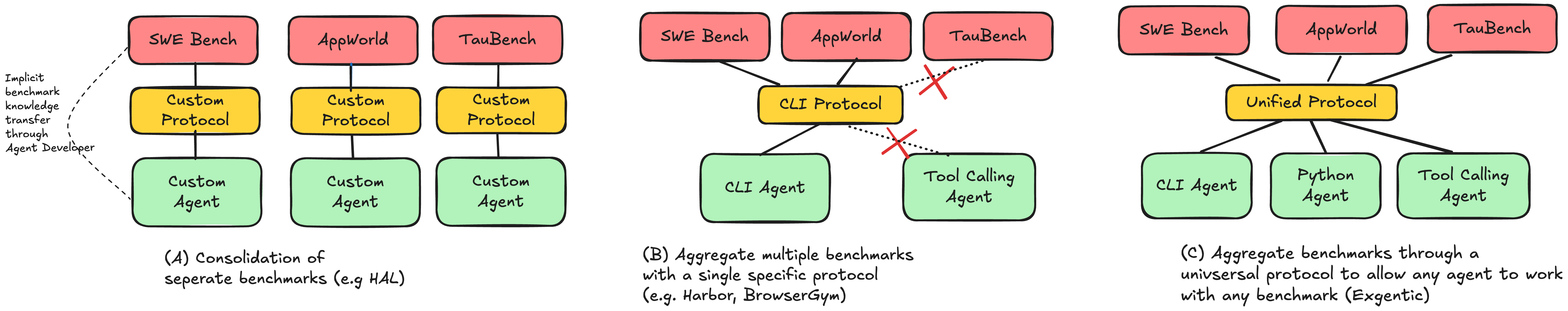}
    \caption{Evolution of Agentic Evaluation. (A) Collection of separate benchmarks, each requiring a custom agent or an agent with specific adaptation per benchmark (\HAL{}). (B) Multiple benchmarks consolidated through a single protocol, such as CLI or Web. (C) Multiple benchmarks consolidated through a common protocol that can be adapted to any agent's protocol (\method{}).}
    \label{fig:agentic_evaluation_evolution}
\end{figure*}

\subsection{Adapting Existing Benchmarks}
\label{app:benchmark-adaptation-method}

Existing agent benchmarks are typically coupled with specific interaction protocols, and often implicitly assume that agents possess prior knowledge of the benchmark's semantics, or that a human will manually perform the integration.

A representative example is \textsc{SWE-Bench Verified}\footnote{\smaller \href{https://huggingface.co/datasets/princeton-nlp/SWE-bench_Verified}{\SWEBench{}}}.
Each task specifies a GitHub repo, a base commit, and a free-text bug description, with the expected output being a patch. The benchmark does not define how agents should access the repo or submit fixes; those details are left to the integrator. For general-purpose agents without human intervention, this interface must be explicit. However, we cannot arbitrarily decide on a setup; instead, we derive the interface from a reference agent implementation.

For \textsc{SWE-Bench Verified}, we examined \textsc{mini-SWE agent}\footnote{\smaller \href{https://github.com/SWE-agent/mini-swe-agent}{\textsc{mini-SWE agent}}} as the reference implementation. There, the agent is placed in a bash environment where the repository has already been cloned. When the agent outputs \texttt{COMPLETE\_TASK\_AND\_SUBMIT\_FINAL\_OUTPUT}, the system automatically generates a patch and submits it for evaluation. This design fully specifies how the agent interacts with the benchmark, what actions it may take, and how it submits solutions, implicitly indicating that repository cloning and patch creation are \textit{not} evaluation targets.

Accordingly, in the \method{} protocol for \textsc{SWE-Bench Verified}, we introduce two explicit actions: one for executing bash commands and another for submitting a patch constructed from the agent's code modifications.

To define the protocol's task and context fields, we review both the benchmark tasks and the reference implementation prompts. Many benchmark tasks include irrelevant implementation details, while key instructions appear only in the reference agent's internal prompts. For instance, in \Tbench, the reference prompt states: \textit{``You are a customer service agent that helps the user according to the \texttt{<policy>} below.''} Such essential information belongs in the benchmark task itself and is included in the \method{} task definition. In contrast, instructions like \textit{``Each turn you may either message the user or make a tool call, but not both''} are excluded because they assume a particular tool-calling protocol.

In summary, we decouple each benchmark from its original protocol by making all agent-visible assumptions explicit. First, we inspect the reference agent to see how it interacts with the environment and what actions and observations it uses. Then we build task descriptions that include only the information needed for the agent to solve the task, omitting implementation-specific details and redundant signals. This yields tasks that preserve the benchmark's intended semantics while remaining independent of any particular agent architecture or communication protocol, making them suitable for evaluating any general agent implementation.

\subsubsection{Per-Benchmark Adaptations}
\label{app:per-benchmark-adaptation}

Table~\ref{tab:benchmark-adaptation-summary} summarizes the six adaptations along the three protocol fields plus the reference implementation we mirrored. The full task / context / actions text for each benchmark is in Appendix~\ref{sec:benchmark-examples}; the released adaptor code lives at \texttt{src/exgentic/benchmarks/<benchmark>/} in the released repository.

\begin{table}[H]
\centering
\scriptsize
\setlength{\tabcolsep}{4pt}
\renewcommand{\arraystretch}{1.15}
\begin{tabular}{@{}l p{2.4cm} p{2.2cm} p{4.6cm}@{}}
\toprule
\textbf{Benchmark} & \textbf{Reference impl.} & \textbf{Actions exposed} & \textbf{Task / context fields} \\ \midrule
\SWEBench{}        & mini-SWE agent           & \texttt{bash}, \texttt{finish}                       & Task: bash-protocol prompt + issue body. Context: empty. \\
\AppWorld{}        & native interpreter       & one per AppWorld API ($\sim$468); \texttt{finish} & Task: supervisor instruction. Context: \texttt{policy}, \texttt{supervisor}, \texttt{datetime}. \\
\BrowseComp{}      & authors' retriever       & \texttt{search}, \texttt{get\_document}, \texttt{submit} & Task: scoring-rubric prompt + question. Context: empty. \\
$\tau^2$-Airline   & $\tau^2$ reference ReAct & \texttt{message} + benchmark tools & Task: ``customer service agent'' line. Context: \texttt{policy} (airline). \\
$\tau^2$-Retail    & $\tau^2$ reference ReAct & \texttt{message} + benchmark tools                  & As above; \texttt{policy} (retail). \\
$\tau^2$-Telecom   & $\tau^2$ reference ReAct & \texttt{message} + benchmark tools                  & As above; \texttt{policy} (telecom). \\
\bottomrule
\end{tabular}
\caption{Per-benchmark adaptation summary. Released adaptor code:
\texttt{src/exgentic/benchmarks/\{appworld, browsecompplus, swebench, tau2\}/}.}
\label{tab:benchmark-adaptation-summary}
\end{table}

Table~\ref{tab:benchmark-reference-agents} pins the canonical source for each reference implementation we mirrored.

\begin{table}[H]
\centering
\scriptsize
\setlength{\tabcolsep}{4pt}
\renewcommand{\arraystretch}{1.2}
\begin{tabular}{@{}l p{3.4cm} p{6.3cm}@{}}
\toprule
\textbf{Benchmark} & \textbf{Reference agent} & \textbf{Source} \\ \midrule
\SWEBench{}     & mini-SWE agent                  & \href{https://github.com/SWE-agent/mini-swe-agent}{github.com/SWE-agent/mini-swe-agent} \\
\AppWorld{}     & \texttt{minimal\_agent.ipynb} (ReAct) & \href{https://github.com/StonyBrookNLP/appworld/blob/main/notebooks/minimal_agent.ipynb}{StonyBrookNLP/appworld/notebooks/minimal\_agent.ipynb} \\
\BrowseComp{}   & OpenAI search-agent baseline    & \href{https://github.com/texttron/BrowseComp-Plus/blob/main/search_agent/openai_client.py}{texttron/BrowseComp-Plus/search\_agent/openai\_client.py} \\
\Tbench{}       & \texttt{LLMAgent} (text default) & \href{https://github.com/sierra-research/tau2-bench/blob/main/src/tau2/agent/llm_agent.py}{sierra-research/tau2-bench/src/tau2/agent/llm\_agent.py} \\
\bottomrule
\end{tabular}
\caption{Reference agent implementations mirrored when defining each benchmark's \UP{} adaptor. Each row links to the canonical implementation in the benchmark's own repository.}
\label{tab:benchmark-reference-agents}
\end{table}

\paragraph{\AppWorld{}.} The native interface is a dedicated Python interpreter exposing the AppWorld APIs as importable functions; the reference agent writes Python code that the interpreter executes. We preserve this interface by treating each AppWorld API ($\sim$468 across the nine apps) as a distinct \UP{} action, generated programmatically from \texttt{world.task.api\_docs.function\_calling()}. The benchmark's native completion call (\texttt{supervisor.complete\_task}) becomes the protocol's \texttt{finish} action. The task field carries the supervisor's natural-language instruction (\texttt{"Task from supervisor:\textbackslash n\{instruction\}"}); environment-level information that the reference agent assumes implicitly --- which apps are available, how supervisor credentials are obtained, time and pagination conventions --- is moved into the context field as a \texttt{policy} entry, alongside the \texttt{supervisor} record and the simulated \texttt{datetime}. Code-generation agents are not privileged: any agent that can call the exposed actions (Python functions, tool-calls, \MCP{}) can solve AppWorld tasks. Because \gpt{}'s tool-use API enforces a 128-tool ceiling per request, \gpt{}-backed configurations without tool shortlisting cannot expose the full 468 actions and score 0.00 on \AppWorld{}; this is a model-API limitation rather than an integration artifact (\S\ref{sec:leaderboard}, \S\ref{sec:generality_sinks}).

\paragraph{\BrowseComp.} \BrowseComp{} provides queries, a fixed corpus, and an authors' retriever; the reference harness expects an agent that can issue searches and (optionally) fetch full documents. We expose three actions: \texttt{search} (single \texttt{query} string, returns the top-$k$ snippets each trimmed to a documented token budget), an optional \texttt{get\_document} (full text by document id), and \texttt{submit} (finish action, with structured fields \texttt{exact\_answer}, \texttt{explanation}, \texttt{confidence}). The task field embeds the question along with the benchmark's scoring rubric and a hard rule that finishing requires \texttt{submit} (since the original benchmark's user interaction does not exist in our setup). Context is empty: the corpus is server-side, accessed only through \texttt{search}/\texttt{get\_document}.

\paragraph{$\tau^2$-Bench (Airline / Retail / Telecom).} The native interface is the bespoke Python API in $\tau^2$-Bench: the agent receives messages from a simulated user and either replies with a message or invokes one of the benchmark-registered tools (e.g., \texttt{cancel\_reservation}, \texttt{search\_direct\_flight}). We preserve this by mapping the user-reply path to the protocol's \texttt{message} action, and translating each $\tau^2$ tool (via its OpenAI schema) into a distinct \UP{} action. The task field is the single line \textit{``You are a customer service agent that helps the user according to the \texttt{<policy>} provided below. Try to be helpful and always follow the policy.''} The domain-specific policy (airline / retail / telecom rules) is exposed through the context field. Tool-calling-protocol-specific instructions present in the reference agent (e.g., ``each turn you may either message the user or make a tool call, but not both'') are deliberately omitted, because they constrain protocol shape rather than task semantics.

\subsection{Adapting Existing Agents}
\label{app:agent-adaptation-method}

Existing agents interface with existing environments through specific protocols such as \MCP{}, Python functions, or tool calls. They also receive the task description through some command line or programmatic API.

Adapting agents to the \UP{} involves deciding how to map the task, context, and actions of the protocol to the agents' specific API. It is important to note that the agent adaptor is benchmark agnostic.

The textual task descriptions are typically concatenated with the context fields into textual instructions passed to the model. While not implemented today, the context may be used in different ways. For example, an \MCP{}-based agent may opt to store the context in \MCP{} resources rather than add them to the instructions.

Action adaptation is straightforward and largely reusable across agents using similar APIs, with each action mapped to a single Python function, OpenAI tool, or \MCP{} tool.

More subtle adaptation is dealing with special actions. One special action type is interacting with a user. Some agents, like tool-calling agents, natively interact with users using dedicated \textit{assistant}- and \textit{user}- messages rather than through tool API. To preserve the principle of presenting the benchmark to the agent in the most natural way, the tool-calling agent adaptor converts \textit{user} and \textit{assistant} messages to the corresponding \textit{message} action.

\subsection{\method{} Orchestrator Design}
\label{app:framework-orchestrator}

General-purpose agents must operate across diverse environments, and hence viable evaluation frameworks must scale across many benchmarks and agents. The \method{} framework enables running any currently supported agent on any supported benchmark task, with any LLM, using only a few lines of standard Python code or a dedicated GUI.

The framework was built for use at scale and supports parallelism and caching. Every run is executed in an isolated environment and is reproducible. Benchmark results, interaction trajectories, and cost reports are created in a standardized format for all benchmarks and agents.

\begin{figure*}[h!]
    \centering
    \includegraphics[width=1.0\textwidth]{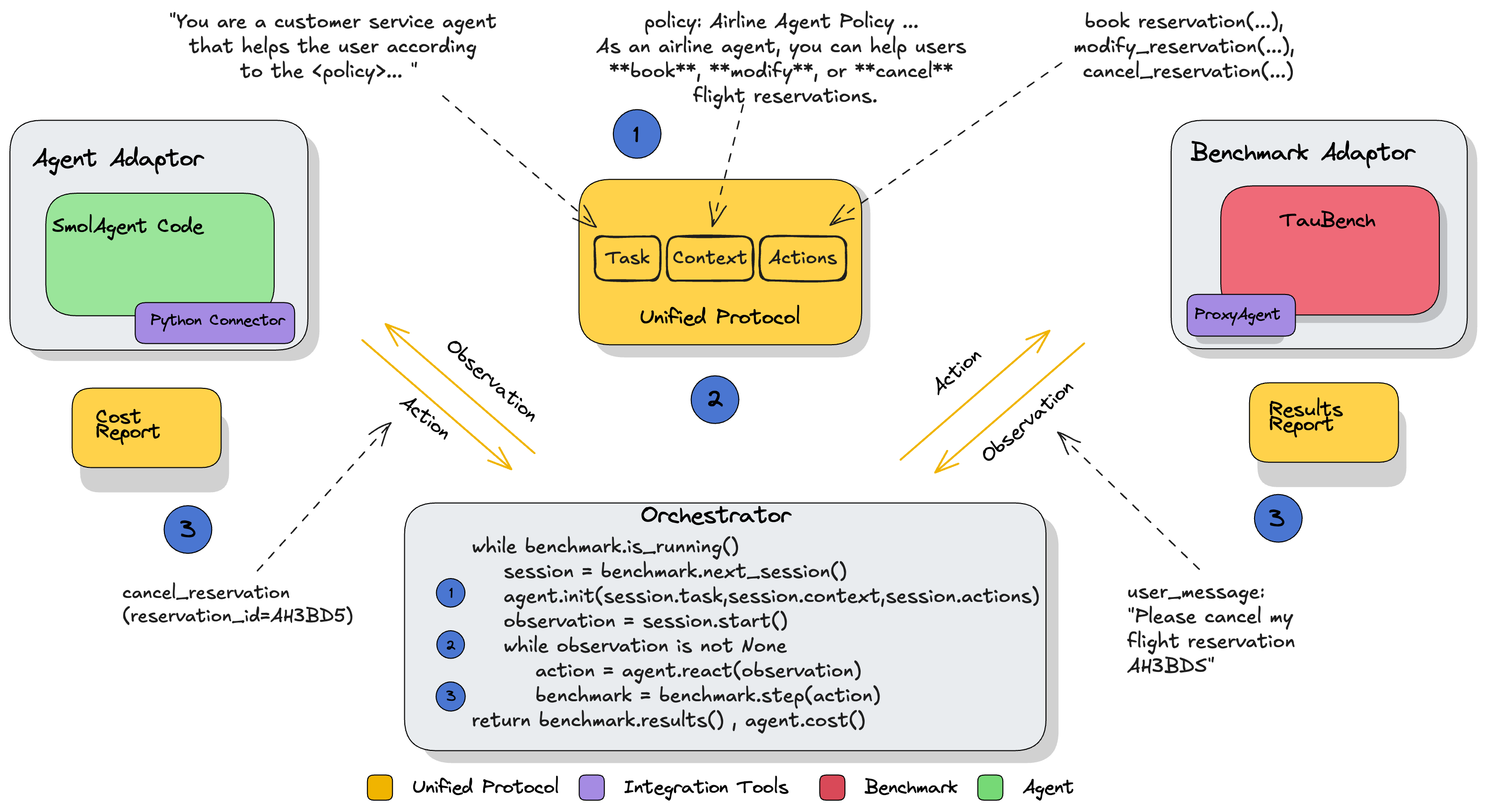}
    \caption{\method{} architecture. \method{} defines a unified protocol between agents and benchmarks. The \method{} Orchestrator connects the agent and the benchmark, first passing the task definition and then mediating the observations and actions that are passed between the benchmark and the agent. \method{} provides adaptors that convert the \UP{} into the specific protocols required by the agents and benchmarks. Finally, the benchmark provides the quality result metrics while the agent provides the agent runtime cost.}
    \label{fig:framework-architecture}
\end{figure*}

The main orchestration loop is illustrated in Figure~\ref{fig:framework-architecture}. Each benchmark generates a set of sessions, where each session corresponds to a single benchmark task the agent must complete (e.g., resolving a GitHub issue, or fulfilling a specific user request). For each session, the orchestrator initializes the agent with the task description, contextual information, and the set of available actions.

Following initialization, the agent receives the first observation from the session environment and responds by selecting one of the permissible actions. This action is executed by the environment, which returns a new observation. The loop continues until either the session concludes or the agent terminates by emitting no further actions. We also terminate if the number of actions/observations exceeds some threshold to avoid deadlocks or excessive costs.

\subsubsection{Solving the Integration Problem}

Adapting existing agents and benchmarks to the \UP{} and integrating them with the \method{} orchestrator is conceptually straightforward but practically challenging. These components are developed independently by third-party authors who are unaware of the \UP{}, the orchestrator's execution model, or each other's design assumptions.

Presumably, one possible solution is to make intrusive modifications to the benchmark and agent code bases to make them use the \UP{}. However, such changes may be extremely costly to implement, difficult to maintain, or even impossible when the agent or benchmark is closed-source.

Instead, we use external adaptor code that handles synchronization and protocol translation. On the agent side, adaptors expose the \UP{} actions in whatever form the agent expects (Python functions, \MCP{} server tools, or OpenAI tools). On the benchmark side, they translate each benchmark's task definition and agent interface into the \UP{}. Since many adaptations repeat across agents and benchmarks, we provide base adaptors that simplify building specific ones.

We allow agents and benchmarks to run natively and independently in separate processes, while all communication between them is mediated by the orchestrator and the corresponding adaptor components. This design ensures that neither the benchmarks nor the agents are affected by the fact that they are running inside the \method{} framework, preserving their original behavior.

For a complete walkthrough of an interaction between a code-generation agent (\smol{}) and \Tbench, see Appendix~\ref{sec:detailed_interaction}.

%% file: tables/eval_methods_comparison_appendix.tex
\begin{table}[H]
\centering
\small
\caption{Positioning of this work relative to prior agent-evaluation studies and frameworks along five axes. \cmark~indicates the property holds; \xmark~indicates it does not; \pmark~indicates it holds within a constrained class. \emph{Multi-Protocol Benchmark}: benchmarks are supported in their native form across heterogeneous protocols, without forcing a fixed transport. \emph{Multi-Protocol Agent}: agents are supported in their native form across heterogeneous protocols. \emph{Agent Plug-and-Play}: an agent integrates without per-benchmark adaptation. \emph{General-Purpose Tasks}: evaluated tasks span domains rather than a single application class. \emph{Systematic Factorial Study}: the work performs a comparative study that decomposes performance across the full agent-architecture $\times$ backbone-model $\times$ benchmark factorial. This is the extended version of Table~\ref{tab:eval_methods_comparison} in the main paper.}
\label{tab:eval_methods_comparison_extended}
\resizebox{\textwidth}{!}{%
\begin{tabular}{@{}lccccc@{}}
\toprule
\textbf{System} & \makecell{\textbf{Multi-Protocol}\\\textbf{Benchmark}} & \makecell{\textbf{Multi-Protocol}\\\textbf{Agent}} & \makecell{\textbf{Agent}\\\textbf{Plug-and-Play}} & \makecell{\textbf{General-Purpose}\\\textbf{Tasks}} & \makecell{\textbf{Systematic}\\\textbf{Factorial Study}} \\
\midrule
agent-eval~\citep{bragg2025astabench}       & \cmark~(Inspect harness)       & \cmark~(InspectAI solvers)     & \xmark~(solver implementation)       & \xmark~(scientific research only) & \pmark~(no model axis; sci only) \\
\HAL{}~\citep{kapoor2025holisticagentleaderboardmissing}      & \cmark                         & \cmark                         & \xmark~(per-benchmark adaptation)    & \cmark                            & \pmark~(architecture per-benchmark wired) \\
Inspect~\citep{inspect_ai_2024}             & \cmark                         & \cmark                         & \xmark~(per-task solver setup)       & \cmark                            & \xmark~(framework only) \\
\browsergym{}~\citep{browsergym}            & \xmark~(web only)              & \xmark~(custom API)   & \cmark                               & \pmark~(web-mediated)             & \pmark~(single architecture across LLMs) \\
\harbor{}~\citep{harbor}                    & \xmark~(CLI only)              & \xmark~(CLI only)              & \cmark                               & \pmark~(CLI-mediated)             & \xmark~(framework only) \\
AgentBeats                                  & \xmark~(A2A+\MCP{})            & \xmark~(A2A+\MCP{})            & \cmark                               & \cmark                            & \xmark~(framework only) \\
CUBE~\citep{lacoste2026cubestandardunifyingagent} & \xmark~(\MCP{}+Gym) & \xmark~(\MCP{}+Gym) & \cmark                               & \xmark                            & \xmark~(position paper; no empirical eval) \\
\midrule
\makecell[l]{\textbf{This work}} & \cmark & \cmark & \cmark & \cmark & \cmark \\
\bottomrule
\end{tabular}%
}
\end{table}

%% file: appendices/agent_adaptation.tex
This appendix documents the five agent integrations behind the
\leaderboard{}: pinned versions, the system prompt / first-turn
instructions each agent receives, and the integration approach. The
released agent adaptor code lives at
\texttt{src/exgentic/agents/<agent>/} in the released repository.

\subsection{Pinned Agent Versions}
\label{appendix:agent-versions}

Table~\ref{tab:agent-versions} pins the exact version string used for
each agent in every cell of the leaderboard; the same strings appear in
the agent column of the released \texttt{results.csv}.

\begin{table}[H]
\centering
\small
\begin{tabular}{@{}l l l@{}}
\toprule
\textbf{Agent} & \textbf{Version} & \textbf{Source} \\ \midrule
\react{}                & \texttt{exgentic\_0.1.0}                                  & first-party (LiteLLM tool-calling) \\
\short{}                & \texttt{exgentic\_0.1.0} + \texttt{litellm\_1.79.1}       & first-party + tool shortlisting \\
\smol{}                 & \texttt{smolagents\_1.24.0}                               & HuggingFace \texttt{smolagents} \\
\solo{}                 & \texttt{openai\_sdk\_0.7.0}                               & \texttt{openai-agents} (Solo + \MCP{}) \\
\cc{}                   & \texttt{claude\_code\_2.1.7}                              & \texttt{@anthropic-ai/claude-code} (Docker) \\
\bottomrule
\end{tabular}
\caption{Agent versions used in every leaderboard cell. Versions for
\react{}/\short{} reflect the first-party adaptor's release; the
\short{} variant additionally pins the LiteLLM tool-calling library it
uses for shortlisting. \cc{} runs unmodified inside a Docker container
that pins the CLI version via \texttt{@anthropic-ai/claude-code@2.1.7}.}
\label{tab:agent-versions}
\end{table}

\subsection{First-Turn Instructions per Agent}
\label{appendix:agent-prompts}

The \UP{} provides each agent with three protocol fields (task,
context, actions). Each agent adaptor decides how to present these
fields to the underlying agent. Most adaptors prepend a small
agent-specific instruction string to make the protocol's expectations
explicit (e.g., that the agent is in solo mode and must finish via a
designated action); the strings are short and stable across all
benchmarks. We reproduce them verbatim below.

\paragraph{\react{} / \short{} (LiteLLM tool-calling).} No agent-side
instruction is added. The first user message simply concatenates the
task and the context dictionary, with each context field tagged by an
XML-style key:
\begin{quote}\small\ttfamily
\{task\}\\
\textless{}context\_key\_1\textgreater\\
\{context\_value\_1\}\\
\textless{}/context\_key\_1\textgreater\\
\textless{}context\_key\_2\textgreater\\
\ldots
\end{quote}
\noindent Actions are exposed as standard tool-call specifications via
\litellm{}.

\paragraph{\short{}: tool-shortlisting prompt.} The \short{} variant
adds a per-turn shortlisting step before each main call: when the
benchmark exposes more than $k=30$ actions, the same backbone LLM is
queried with the conversation-so-far and the full action catalog and
asked to return the $k$ most relevant tools as a JSON list; only those
$k$ tools are passed to the next main call. If at most $k$ actions are
available, the step is bypassed and all tools are forwarded
unchanged. The shortlisting call uses the following two-message prompt
(\texttt{developer} role + \texttt{user} role); \texttt{\{tool\_catalog\}}
is the bullet-listed \texttt{name: description} for every available
action, and \texttt{\{history\}} is a plain-text rendering of the
conversation so far:
\begin{lstlisting}[basicstyle=\small\ttfamily,frame=single,framesep=4pt,breaklines=true,breakindent=0pt,xleftmargin=4pt,xrightmargin=4pt,rulecolor=\color{black!30},backgroundcolor=\color{gray!5}]
[developer]
Please before providing your next move list the names of the top 30 tools that are somewhat relevant for the next step, ordered by relevancy (most to least). Return ONLY a JSON object with this shape: {
  "tools": ["tool_name_1", "tool_name_2", ...]
}.
Choose from these tools only: {tool_catalog}.
Do not call any of those tools just return the list of the top 30 relevant tools names in the required format.

[user]
Conversation so far (plain text):
{history}
\end{lstlisting}
\noindent A high-level component view of shortlisting is in
\S\ref{appendix:agent-components}.

\paragraph{\smol{} (Smolagent code agent).} Actions are exposed as
Python functions. The first user message is:
\begin{quote}\small\ttfamily
Task: \{task\}\\\\
Context: \{context\}\\\\
Complete this task using the available functions. Each function
corresponds to an action you can take to solve the given task.
Every action should be taken only by calling one of the functions. If
one function fail, consider using another, at any given point one of
the functions can be a valid next step. At any point you should
executing actions by writing code. do not call tools with tool calling
mechanism.\\\\
Printing or any other code will be visible only by you alone.
\end{quote}
\noindent The built-in \texttt{final\_answer} tool is removed so the
agent terminates only via the benchmark-defined finish action.

\paragraph{\solo{} (\MCP{} solo mode).} Actions are exposed as
\MCP{} tools. The first user message is:
\begin{quote}\small\ttfamily
Context: \{context\}\\\\
Complete this task using the available tools. Each tool corresponds to
an action you can take in the environment. Do not respond or ask
clarification questions unless done through a dedicated tool, and only
if such tool exist. Any plain message that is not a tool call will end
the run in failure.\\
\{task\}
\end{quote}
\noindent The agent runs in OpenAI's documented ``solo mode'' so all
agent--environment interactions go through \MCP{} tool calls.

\paragraph{\cc{} (Claude Code CLI).} Actions are exposed as
\MCP{} tools through a sidecar server; the CLI runs unmodified inside a
Docker container. The first user message is:
\begin{quote}\small\ttfamily
Context: \{context\}\\\\
Complete this task using the available environment tools. Each tool
corresponds to an action you can take in the task environment.\\
\# Important: You are on solo mode. Do not reply back or message
unless its through a dedicated environment tool call, every such
attempt will finish the session with failure.\\
All your actions on with regard to the task must go through
environment tool calls.\\
Use the designated finish tool(s): \{finish\_tool\_names\}. Always
conclude by invoking the designated finish tool for this task
environment.\\
\{task\}
\end{quote}

\paragraph{Action translation.} Each protocol action is mapped to a Python function (\smol{}), an OpenAI tool spec (\react{}, \short{}), or an \MCP{} tool (\solo{}, \cc{}); special \texttt{message} and \texttt{finish} actions follow the methodology in Appendix~\ref{app:agent-adaptation-method}.

\subsection{Architectural Components}
\label{appendix:agent-components}

We outline the key components common to general-purpose agents and, in Table~\ref{tab:agent-comparison}, summarize which components each evaluated agent provides.

\textbf{Execution Runtime.} Agents may have access to sandboxed execution environments where they can run code dynamically. For example, SmolAgents provides a Python interpreter, while Claude Code operates within a Linux machine environment. These runtime environments enable agents to execute and test code as part of their problem-solving process.

\textbf{Tool Shortlisting.} A preprocessing component that filters the available tool set before each action step, selecting a relevant subset based on current context. This improves efficiency and decision quality by focusing the agent on contextually appropriate tools, and addresses LLM constraints on tool count: when the full tool set exceeds model limits, shortlisting becomes necessary for task completion.

\textbf{Tool Schema Guard.} A component that validates actions against expected schemas before execution. When an agent attempts to call a tool or execute an environment action with incorrect parameters or structure, the schema validator raises an internal error, allowing the agent to detect and correct the mistake. This component is implemented differently across agent types: tool-calling agents typically lack explicit schema validation (relying on the LLM to generate correct calls), \MCP{}-based agents include built-in schema validation as part of the protocol, and Python-based agents receive runtime errors from the interpreter that serve a similar validation function.

\textbf{Communication Protocol.} The interface through which agents invoke tools and receive results. Agents may use direct tool-calling APIs (e.g., OpenAI function calling), code-generation approaches where the agent writes executable code, or standardized protocols like \MCP{}. Protocol choice affects action expressiveness and error handling mechanisms available to the agent.

\textbf{Memory.} Explicit storage and retrieval mechanisms beyond the conversation history. Memory components allow agents to maintain working state across turns, recall previous observations, and avoid redundant actions. Without explicit memory, agents rely solely on the LLM's context window.

\textbf{Planning.} Components that decompose tasks into structured subgoals before execution. Planning modules may generate explicit task hierarchies or action sequences, enabling more directed problem-solving. Agents without planning components select actions reactively at each step based on immediate observations.

\begin{table}[H]
\centering
\caption{Architectural components of evaluated agents.
\cmark~denotes an explicit, modular component;
\pmark~denotes an implicit or non-modular capability;
\xmark~denotes absence.}
\renewcommand{\arraystretch}{1.2}
\resizebox{1\textwidth}{!}{
    \begin{tabular}{lcccccc}
\toprule
\textbf{Agent} & \textbf{Execution Runtime} & \textbf{Tool Shortlisting} & \textbf{Tool Schema Guard} & \textbf{Communication Protocol} & \textbf{Memory} & \textbf{Planning} \\
\midrule
\react{} & \xmark & \xmark & \xmark & Tool-calling & \pmark  & \pmark  \\
\rowcolor{gray!5}
\short{} & \xmark & \cmark & \xmark & Tool-calling & \pmark  &  \pmark  \\
\smol{} & \cmark & \xmark  & \cmark & Python-Functions & \pmark  &  \pmark \\
\rowcolor{gray!5}
\solo{} & \xmark & \xmark & \cmark & \MCP{} & \pmark &  \pmark \\
\cc{} & \cmark & \xmark & \cmark & \MCP{} & \cmark & \cmark \\
\bottomrule
    \end{tabular}
}
\label{tab:agent-comparison}
\end{table}

%% file: appendices/benchmark_adaptation.tex
\begin{figure*}[h!]
    \centering
    \includegraphics[width=0.7\textwidth]{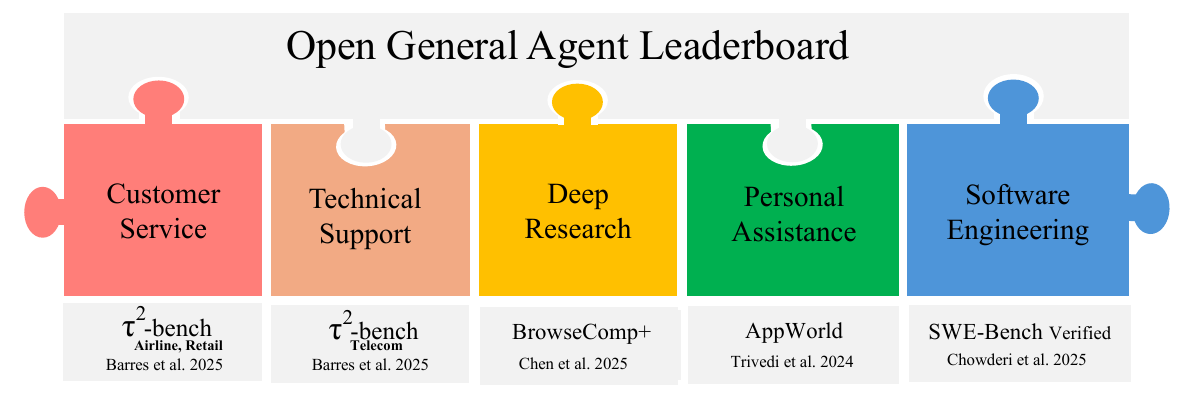}
    \caption{The six benchmarks evaluated in the \leaderboard{}, spanning software engineering, customer service, deep research, and personal assistance.}
    \label{fig:benchmarks}
\end{figure*}

For each benchmark, this appendix shows a concrete task / context /
actions example as it appears to the agent. The integration process
itself --- which reference agent we mirrored, what was kept vs.\
omitted, and what each task field encodes --- is documented in
Appendix~\ref{app:per-benchmark-adaptation}, and the released adaptor
code lives at \texttt{src/exgentic/benchmarks/<benchmark>/} in the
released repository.

\subsection{\SWEBench{} Task Definition Example}

\vspace{0.75em}
\subsubsection*{Task}
{\small
\begin{lstlisting}[basicstyle=\small\ttfamily,frame=single,framesep=4pt,breaklines=true,breakindent=0pt,xleftmargin=4pt,xrightmargin=4pt,rulecolor=\color{black!30},backgroundcolor=\color{gray!5}]
Resolve the given issue by editing the repository files directly on a remote machine.

Repository directory on the remote machine: /testbed

## Issue to resolve:
Missing call `make_hashable` on `through_fields` in `ManyToManyRel`
Description
        
In 3.2 identity property has been added to all ForeignObjectRel to make it possible to 
compare them.  A hash is derived from said identity and it's possible because identity
is a tuple.   To make limit_choices_to 
hashable (one of this tuple elements), there's a call to make_hashable.
It happens that through_fields can be a list. In such case, this make_hashable call is 
missing in ManyToManyRel.
For some reason it only fails on checking proxy model. I think proxy models have 
29 checks and normal ones 24, hence the issue, but that's just a guess.

Minimal repro:
class Parent(models.Model):
    name = models.CharField(max_length=256)
class ProxyParent(Parent):
    class Meta:
        proxy = True
class Child(models.Model):
    parent = models.ForeignKey(Parent, on_delete=models.CASCADE)
    many_to_many_field = models.ManyToManyField(
        to=Parent,
        through="ManyToManyModel",
        through_fields=['child', 'parent'],
        related_name="something"
    )
class ManyToManyModel(models.Model):
    parent = models.ForeignKey(Parent, on_delete=models.CASCADE, related_name='+')
    child = models.ForeignKey(Child, on_delete=models.CASCADE, related_name='+')
    second_child = models.ForeignKey(Child, on_delete=models.CASCADE, null=True, 
    default=None)

Which will result in:
    File "manage.py", line 23, in <module>
        main()
    File "manage.py", line 19, in main
        execute_from_command_line(sys.argv)
    
    ...
    
    File ".../django/db/models/fields/reverse_related.py", line 140, in __hash__
        return hash(self.identity)
TypeError: unhashable type: 'list'

Solution:
Add missing make_hashable call on self.through_fields in ManyToManyRel.

## Execution Environment & Access (STRICT):
All commands are executed on a remote machine that already contains the full
repository and all required system dependencies and prerequisites.

The remote machine is accessible ONLY via the `bash` action.
All interactions with the filesystem and environment, including reading files,
editing files, and running scripts, MUST be done through `bash`.

Each command is executed in a separate shell invocation; working directory changes
and environment variables do not persist between commands. Only files written to disk 
persist. All bash commands are executed with the working directory set to /.

## Instructions:
You must fix the issue by directly modifying files in this repository.

ALL file edits MUST be performed using `bash`.
Explanations or code snippets in chat are not sufficient.

Only changes written to the actual repository files will be included in the final patch.

Modify only non-test source files in /testbed.

## Hard Boundaries:
- MODIFY ONLY: regular source code files in /testbed  
- DO NOT MODIFY: tests or configuration files  
- DO NOT use interactive editors  
- DO NOT suggest changes without implementing them

## Patch & Submission Mechanics:
Submission captures the entire working tree using:
    git add -A && git diff --staged C0

This means:
- All intended fixes MUST appear in repository files
- Any file present at submission time WILL be included in the patch
- Temporary files must be removed before submission

## Recommended Workflow:
1. Analyze the codebase
2. Reproduce the issue
3. Edit using bash
4. Verify fix
5. Test edge cases

## Submission:
Use submit_patch exactly once with a short summary.

## Evaluation:
The patch will be applied to a hidden test suite and must pass all checks.
\end{lstlisting}
}

\subsubsection*{Context}
{\small
\begin{tabular}{@{}p{0.28\textwidth} p{0.68\textwidth}@{}}
\toprule
\textbf{Key} & \textbf{Value} \\
\midrule
\emph{(no entries)} &  \\
\bottomrule
\end{tabular}
}

\subsubsection*{Actions}
\begin{itemize}
  \item \texttt{bash(command: str)} --- Run a bash command in the repo root and get the output.
  \item \texttt{finish(summary: str)} --- Finish the task by submitting a brief summary; the system automatically computes the git patch from the repository changes.
\end{itemize}

\subsection{\BrowseComp{} Task Definition Example}

\vspace{0.75em}
\subsubsection*{Task}
{\small
\begin{lstlisting}[basicstyle=\small\ttfamily,frame=single,framesep=4pt,breaklines=true,breakindent=0pt,xleftmargin=4pt,xrightmargin=4pt,rulecolor=\color{black!30},backgroundcolor=\color{gray!5}]
Answer the provided question by performing search and document expansion as needed,
and submit your final answer. 
Question: I need you to name the first and last name of the production controller 
of a specific Indian film. The director of this film made one other movie the same year, 
made his directorial debut in the 1980s, and has since directed over thirty films. 
The film features the debut of an actor known for starring in a satirical
show influenced by real-world events, which is loosely adapted from a cartoon series 
published in a famous magazine. This show has more than 4,000 episodes.
This actor has also played small roles in over 60 films. The film was released between 
1990 and 2020, and a 1980s Hollywood movie inspired its storyline. 
Additionally, an actor in the main cast is featured in a film with a title connected to 
a  classic game of strategy, this film was released in the 2000s.

Note:
- The question has an answer discoverable through proper search.
- The question requires putting together information from different sources.

Your performance is scored based on:
 1. Most importantly, the correctness of the answer you assembled from different 
 searches.
 2. Your effective use of search and your ability to retrieve all relevant 
 information for the question.
 3. How efficiently you find all the relevant information, using as few searches
 as possible.

Important: During your work, Do NOT interact with the user or send any messages at any 
point - messages will be ignored and are NOT considered a valid final answer. 
The ONLY acceptable way to finish is by calling 'submit' with the required structured 
fields.

Finish the session always by calling `submit`. 
If you fail to find the answer, submit with exact_answer: "Can't find the answer.".
\end{lstlisting}
}

\subsubsection*{Context}
{\small
\begin{tabular}{@{}p{0.28\textwidth} p{0.68\textwidth}@{}}
\toprule
\textbf{Key} & \textbf{Value} \\
\midrule
\emph{(no entries)} &  \\
\bottomrule
\end{tabular}
}

\subsubsection*{Actions}
\begin{itemize}
  \item \texttt{search(query: str)} --- Perform a search on the knowledge source; retrieves the top-$k$ most relevant snippets, each trimmed to a documented token budget.
  \item \texttt{get\_document(docid: str)} --- Retrieve the full document by its document id (optional; enabled per benchmark configuration).
  \item \texttt{submit(exact\_answer: str, explanation: str, confidence: float)} --- Submit the final answer and end the session (finish action).
\end{itemize}

\subsection{\AppWorld{} Task Definition Example}

\vspace{0.75em}
\subsubsection*{Task}
{\small
\begin{lstlisting}[basicstyle=\small\ttfamily,frame=single,framesep=4pt,breaklines=true,breakindent=0pt,xleftmargin=4pt,xrightmargin=4pt,rulecolor=\color{black!30},backgroundcolor=\color{gray!5}]
Task from supervisor:
I have invited some of my friends to a reunion party via phone messages. I have made a 
CSV to track who is coming or not in ~/documents/personal/ in my file system. 
Please update RSVPs in it as per their latest replies.
\end{lstlisting}
}

\subsubsection*{Context}
{\small
\begin{tabular}{@{}p{0.28\textwidth} p{0.68\textwidth}@{}}
\toprule
\textbf{Key} & \textbf{Value} \\
\midrule
policy &  This environment provides a set of applications, each exposing a predefined set of APIs that may be used to perform tasks on behalf of the supervisor. The applications include: api\_docs, supervisor, amazon, phone, file\_system, spotify, venmo, gmail, splitwise, simple\_note, todoist.  The available applications and their APIs are fixed for the task. Supervisor account credentials (such as emails, usernames, and passwords) are available through the supervisor application's APIs and are accessed from there when required. If an application requires an access token to perform authenticated operations, the access token is obtained by calling that application's authentication/login API using the credentials retrieved from the supervisor application. Access tokens are not provided by the supervisor application. References to people (e.g., friends, family, roommates) correspond to entries in the phone\_contacts application. References to files or storage correspond to the file\_system application, not the local machine filesystem. Time-based instructions (e.g., 'this month', 'yesterday') are interpreted with full calendar boundary ranges. If an API returns paginated results, all pages constitute the complete result. The environment consists only of the provided applications and their documented APIs and parameters. No additional endpoints, methods, arguments, or capabilities are assumed beyond those explicitly defined. When task execution is finished, the designated task-completion API is used to signal completion. If the task requires a final answer value, the answer is returned through that completion API. If the task cannot be completed using the available applications and APIs, the task may be marked as failed.\\
supervisor & \{
      "first\_name": "Ashley",
      "last\_name": "Moore",
      "email": "as\_moore@gmail.com",
      "phone\_number": "7336094411"
    \}
\\    
datetime &  2023-05-18T12:00:00 \\

\bottomrule
\end{tabular}
}

\subsubsection*{Actions ($\sim$468 total)}
Generated programmatically from
\texttt{world.task.api\_docs.function\_calling()}; one \UP{} action per
AppWorld API across the nine apps, plus a benchmark-level
\texttt{finish} mapped to \texttt{supervisor.complete\_task}. Each
action's name and signature mirror the AppWorld API; descriptions are
the AppWorld-provided docstrings. Selected examples:
\begin{itemize}
  \item \texttt{finish} --- Mark the task as completed and (optionally) return a final answer; mapped from \texttt{supervisor.complete\_task}.
  \item \texttt{supervisor.show\_profile}
  \item \texttt{supervisor.show\_addresses}
  \item \texttt{supervisor.show\_payment\_cards}
  \item \texttt{supervisor.show\_account\_passwords}
  \item \texttt{amazon.show\_account}
  \item \texttt{amazon.signup}
  \item \texttt{amazon.delete\_account}
  \item \texttt{amazon.update\_account\_name}
  \item \texttt{amazon.login}

  \item \texttt{amazon.logout}
  \item \texttt{amazon.clear\_browsing\_history}
  \item \texttt{amazon.search\_sellers}
  \item \texttt{amazon.show\_cart}
  \item \texttt{amazon.update\_product\_quantity\_in\_cart}
  \item \texttt{amazon.show\_wish\_list}
  \item \texttt{amazon.update\_address}
  \item \texttt{amazon.show\_product\_reviews}
  \item \texttt{amazon.write\_product\_review}
  \item \texttt{amazon.show\_product\_questions}
  \item \textit{(\ldots\ 460+ additional tools omitted for brevity \ldots)}
  \item \texttt{phone.search\_contacts}
  \item \texttt{phone.send\_text\_message}
  \item \texttt{phone.show\_alarm}
  \item \texttt{phone.update\_alarm}

  \item \texttt{file\_system.create\_directory}
  \item \texttt{file\_system.show\_file}

  \item \texttt{spotify.show\_account}
  \item \texttt{spotify.search\_songs}

  \item \texttt{simple\_note.create\_note}

  \item \texttt{todoist.create\_task}
\end{itemize}

\subsection{\Tbench{} Task Definition Example}

\vspace{0.75em}
\subsubsection*{Task}
{\small
\begin{lstlisting}[basicstyle=\small\ttfamily,frame=single,framesep=4pt,breaklines=true,breakindent=0pt,xleftmargin=4pt,xrightmargin=4pt,rulecolor=\color{black!30},backgroundcolor=\color{gray!5}]
You are a customer service agent that helps the user according to the <policy> provided 
below.  Try to be helpful and always follow the policy.
\end{lstlisting}
}

\subsubsection*{Context}
{\small
\begin{tabular}{@{}p{0.28\textwidth} p{0.68\textwidth}@{}}
\toprule
\textbf{Key} & \textbf{Value} \\
\midrule
policy &  \# Airline Agent Policy

The current time is 2024-05-15 15:00:00 EST.

As an airline agent, you can help users **book**, **modify**, or **cancel** flight reservations. You also handle **refunds and compensation**.

Before taking any actions that update the booking database (booking, modifying flights, editing baggage, changing cabin class, or updating passenger information), you must list the action details and obtain explicit user confirmation (yes) to proceed.

You should not provide any information, knowledge, or procedures not provided by the user or available tools, or give subjective recommendations or comments.

You should only make one tool call at a time, and if you make a tool call, you should not respond to the user simultaneously. If you respond to the user, you should not make a tool call at the same time.

You should deny user requests that are against this policy.

You should transfer the user to a human agent if and only if the request cannot be handled within the scope of your actions. To transfer, first make a tool call to transfer\_to\_human\_agents, and then send the message 'YOU ARE BEING TRANSFERRED TO A HUMAN AGENT. PLEASE HOLD ON.' to the user.
[\ldots] (full task text in released repository)

\\
\bottomrule
\end{tabular}
}

\subsubsection*{Actions}
The protocol \texttt{message} action carries replies to the simulated
user; the remaining actions are translated from the
\Tbench{} domain's tool registry (one \UP{} action per registered tool,
schema and description preserved). Listing for the airline subset:
\begin{itemize}
  \item \texttt{message} --- Send a message to the user.
  \item \texttt{book\_reservation}
  \item \texttt{calculate}
  \item \texttt{cancel\_reservation}
  \item \texttt{get\_reservation\_details}
  \item \texttt{get\_user\_details}
  \item \texttt{list\_all\_airports}
  \item \texttt{search\_direct\_flight}
  \item \texttt{search\_onestop\_flight}
  \item \texttt{send\_certificate}
  \item \texttt{transfer\_to\_human\_agents}
  \item \texttt{update\_reservation\_baggages}
  \item \texttt{update\_reservation\_flights}
  \item \texttt{update\_reservation\_passengers}
  \item \texttt{get\_flight\_status}
\end{itemize}

%% file: appendices/detailed_interaction.tex
This section demonstrates a complete interaction between a code-generation agent such as SmolAgents and the \Tbench benchmark.

\textbf{Agent Side.}  
During initialization, the SmolAgent adaptor converts all \method{} actions into
lightweight Python wrapper functions. A standard SmolAgent instance is then created
using the session's task definition and the set of wrapper functions.

When the agent invokes one of these wrapper functions, the wrapper places the
corresponding action into an \emph{action queue} and blocks while waiting for a response in
an \emph{observation queue}.

Later, when the orchestrator calls:
$
\text{action} = \operatorname{CodeAgentWrapper.react}(\text{observation}),
$
the adaptor stores the observation in the \emph{observation queue}, unblocking the agent-side wrapper
function. The wrapper retrieves the observation and returns it to the agent as the result of
the function call. Meanwhile, \(\operatorname{react}(\cdot)\) waits for the next action to appear in
the \emph{action queue}.

On the next invocation of a wrapper function, the agent places a new action in the
\emph{action queue}, which releases the blocked \(\operatorname{react}(\cdot)\) call. The action is then
returned to the orchestrator, which forwards it to the benchmark session, obtains the next
observation, and calls \(\operatorname{react}(\cdot)\) again.

This cycle continues until either the agent produces no further actions or the benchmark
provides no further observations, signaling the end of the session.

\textbf{Benchmark Side.}  
During initialization in \(\operatorname{TauBenchBenchmark.start}()\), the list of available task
names is retrieved from the \Tbench codebase. When
\(\operatorname{TauBenchBenchmark.next\_session}()\) is invoked, a Session wrapper object is
constructed. This wrapper defines the textual task description for the selected task and
translates \Tbench's OpenAI tool specifications into \method{} protocol actions. It then
builds a proxy agent compatible with \Tbench's internal agent API and begins executing
\Tbench code for the selected task.

When \Tbench calls the proxy agent to obtain the next action given a simulated user
message, the proxy agent stores the message in an \emph{observation queue} and waits for an
action to appear in the \emph{action queue}. Once the orchestrator executes
$
\text{observation} = \operatorname{TauBenchBenchmark.step}(\text{action}),
$
the benchmark wrapper stores the action in the \emph{action queue}, allowing the proxy agent to resume and
forward the action to \Tbench{}. Meanwhile,
\(\operatorname{TauBenchBenchmark.step}(\cdot)\) blocks on the \emph{observation queue}. When the proxy agent is called again by the \Tbench{} code with the next simulated user message, it stores the message in the \emph{observation queue}, enabling the
observation to be returned to the orchestrator, which then passes it to the real agent.

%% file: appendices/detailed_results.tex
This appendix collects the full per-(agent, model, benchmark) results that back the headline claims in \S\ref{sec:leaderboard}: the complete leaderboard, cost-efficiency for every (architecture, model) configuration, the generalist-vs-specialist Wilson-CI comparison, the tool-shortlisting ablation, average step counts split by success/failure, and per-model failure-step dynamics. Each subsection is self-contained: the same conclusions appear in compressed form in the main body, while here we give the raw numbers, sample sizes, and methodology footnotes needed to reproduce them.

Table~\ref{tab:auto_results_leaderboard}\input{tables/auto_results_leaderboard} presents the complete per-(agent, model, benchmark) leaderboard for all five models.

\subsection{Cost-Efficiency per Configuration}
\label{app:cost_efficiency}

Table~\ref{tab:cost_efficiency} reports cost-efficiency (bench-weighted success divided by bench-weighted cost-per-task in dollars) for all 25 configurations, sorted from most to least efficient.

Tool-shortlisting cost effects (\S\ref{sec:leaderboard}): adding shortlisting to \react{} reduces \opus{} cost from \$5.75 to \$3.78 per task ($-\$1.97$), while leaving \gpt{} approximately flat (\$0.17$\to$\$0.26).

\begin{table}[H]
\small
\centering
\caption{Cost-efficiency for all 25 (architecture, model) configurations: bench-weighted success divided by bench-weighted cost per task (score/\$). Sorted by efficiency, descending.}
\label{tab:cost_efficiency}
\resizebox{0.55\textwidth}{!}{%
\begin{tabular}{@{}lccc@{}}
\toprule
\textbf{Configuration} & \textbf{Score} & \textbf{Cost/Task} & \textbf{Efficiency} \\
\midrule
\react{} + \gpt{}      & 0.41 & \$0.17 & 2.43 \\
\solo{} + \deepseek{}  & 0.32 & \$0.13 & 2.43 \\
\solo{} + \gpt{}       & 0.39 & \$0.19 & 2.03 \\
\short{} + \deepseek{} & 0.45 & \$0.23 & 1.95 \\
\cc{} + \deepseek{}    & 0.42 & \$0.21 & 1.94 \\
\short{} + \gpt{}      & 0.46 & \$0.26 & 1.79 \\
\smol{} + \deepseek{}  & 0.41 & \$0.23 & 1.76 \\
\react{} + \deepseek{} & 0.46 & \$0.38 & 1.21 \\
\solo{} + \kimi{}      & 0.25 & \$0.22 & 1.12 \\
\short{} + \kimi{}     & 0.43 & \$0.41 & 1.05 \\
\smol{} + \gpt{}       & 0.38 & \$0.36 & 1.04 \\
\cc{} + \gpt{}         & 0.39 & \$0.38 & 1.04 \\
\short{} + \gemini{}   & 0.62 & \$0.66 & 0.94 \\
\react{} + \kimi{}     & 0.43 & \$0.48 & 0.89 \\
\react{} + \gemini{}   & 0.61 & \$0.81 & 0.76 \\
\smol{} + \kimi{}      & 0.42 & \$0.66 & 0.63 \\
\cc{} + \kimi{}        & 0.30 & \$0.61 & 0.50 \\
\smol{} + \gemini{}    & 0.56 & \$1.85 & 0.30 \\
\cc{} + \gemini{}      & 0.56 & \$2.47 & 0.23 \\
\solo{} + \gemini{}    & 0.59 & \$2.81 & 0.21 \\
\short{} + \opus{}     & 0.62 & \$3.78 & 0.16 \\
\smol{} + \opus{}      & 0.66 & \$4.39 & 0.15 \\
\react{} + \opus{}     & 0.61 & \$5.75 & 0.11 \\
\solo{} + \opus{}      & 0.73 & \$8.54 & 0.09 \\
\cc{} + \opus{}        & 0.67 & \$8.03 & 0.08 \\
\bottomrule
\end{tabular}%
}
\end{table}

\subsection{References to Leaderboards}\label{app:ref_leaderboard}
For reference, \SWEBench
leaderboard top reported domain-specific agent achieves 0.79\footnote{\label{fn:swe}\smaller{\href{https://www.swebench.com/}{\SWEBench{}}}},
\BrowseComp and \AppWorld are 0.80\footnote{\label{fn:bcomp}\smaller{\href{https://huggingface.co/spaces/Tevatron/BrowseComp-Plus}{\BrowseComp{}}}}, and $0.73$\footnote{\label{fn:aworld}\smaller{\href{https://appworld.dev/leaderboard}{\AppWorld{}}}}, respectively. \Tbench Airline ($0.73$), Retail ($0.86$), and Telecom ($0.98$)\footnote{\label{fn:tau}\smaller{\href{https://taubench.com/\#leaderboard}{\Tbench{}}}}.

\subsection{Generalist vs.\ Specialist Wilson CI Comparison}
\label{app:wilson_specialist}

Table~\ref{tab:wilson_specialist} reports the per-benchmark Wilson 95\% half-width and the gap between our best generalist configuration and the top reported specialist score (\S\ref{sec:leaderboard}, App.~\ref{app:ref_leaderboard}). The \emph{Indistinguishable} verdict applies when the gap is within the per-benchmark Wilson half-width (or favors the generalist); on \BrowseComp{} (gap 19pp) and \Tbench-Telecom (gap 10pp) the specialist leads beyond Wilson uncertainty.

\begin{table}[H]
\centering
\small
\caption{Per-benchmark Wilson 95\% half-widths and generalist--specialist gaps. $n$ is the number of evaluated instances. ``HW gen/spec'' are Wilson half-widths in percentage points at the respective scores. ``Gap'' is specialist minus generalist (positive favors specialist). The four ``Indistinguishable'' rows back the body claim of parity on 4 of 6 benchmarks.}
\label{tab:wilson_specialist}
\resizebox{0.8\textwidth}{!}{%
\begin{tabular}{@{}llrrrrrl@{}}
\toprule
\textbf{Benchmark} & \textbf{Best Generalist} & \textbf{$n$} & \textbf{Gen} & \textbf{Spec} & \textbf{Gap (pp)} & \textbf{HW gen/spec (pp)} & \textbf{Verdict} \\
\midrule
\SWEBench{}     & \solo{}+\opus{}    & 100 & 0.81 & 0.79 & $-2$  & 7.5 / 8.0   & Indistinguishable \\
\BrowseComp{}   & \smol{}+\opus{}    & 100 & 0.61 & 0.80 & $+19$ & 9.5 / 8.0   & Specialist leads \\
\Tbench-Airline & \solo{}+\opus{}    & 50  & 0.74 & 0.73 & $-1$  & 12.0 / 12.5 & Indistinguishable \\
\Tbench-Retail  & \solo{}+\opus{}    & 100 & 0.85 & 0.86 & $+1$  & 7.0 / 7.0   & Indistinguishable \\
\Tbench-Telecom & \smol{}+\gemini{}  & 100 & 0.88 & 0.98 & $+10$ & 6.5 / 3.5   & Specialist leads \\
\AppWorld{}     & \smol{}+\opus{}    & 100 & 0.70 & 0.73 & $+3$  & 9.0 / 8.5   & Indistinguishable \\
\bottomrule
\end{tabular}%
}
\end{table}

\subsection{Tool Shortlisting Ablation}
\label{app:shortlisting_ablation}

Table~\ref{tab:shortlisting_ablation} isolates the per-cell impact of adding tool shortlisting to a vanilla \react{} architecture, by model. \AppWorld{} is the only tool-rich benchmark in the suite ($\sim$468 actions); shortlisting changes nothing on the others (whose action sets fit within all five LLMs' tool limits) so the bench-weighted column reduces to \nicefrac{1}{4} of the \AppWorld{} delta. The largest gain is on \gpt{} ($+22$pp on \AppWorld{}, $+5$pp aggregate), where the 128-tool API limit otherwise drives the cell to 0.00. \deepseek{} is the lone regression ($-5$pp on \AppWorld{}, $-1$pp aggregate), indicating shortlisting is broadly but not universally helpful (cf.\ \S\ref{sec:leaderboard}).

\begin{table}[H]
\centering
\small
\caption{Tool-shortlisting ablation: success-rate delta from adding shortlisting to a vanilla \react{} agent (\react{}+\short{} minus \react{}), in percentage points. Aggregate is bench-weighted across all six benchmarks.}
\label{tab:shortlisting_ablation}
\begin{tabular}{@{}lrrr@{}}
\toprule
\textbf{Model} & \textbf{\AppWorld{} $\Delta$ (pp)} & \textbf{Aggregate $\Delta$ (pp)} & \textbf{Cost $\Delta$/task} \\
\midrule
\opus{}     & $+3$  & $+0.7$ & $-\$1.97$ \\
\gemini{}   & $+5$  & $+1.2$ & $-\$0.15$ \\
\gpt{}      & $+22$ & $+5.5$ & $+\$0.09$ \\
\deepseek{} & $-5$  & $-1.3$ & $-\$0.15$ \\
\kimi{}     & $+1$  & $+0.3$ & $-\$0.07$ \\
\bottomrule
\end{tabular}
\end{table}

\subsection{Step Counts}\label{app:steps_counts}

\begin{table}[H]
  \centering
  \small
  \caption{Average steps per benchmark and architecture, split by successful vs.\ failed sessions. The three closed-source backbones (\opus{}, \gemini{}, \gpt{}) are aggregated within each (benchmark, architecture) cell. Zero-step sessions are excluded (4.0\% of closed-source runs); step counts are capped at 50 to limit outlier influence (7.1\% of remaining runs). \SWEBench{} success is patch-pass (\texttt{score}~$\geq 1$); other benchmarks use binary session success. Per-model raw values for all five backbones are in App.~\ref{app:per_model_steps}.}
  \label{tab:avg-steps-split-success-failed-cap50}
  \resizebox{\textwidth}{!}{%
  \begin{tabular}{lcccccccccc}
  \toprule
  \textbf{Benchmark} &
  \textbf{\cc{} Succ} & \textbf{\cc{} Fail} &
  \textbf{\solo{} Succ} & \textbf{\solo{} Fail} &
  \textbf{\smol{} Succ} & \textbf{\smol{} Fail} &
  \textbf{\react{} Succ} & \textbf{\react{} Fail} &
  \textbf{\short{} Succ} & \textbf{\short{} Fail} \\
  \midrule
  \AppWorld{}   & 23.67 & 38.56 & 26.41 & 39.39 & 25.69 & 34.17 & 13.24 & 27.91 & 12.34 & 21.41 \\
  \BrowseComp{} & 13.90 & 23.64 & 15.23 & 17.96 & 15.89 & 23.83 & 9.28  & 15.53 & 9.28  & 15.53 \\
  \SWEBench{}   & 27.69 & 32.21 & 27.76 & 29.30 & 29.28 & 32.03 & 28.38 & 34.22 & 28.38 & 34.22 \\
  airline       & 10.43 & 13.02 & 10.19 & 13.65 & 10.39 & 14.06 & 9.31  & 12.18 & 9.31  & 12.18 \\
  retail        & 11.73 & 11.54 & 11.36 & 9.81  & 11.48 & 11.31 & 10.81 & 11.52 & 10.81 & 11.52 \\
  telecom       & 12.98 & 12.25 & 13.06 & 13.26 & 11.99 & 12.75 & 13.23 & 15.84 & 13.23 & 15.84 \\
  \midrule
  \textbf{weighted\_avg} & \textbf{19.24} & \textbf{26.67} & \textbf{20.24} & \textbf{24.72} & \textbf{20.54} & \textbf{25.68} & \textbf{15.50} & \textbf{22.71} & \textbf{15.28}
  & \textbf{21.08} \\
  \bottomrule
  \end{tabular}%
  }
  \end{table}

\begin{table}[H]
  \centering
  \small
  \caption{Percentage difference in interactions between failed and successful runs (from the absolute counts above). Positive values mean failures take more interactions; negative values mean they take fewer. The Average row is the bench-weighted simple mean of per-benchmark percentages (1/4 each non-tau, 1/12 each tau subdomain).}
  \label{tab:failed-vs-success-steps-pct-cap50}
  \resizebox{0.65\textwidth}{!}{%
  \begin{tabular}{lrrrrr}
  \toprule
  \textbf{Benchmark} & \textbf{\shortstack{Claude\\Code}} & \textbf{\shortstack{OpenAI\\Solo}}  & \textbf{\smol{}} & \textbf{\react{}} & \textbf{\shortstack{\react{}\\Short}}  \\
  \midrule
  \AppWorld{} & 63\% & 49\% & 33\% & 111\% & 74\% \\
  \BrowseComp{}    & 70\% & 18\% & 50\% & 67\%  & 67\% \\
  \SWEBench           & 16\% & 6\%  & 9\%  & 21\%  & 21\% \\
  \Tbench-Airline         & 25\% & 34\% & 35\% & 31\%  & 31\% \\
  \Tbench-Retail           & -2\% & -14\% & -2\% & 7\%  & 7\% \\
  \Tbench-Telecom          & -6\% & 2\%  & 6\%  & 20\%  & 20\% \\
  \midrule
  \textbf{Average} & 39\% & 20\% & 26\% & 54\% & 45\% \\
  \bottomrule
  \end{tabular}%
  }
\end{table}

\subsection{Per-Model Failure-Step Dynamics}
\label{app:per_model_steps}

Table~\ref{tab:per_model_failure_steps} reports the same percentage-difference computation as Table~\ref{tab:failed-vs-success-steps-pct-cap50} but un-pooled across backbone models, so the open-weight rows are directly visible. The closed-source rows reproduce the body's bench-weighted averages within rounding when reaggregated across (\opus{}, \gemini{}, \gpt{}). The two open-weight backbones depart from the closed-source pattern most sharply on autonomous architectures (\solo{}, \cc{}): \deepseek{}+\solo{} reaches $-27\%$ (failures shorter than success), driven by the protocol-violation early-termination dynamic characterized in \S\ref{sec:generality_sinks}; \kimi{}+\solo{} and \kimi{}+\cc{} have insufficient \Tbench{} data after the same filter for some cells (``--''). Methodology is identical to Table~\ref{tab:avg-steps-split-success-failed-cap50}: zero-step sessions excluded, step counts capped at 50, \SWEBench{} success is patch-pass. Reproduction script: \texttt{tools/audit\_failure\_steps.py} in the released repository.

\begin{table}[H]
\centering
\scriptsize
\caption{Per-(model, agent) bench-weighted percentage difference in steps between failed and successful runs. Positive values mean failures take more steps; negative values mean failures terminate earlier than success. \AppWorld{}, \BrowseComp{}, and \SWEBench{} contribute weight 1/4 each; the three \Tbench{} subdomains contribute weight 1/12 each. ``--'' indicates a cell with insufficient succ/fail data after filtering.}
\label{tab:per_model_failure_steps}
\resizebox{\textwidth}{!}{%
\begin{tabular}{@{}llrrrrrrr@{}}
\toprule
\textbf{Model} & \textbf{Agent} & \textbf{App} & \textbf{BC+} & \textbf{SWE} & \textbf{T-Air} & \textbf{T-Ret} & \textbf{T-Tel} & \textbf{Bench-wgt} \\
\midrule
\opus{}     & \react{} & $+127\%$ & $+115\%$ & $+41\%$ & $+33\%$ & $+10\%$ & $+63\%$ & $+79.8\%$ \\
\opus{}     & \short{} & $+163\%$ & $+115\%$ & $+41\%$ & $+33\%$ & $+10\%$ & $+63\%$ & $+88.7\%$ \\
\opus{}     & \smol{}  & $+40\%$  & $+97\%$  & $+26\%$ & $+32\%$ & $-1\%$  & $-12\%$ & $+42.4\%$ \\
\opus{}     & \solo{}  & $+71\%$  & $+86\%$  & $+23\%$ & $+32\%$ & $-9\%$  & $+38\%$ & $+50.1\%$ \\
\opus{}     & \cc{}    & $+93\%$  & $+56\%$  & $+29\%$ & $+28\%$ & $+8\%$  & $+43\%$ & $+51.1\%$ \\
\midrule
\gemini{}   & \react{} & $+98\%$  & $+38\%$  & $+4\%$  & $+26\%$ & $+4\%$  & $+22\%$ & $+39.3\%$ \\
\gemini{}   & \short{} & $+136\%$ & $+38\%$  & $+4\%$  & $+26\%$ & $+4\%$  & $+22\%$ & $+49.0\%$ \\
\gemini{}   & \smol{}  & $+76\%$  & $+151\%$ & $+30\%$ & $+53\%$ & $+2\%$  & $+35\%$ & $+71.6\%$ \\
\gemini{}   & \solo{}  & $+33\%$  & $+19\%$  & $+14\%$ & $+20\%$ & $-9\%$  & $+24\%$ & $+19.4\%$ \\
\gemini{}   & \cc{}    & $+58\%$  & $+154\%$ & $+23\%$ & $+41\%$ & $+3\%$  & $-40\%$ & $+59.2\%$ \\
\midrule
\gpt{}      & \react{} & --       & $+16\%$  & $+20\%$ & $+31\%$ & $+8\%$  & $+11\%$ & $+17.6\%$ \\
\gpt{}      & \short{} & $+11\%$  & $+16\%$  & $+20\%$ & $+31\%$ & $+8\%$  & $+11\%$ & $+15.9\%$ \\
\gpt{}      & \smol{}  & $+91\%$  & $-23\%$  & $-3\%$  & $+27\%$ & $-3\%$  & $+15\%$ & $+19.8\%$ \\
\gpt{}      & \solo{}  & --       & $+4\%$   & $+20\%$ & $+63\%$ & $-7\%$  & $-1\%$  & $+13.9\%$ \\
\gpt{}      & \cc{}    & --       & $+14\%$  & $+7\%$  & $+20\%$ & $-9\%$  & $-4\%$  & $+7.5\%$ \\
\midrule
\rowcolor{gray!6}
\deepseek{} & \react{} & $+54\%$  & $+43\%$  & $-2\%$  & $+30\%$ & $+13\%$ & $+70\%$ & $+33.1\%$ \\
\rowcolor{gray!6}
\deepseek{} & \short{} & $+111\%$ & $+43\%$  & $-2\%$  & $+30\%$ & $+13\%$ & $+70\%$ & $+47.3\%$ \\
\rowcolor{gray!6}
\deepseek{} & \smol{}  & $+52\%$  & $-26\%$  & $+6\%$  & $+47\%$ & $+8\%$  & $+54\%$ & $+17.2\%$ \\
\rowcolor{gray!6}
\deepseek{} & \solo{}  & $-19\%$  & $-34\%$  & $-37\%$ & $+25\%$ & $-47\%$ & $-33\%$ & $-27.1\%$ \\
\rowcolor{gray!6}
\deepseek{} & \cc{}    & $+92\%$  & $+68\%$  & $-8\%$  & $+38\%$ & $+0\%$  & $+29\%$ & $+43.7\%$ \\
\midrule
\rowcolor{gray!6}
\kimi{}     & \react{} & $+192\%$ & $+58\%$  & $+0\%$  & $+27\%$ & $+11\%$ & $+20\%$ & $+67.6\%$ \\
\rowcolor{gray!6}
\kimi{}     & \short{} & $+33\%$  & $+58\%$  & $+0\%$  & $+27\%$ & $+11\%$ & $+20\%$ & $+27.7\%$ \\
\rowcolor{gray!6}
\kimi{}     & \smol{}  & $+42\%$  & $+62\%$  & $+9\%$  & $+75\%$ & $+1\%$  & $+28\%$ & $+36.7\%$ \\
\rowcolor{gray!6}
\kimi{}     & \solo{}  & $+71\%$  & $+76\%$  & $-0\%$  & --      & $-61\%$ & --      & $+37.8\%$ \\
\rowcolor{gray!6}
\kimi{}     & \cc{}    & $+37\%$  & $+131\%$ & $+2\%$  & $-5\%$  & $-35\%$ & --      & $+43.0\%$ \\
\bottomrule
\end{tabular}%
}
\end{table}

%% file: tables/auto_results_leaderboard.tex
\begin{table}[H]
\centering
\small
\caption{Full agent-model configuration leaderboard (binary success rates per benchmark, deduped from runs.csv). All 25 (5 agents × 5 models) configurations across 6 benchmarks.}
\label{tab:auto_results_leaderboard}
\begin{tabular}{llccccccc}
\toprule
\textbf{Agent} & \textbf{Model} & \textbf{App} & \textbf{BC+} & \textbf{SWE} & \textbf{T-Air} & \textbf{T-Ret} & \textbf{T-Tel} & \textbf{Mean} \\
\midrule
Solo & Opus & 0.68 & 0.61 & 0.81 & 0.74 & 0.85 & 0.84 & 0.73 \\
CC & Opus & 0.66 & 0.53 & 0.74 & 0.66 & 0.83 & 0.76 & 0.67 \\
Smol & Opus & 0.70 & 0.61 & 0.65 & 0.72 & 0.78 & 0.58 & 0.66 \\
React+Short & Gemini & 0.55 & 0.48 & 0.71 & 0.70 & 0.82 & 0.73 & 0.62 \\
React+Short & Opus & 0.64 & 0.49 & 0.61 & 0.66 & 0.78 & 0.76 & 0.62 \\
React & Gemini & 0.50 & 0.48 & 0.71 & 0.70 & 0.82 & 0.73 & 0.61 \\
React & Opus & 0.61 & 0.49 & 0.61 & 0.66 & 0.78 & 0.76 & 0.61 \\
Solo & Gemini & 0.57 & 0.33 & 0.72 & 0.62 & 0.73 & 0.79 & 0.59 \\
CC & Gemini & 0.36 & 0.51 & 0.67 & 0.70 & 0.71 & 0.71 & 0.56 \\
Smol & Gemini & 0.13 & 0.57 & 0.76 & 0.68 & 0.75 & 0.88 & 0.56 \\
React+Short & GPT & 0.22 & 0.46 & 0.57 & 0.54 & 0.73 & 0.53 & 0.46 \\
React & DeepSeek & 0.09 & 0.36 & 0.69 & 0.56 & 0.82 & 0.71 & 0.46 \\
React+Short & DeepSeek & 0.04 & 0.36 & 0.69 & 0.56 & 0.82 & 0.71 & 0.45 \\
React+Short & Kimi & 0.10 & 0.34 & 0.57 & 0.62 & 0.65 & 0.83 & 0.43 \\
React & Kimi & 0.09 & 0.34 & 0.57 & 0.62 & 0.65 & 0.83 & 0.43 \\
Smol & Kimi & 0.11 & 0.33 & 0.58 & 0.56 & 0.72 & 0.71 & 0.42 \\
CC & DeepSeek & 0.03 & 0.48 & 0.64 & 0.28 & 0.65 & 0.61 & 0.42 \\
Smol & DeepSeek & 0.13 & 0.21 & 0.56 & 0.60 & 0.77 & 0.84 & 0.41 \\
React & GPT & 0.00 & 0.46 & 0.57 & 0.54 & 0.73 & 0.53 & 0.41 \\
CC & GPT & 0.00 & 0.43 & 0.58 & 0.48 & 0.64 & 0.55 & 0.39 \\
Solo & GPT & 0.00 & 0.48 & 0.55 & 0.50 & 0.53 & 0.53 & 0.39 \\
Smol & GPT & 0.07 & 0.26 & 0.53 & 0.60 & 0.68 & 0.71 & 0.38 \\
Solo & DeepSeek & 0.06 & 0.30 & 0.74 & 0.20 & 0.19 & 0.18 & 0.32 \\
CC & Kimi & 0.08 & 0.56 & 0.52 & 0.12 & 0.03 & 0.00 & 0.30 \\
Solo & Kimi & 0.08 & 0.35 & 0.57 & 0.00 & 0.01 & 0.00 & 0.25 \\
\bottomrule
\end{tabular}
\end{table}

%% file: appendices/significance.tex
We assess the statistical significance of the benchmark results using a coordinated battery of tests matched to each axis of comparison: paired McNemar tests on binary success outcomes for shared (benchmark, task) pairs, paired \(t\)-tests on continuous scores for cross-architecture and cross-model contrasts, Wilson 95\% confidence intervals for per-benchmark success rates, and two-proportion $z$-tests for protocol-violation rate comparisons. Multiple-testing is controlled with Benjamini--Hochberg and Benjamini--Yekutieli corrections at $\alpha=0.05$; all qualitative conclusions in this appendix and \S\ref{sec:leaderboard} survive both. The evaluation consists of six benchmark configurations with the following instance counts: AppWorld (100), BrowseComp+ (100), SWE-Bench Verified (100), \Tbench Airline (50), \Tbench Retail (100), and \Tbench Telecom (100), for a total of 550 instances per agent-model configuration --- an aggregation level large enough that cross-model differences of even a few percentage points reach $p<0.0001$, while within-benchmark single-cell contrasts at $n=100$ remain noisy and are reported with their Wilson half-widths throughout.

For a single benchmark with n = 100 binary trials, the 95\% Wilson~\citep{wilson1927probable} confidence-interval half-width typically ranges from 7 to 9.5 percentage points when the observed success rate lies between 0.3 and 0.8, the region where most leading agent configurations perform. This means that differences smaller than approximately 8--10 percentage points on individual benchmarks should be interpreted cautiously, as they fall within normal statistical uncertainty.
To obtain a more stable measure, we compute a weighted aggregate score across all benchmark instances. Under the assumption that benchmarks are independent of one another, this yields an effective sample size of $n=550$ per agent-model configuration. The corresponding 95\% delta-method confidence-interval half-width for the aggregated score is substantially smaller, typically in the range of 4--5 percentage points. The paper's central claims operate at even larger aggregation levels: backbone-model contrasts aggregate across 5 architectures $\times$ 550 instances $= 2{,}750$ observations per model (giving a 95\% CI half-width of approximately $\pm 1.8$pp), and the $\eta^2$ variance decomposition in Section~\ref{sec:variance} uses all ${\sim}8{,}250$ observations across 15 closed-source agent configurations. At these sample sizes, the model-axis \opus{}-vs-\gpt{} gap (26pp, aggregated across architectures and benchmarks) and similar cross-model differences are highly significant, even though within-benchmark comparisons at $n=100$ remain noisy.
These levels of statistical uncertainty are standard across existing agentic leaderboards: most widely used agent-evaluation platforms report confidence intervals on the order of only a few percentage points, reflecting the inherent variability of evaluations on datasets of similar size.

We use two paired tests on shared (benchmark, task) outcomes to enhance statistical power. A pooled McNemar test on binary success outcomes compares each configuration against the top-ranked configuration; this is the basis for the leaderboard ranking claims in the body. The paired \(t\)-test reported in the tables below operates on continuous scores and isolates per-model agent-architecture contrasts. For the model-axis ranking in \S\ref{sec:leaderboard}, we run pairwise paired tests on continuous scores aligned across (benchmark, task, agent) tuples between each pair of models; all pairs differ at $p<0.0001$. For the agent-axis ranking we run the same paired tests on (benchmark, task, model) tuples between each pair of agent architectures; no pair reaches $p<0.1$.

We apply Benjamini--Hochberg and Benjamini--Yekutieli corrections at $\alpha=0.05$. All qualitative conclusions in this appendix and the \S\ref{sec:leaderboard} leaderboard ranking claims (cross-model paired-$t$ separation $p<0.0001$ between any two model tiers; within-model architecture pairs largely $p>0.1$) survive both corrections. Raw $p$-values are shown in the tables for transparency.

\subsection{Within-Model Scaffold Sensitivity}
\label{app:within-model}
Table~\ref{tab:auto_within_model} reports, for each model, the best vs.\ worst agent architecture by bench-weighted mean success and the paired \(t\)-test on shared (benchmark, task) pairs. This is the basis for our claim that agent-architecture choice has small effect on closed-source models (7--12pp swing) but largely determines open-weight performance (14--18pp swing).
\input{tables/auto_within_model}

\subsection{Pairwise Scaffold Comparisons}
\label{app:pairwise}
For each model, we report all pairwise agent-architecture $p$-values from a paired \(t\)-test on shared (benchmark, task) pairs (Tables~\ref{tab:auto_pairwise_opus}--\ref{tab:auto_pairwise_kimi}).
\input{tables/auto_pairwise_matrices}

\subsection{Variance Decomposition (Closed-Source)}
\label{app:variance-decomposition}
Table~\ref{tab:auto_variance} reproduces the variance decomposition reported in \S\ref{sec:variance} using two methodologies that both match the \method{} \texttt{analyse model-agent} CLI.

Two-way ANOVA on cell-level success scores for the 15 closed-source configurations yields model main effect $F(2,78)=34.95$, $p<10^{-10}$; agent main effect $F(4,78)=0.16$, $p=0.96$; benchmark $F(5,78)=25.20$, $p<10^{-14}$. The agent main effect emerges only when open-weight models are included ($F(4,136)=3.82$, $p=0.006$), reflecting their architectural sensitivity (\S\ref{sec:generality_sinks}).
\input{tables/auto_variance}

\subsection{Generality Sinks Decomposition}
\label{app:sinks-decomposition}
Per-benchmark decomposition of the open-weight gap to closed-source: a \emph{capability gap} (closed-best cell minus open-best cell) and an \emph{architectural-sensitivity gap} (open-weight per-model architectural spread minus closed-source per-model spread, averaged across models). 95\% bootstrap CIs (2{,}000 resamples, parametric Binomial bootstrap on per-cell success counts) in brackets.
\begin{table}[H]
\centering
\small
\begin{tabular}{lrrl}
\toprule
\textbf{Benchmark} & \textbf{Capability gap (pp)} & \textbf{Arch.-sens.\ gap (pp)} & \textbf{Sink type} \\
\midrule
\AppWorld{}     & $+57.0$ \hfill $[+50, +66]$ & $-18.5$ \hfill $[-25, -11]$ & Benchmark sink \\
\SWEBench{}     & $+7.0$  \hfill $[-2, +16]$  & $+0.2$  \hfill $[-9, +9]$   & (none) \\
\BrowseComp{}   & $+5.0$  \hfill $[-3, +19]$  & $+5.8$  \hfill $[-6, +16]$  & (none) \\
\Tbench-Airline & $+12.0$ \hfill $[+3, +20]$  & $+41.7$ \hfill $[+30, +47]$ & Architecture sink \\
\Tbench-Retail  & $+3.0$  \hfill $[-5, +10]$  & $+54.6$ \hfill $[+44, +60]$ & Architecture sink \\
\Tbench-Telecom & $+4.0$  \hfill $[-5, +10]$  & $+54.2$ \hfill $[+44, +61]$ & Architecture sink \\
\bottomrule
\end{tabular}
\end{table}

\subsection{Step-Zero Rates}
\label{app:zero-step}
Two-proportion $z$ tests on \Tbench{} autonomous (\cc{}~+~\solo{}) sessions, $n=500$ for each open-weight model and $n=1{,}500$ for closed-source: \kimi{} (94\%) vs closed-source (1.7\%), $z=+41.4$; \deepseek{} (31\%) vs closed-source (1.7\%), $z=+20.0$; \kimi{} vs \deepseek{}, $z=+20.5$. All pairwise $p<10^{-15}$.

\subsection{Cross-Benchmark Spearman Correlations}
\label{app:spearman}
Table~\ref{tab:auto_spearman} reports Spearman rank correlations between benchmarks across the 15 closed-model configurations.
\input{tables/auto_spearman}

\subsection{Full Leaderboard with Open-Weight Models}
\label{app:full-leaderboard}
Table~\ref{tab:auto_full_leaderboard} extends the headline leaderboard with all 25 (5 agents \(\times\) 5 models) configurations.
\input{tables/auto_full_leaderboard}

\subsection{Data Quality and Run Provenance}
\label{app:data-quality}
Table~\ref{tab:auto_data_quality} lists, per cell, the canonical run's completed/planned task counts and the total number of run rows present in \texttt{runs.csv}.
\input{tables/auto_data_quality}

%% file: tables/auto_within_model.tex
\begin{table}[H]
\centering
\small
\caption{Within-model agent-architecture sensitivity. For each model we report the best and worst agent architecture by bench-weighted mean success, plus a paired \(t\)-test on shared (benchmark, task) pairs comparing the two. Closed-source models swing 7--12pp; open-weight models swing 14--18pp with highly significant best-vs-worst tests.}
\label{tab:auto_within_model}
\begin{tabular}{lllllrrl}
\toprule
Model & Best arch.\ & Mean & Worst arch.\ & Mean & \(\Delta\)~(pp) & \(n\) & \(p\) \\
\midrule
Opus & Solo & 0.727 & React & 0.610 & +11.7 & 105 & 0.100 \\
Gemini & React+Short & 0.622 & Smol & 0.557 & +6.6 & 5 & 0.317 \\
GPT & React+Short & 0.463 & Smol & 0.380 & +8.3 & 105 & 0.005 \\
DeepSeek & React & 0.459 & Solo & 0.322 & +13.7 & 550 & \(<\)0.001 \\
Kimi & React+Short & 0.428 & Solo & 0.250 & +17.7 & 547 & \(<\)0.001 \\
\bottomrule
\end{tabular}
\end{table}

%% file: tables/auto_pairwise_matrices.tex
\begin{table}[H]
\centering
\scriptsize
\caption{Pairwise agent-architecture paired-\(t\)-test \(p\)-values for Opus. Lower triangle: \(p\); upper triangle: \(\Delta\) (row \(-\) column, in pp). \(***\,p<0.001\), \(**\,p<0.01\), \(*\,p<0.05\).}
\label{tab:auto_pairwise_opus}
\begin{tabular}{lrrrrr}
\toprule
~ & React & React+Short & Smol & Solo & CC \\
\midrule
React & --- & +0.0 & -20.0 & -7.6 & -20.0 \\
React+Short & \(<\)0.001*** & --- & +0.0 & -20.0 & -16.7 \\
Smol & 0.317 & 1.000 & --- & +0.0 & -16.7 \\
Solo & 0.100 & 0.317 & \(<\)0.001*** & --- & +0.0 \\
CC & 0.317 & 0.317 & 0.317 & \(<\)0.001*** & --- \\
\bottomrule
\end{tabular}
\end{table}

\begin{table}[H]
\centering
\scriptsize
\caption{Pairwise agent-architecture paired-\(t\)-test \(p\)-values for Gemini. Lower triangle: \(p\); upper triangle: \(\Delta\) (row \(-\) column, in pp). \(***\,p<0.001\), \(**\,p<0.01\), \(*\,p<0.05\).}
\label{tab:auto_pairwise_gemini}
\begin{tabular}{lrrrrr}
\toprule
~ & React & React+Short & Smol & Solo & CC \\
\midrule
React & --- & -4.8 & +20.0 & -5.7 & -20.0 \\
React+Short & 0.224 & --- & +20.0 & -1.0 & -20.0 \\
Smol & 0.317 & 0.317 & --- & +0.0 & -33.3 \\
Solo & 0.256 & 0.836 & 1.000 & --- & -40.0 \\
CC & 0.317 & 0.317 & 0.114 & 0.102 & --- \\
\bottomrule
\end{tabular}
\end{table}

\begin{table}[H]
\centering
\scriptsize
\caption{Pairwise agent-architecture paired-\(t\)-test \(p\)-values for GPT. Lower triangle: \(p\); upper triangle: \(\Delta\) (row \(-\) column, in pp). \(***\,p<0.001\), \(**\,p<0.01\), \(*\,p<0.05\).}
\label{tab:auto_pairwise_gpt}
\begin{tabular}{lrrrrr}
\toprule
~ & React & React+Short & Smol & Solo & CC \\
\midrule
React & --- & -21.0 & -7.6 & +0.0 & -1.9 \\
React+Short & \(<\)0.001*** & --- & +13.3 & +21.0 & +19.0 \\
Smol & 0.003** & 0.005** & --- & +7.6 & +5.7 \\
Solo & 1.000 & \(<\)0.001*** & 0.003** & --- & -1.9 \\
CC & 0.155 & \(<\)0.001*** & 0.031* & 0.155 & --- \\
\bottomrule
\end{tabular}
\end{table}

\begin{table}[H]
\centering
\scriptsize
\caption{Pairwise agent-architecture paired-\(t\)-test \(p\)-values for DeepSeek. Lower triangle: \(p\); upper triangle: \(\Delta\) (row \(-\) column, in pp). \(***\,p<0.001\), \(**\,p<0.01\), \(*\,p<0.05\).}
\label{tab:auto_pairwise_deepseek}
\begin{tabular}{lrrrrr}
\toprule
~ & React & React+Short & Smol & Solo & CC \\
\midrule
React & --- & +0.9 & -0.5 & +36.0 & +7.1 \\
React+Short & 0.025* & --- & -1.5 & +35.1 & +6.2 \\
Smol & 0.768 & 0.442 & --- & +36.5 & +7.6 \\
Solo & \(<\)0.001*** & \(<\)0.001*** & \(<\)0.001*** & --- & -28.9 \\
CC & \(<\)0.001*** & 0.003** & \(<\)0.001*** & \(<\)0.001*** & --- \\
\bottomrule
\end{tabular}
\end{table}

\begin{table}[H]
\centering
\scriptsize
\caption{Pairwise agent-architecture paired-\(t\)-test \(p\)-values for Kimi. Lower triangle: \(p\); upper triangle: \(\Delta\) (row \(-\) column, in pp). \(***\,p<0.001\), \(**\,p<0.01\), \(*\,p<0.05\).}
\label{tab:auto_pairwise_kimi}
\begin{tabular}{lrrrrr}
\toprule
~ & React & React+Short & Smol & Solo & CC \\
\midrule
React & --- & -0.2 & +1.5 & +34.2 & +29.5 \\
React+Short & 0.564 & --- & +1.7 & +34.4 & +29.7 \\
Smol & 0.442 & 0.389 & --- & +33.0 & +28.2 \\
Solo & \(<\)0.001*** & \(<\)0.001*** & \(<\)0.001*** & --- & -4.6 \\
CC & \(<\)0.001*** & \(<\)0.001*** & \(<\)0.001*** & 0.004** & --- \\
\bottomrule
\end{tabular}
\end{table}

%% file: tables/auto_variance.tex
\begin{table}[H]
\centering
\small
\caption{Variance decomposition (closed-only, 90 deduped (agent, model, bench) cells). Both methodologies match the \method{} \texttt{analyse model-agent} CLI exactly. Axis decomposition (model 27.8\% vs.\ agent 0.5\%) is the headline ratio cited in \S\ref{sec:variance}; additive decomposition over (model, architecture) cell means yields a 5.4\% interaction term.}
\label{tab:auto_variance}
\begin{tabular}{lrr}
\toprule
Component & Axis decomp. (\%) & Additive on cells (\%) \\
\midrule
Model      & 27.8 & 93.9 \\
Agent      & 0.5 & 0.7 \\
Benchmark  & 40.7 & -- \\
Interaction & -- & 5.4 \\
Residual (within-cell + interactions) & 31.0 & -- \\
\midrule
Model / Agent ratio & 57.8\(\times\) & 128.6\(\times\) \\
\bottomrule
\end{tabular}
\end{table}

%% file: tables/auto_spearman.tex
\begin{table}[H]
\centering
\small
\caption{Cross-benchmark Spearman rank correlations across the 15 closed-model configurations. Predominantly positive correlations indicate that model identity drives consistency across benchmarks.}
\label{tab:auto_spearman}
\begin{tabular}{lrrrrrr}
\toprule
~ & App & BC+ & SWE & T-Air & T-Ret & T-Tel \\
\midrule
App & --- & +0.67 & +0.60 & +0.73 & +0.81 & +0.57 \\
BC+ & +0.67 & --- & +0.62 & +0.75 & +0.60 & +0.44 \\
SWE & +0.60 & +0.62 & --- & +0.69 & +0.73 & +0.80 \\
T-Air & +0.73 & +0.75 & +0.69 & --- & +0.73 & +0.56 \\
T-Ret & +0.81 & +0.60 & +0.73 & +0.73 & --- & +0.66 \\
T-Tel & +0.57 & +0.44 & +0.80 & +0.56 & +0.66 & --- \\
\bottomrule
\end{tabular}
\end{table}

%% file: tables/auto_full_leaderboard.tex
\begin{table}[H]
\centering
\scriptsize
\caption{Full agent-configuration leaderboard: per-(agent architecture, backbone model, benchmark) success rates (binary, deduped from runs.csv via the same path the \method{} CLI uses) for all 25 agent configurations. Rightmost column is the bench-weighted mean (1/4 each non-tau2, 1/12 each tau2 subdomain).}
\label{tab:auto_full_leaderboard}
\begin{tabular}{llrrrrrrr}
\toprule
Agent & Model & App & BC+ & SWE & T-Air & T-Ret & T-Tel & Mean \\
\midrule
React & Opus & 0.61 & 0.49 & 0.61 & 0.66 & 0.78 & 0.76 & 0.610 \\
React+Short & Opus & 0.64 & 0.49 & 0.61 & 0.66 & 0.78 & 0.76 & 0.617 \\
Smol & Opus & 0.70 & 0.61 & 0.65 & 0.72 & 0.78 & 0.58 & 0.663 \\
Solo & Opus & 0.68 & 0.61 & 0.81 & 0.74 & 0.85 & 0.84 & 0.727 \\
CC & Opus & 0.66 & 0.53 & 0.74 & 0.66 & 0.83 & 0.76 & 0.670 \\
\midrule
React & Gemini & 0.50 & 0.48 & 0.71 & 0.70 & 0.82 & 0.73 & 0.610 \\
React+Short & Gemini & 0.55 & 0.48 & 0.71 & 0.70 & 0.82 & 0.73 & 0.622 \\
Smol & Gemini & 0.13 & 0.57 & 0.76 & 0.68 & 0.75 & 0.88 & 0.557 \\
Solo & Gemini & 0.57 & 0.33 & 0.72 & 0.62 & 0.73 & 0.79 & 0.585 \\
CC & Gemini & 0.36 & 0.51 & 0.67 & 0.70 & 0.71 & 0.71 & 0.562 \\
\midrule
React & GPT & 0.00 & 0.46 & 0.57 & 0.54 & 0.73 & 0.53 & 0.408 \\
React+Short & GPT & 0.22 & 0.46 & 0.57 & 0.54 & 0.73 & 0.53 & 0.463 \\
Smol & GPT & 0.07 & 0.26 & 0.53 & 0.60 & 0.68 & 0.71 & 0.380 \\
Solo & GPT & 0.00 & 0.48 & 0.55 & 0.50 & 0.53 & 0.53 & 0.386 \\
CC & GPT & 0.00 & 0.43 & 0.58 & 0.48 & 0.64 & 0.55 & 0.392 \\
\midrule
React & DeepSeek & 0.09 & 0.36 & 0.69 & 0.56 & 0.82 & 0.71 & 0.459 \\
React+Short & DeepSeek & 0.04 & 0.36 & 0.69 & 0.56 & 0.82 & 0.71 & 0.446 \\
Smol & DeepSeek & 0.13 & 0.21 & 0.56 & 0.60 & 0.77 & 0.84 & 0.409 \\
Solo & DeepSeek & 0.06 & 0.30 & 0.74 & 0.20 & 0.19 & 0.18 & 0.322 \\
CC & DeepSeek & 0.03 & 0.48 & 0.64 & 0.28 & 0.65 & 0.61 & 0.416 \\
\midrule
React & Kimi & 0.09 & 0.34 & 0.57 & 0.62 & 0.65 & 0.83 & 0.425 \\
React+Short & Kimi & 0.10 & 0.34 & 0.57 & 0.62 & 0.65 & 0.83 & 0.428 \\
Smol & Kimi & 0.11 & 0.33 & 0.58 & 0.56 & 0.72 & 0.71 & 0.420 \\
Solo & Kimi & 0.08 & 0.35 & 0.57 & 0.00 & 0.01 & 0.00 & 0.250 \\
CC & Kimi & 0.08 & 0.56 & 0.52 & 0.12 & 0.03 & 0.00 & 0.303 \\
\bottomrule
\end{tabular}
\end{table}

%% file: tables/auto_data_quality.tex
\begingroup
\scriptsize
\begin{longtable}{lllrrr}
\caption{Data quality per cell. ``Completed/Planned'' is the canonical (most-completed) run; ``\#runs'' is the number of run rows for that cell in raw \texttt{runs.csv} (a value above 1 indicates re-runs or partial-run artifacts that the dedup step collapses to a single row).}
\label{tab:auto_data_quality} \\
\toprule
Agent & Model & Benchmark & Completed & Planned & \#runs \\
\midrule
\endfirsthead
\multicolumn{6}{l}{\emph{(Table~\ref{tab:auto_data_quality} continued)}} \\
\toprule
Agent & Model & Benchmark & Completed & Planned & \#runs \\
\midrule
\endhead
\midrule
\multicolumn{6}{r}{\emph{Continued on next page}} \\
\endfoot
\bottomrule
\endlastfoot
React & Opus & App & 100 & 100 & 1 \\
React & Opus & BC+ & 0 & 100 & 1 \\
React & Opus & SWE & 99 & 99 & 1 \\
React & Opus & T-Air & 50 & 50 & 1 \\
React & Opus & T-Ret & 100 & 100 & 1 \\
React & Opus & T-Tel & 100 & 100 & 1 \\
React+Short & Opus & App & 100 & 100 & 1 \\
React+Short & Opus & BC+ & 0 & 100 & 1 \\
React+Short & Opus & SWE & 99 & 99 & 1 \\
React+Short & Opus & T-Air & 50 & 50 & 1 \\
React+Short & Opus & T-Ret & 100 & 100 & 1 \\
React+Short & Opus & T-Tel & 100 & 100 & 1 \\
Smol & Opus & App & 100 & 100 & 1 \\
Smol & Opus & BC+ & 100 & 100 & 1 \\
Smol & Opus & SWE & 100 & 100 & 1 \\
Smol & Opus & T-Air & 50 & 50 & 1 \\
Smol & Opus & T-Ret & 100 & 100 & 1 \\
Smol & Opus & T-Tel & 100 & 100 & 1 \\
Solo & Opus & App & 100 & 100 & 1 \\
Solo & Opus & BC+ & 100 & 100 & 1 \\
Solo & Opus & SWE & 83 & 83 & 1 \\
Solo & Opus & T-Air & 50 & 50 & 1 \\
Solo & Opus & T-Ret & 100 & 100 & 1 \\
Solo & Opus & T-Tel & 100 & 100 & 1 \\
CC & Opus & App & 100 & 100 & 1 \\
CC & Opus & BC+ & 0 & 51 & 1 \\
CC & Opus & SWE & 97 & 97 & 1 \\
CC & Opus & T-Air & 50 & 50 & 1 \\
CC & Opus & T-Ret & 100 & 100 & 1 \\
CC & Opus & T-Tel & 100 & 100 & 1 \\
React & Gemini & App & 99 & 100 & 1 \\
React & Gemini & BC+ & 0 & 100 & 1 \\
React & Gemini & SWE & 100 & 100 & 1 \\
React & Gemini & T-Air & 50 & 50 & 1 \\
React & Gemini & T-Ret & 100 & 100 & 1 \\
React & Gemini & T-Tel & 100 & 100 & 1 \\
React+Short & Gemini & App & 100 & 100 & 1 \\
React+Short & Gemini & BC+ & 0 & 100 & 1 \\
React+Short & Gemini & SWE & 100 & 100 & 1 \\
React+Short & Gemini & T-Air & 50 & 50 & 1 \\
React+Short & Gemini & T-Ret & 100 & 100 & 1 \\
React+Short & Gemini & T-Tel & 100 & 100 & 1 \\
Smol & Gemini & App & 100 & 100 & 1 \\
Smol & Gemini & BC+ & 0 & 100 & 1 \\
Smol & Gemini & SWE & 99 & 99 & 1 \\
Smol & Gemini & T-Air & 50 & 50 & 1 \\
Smol & Gemini & T-Ret & 100 & 100 & 1 \\
Smol & Gemini & T-Tel & 100 & 100 & 1 \\
Solo & Gemini & App & 98 & 100 & 1 \\
Solo & Gemini & BC+ & 0 & 99 & 1 \\
Solo & Gemini & SWE & 94 & 94 & 1 \\
Solo & Gemini & T-Air & 50 & 50 & 1 \\
Solo & Gemini & T-Ret & 100 & 100 & 1 \\
Solo & Gemini & T-Tel & 89 & 100 & 1 \\
CC & Gemini & App & 100 & 100 & 1 \\
CC & Gemini & BC+ & 0 & 100 & 1 \\
CC & Gemini & SWE & 100 & 100 & 1 \\
CC & Gemini & T-Air & 50 & 50 & 1 \\
CC & Gemini & T-Ret & 100 & 100 & 1 \\
CC & Gemini & T-Tel & 100 & 100 & 1 \\
React & GPT & App & 0 & 100 & 1 \\
React & GPT & BC+ & 0 & 100 & 1 \\
React & GPT & SWE & 100 & 100 & 1 \\
React & GPT & T-Air & 50 & 50 & 1 \\
React & GPT & T-Ret & 100 & 100 & 1 \\
React & GPT & T-Tel & 100 & 100 & 1 \\
React+Short & GPT & App & 100 & 100 & 1 \\
React+Short & GPT & BC+ & 0 & 100 & 1 \\
React+Short & GPT & SWE & 100 & 100 & 1 \\
React+Short & GPT & T-Air & 50 & 50 & 1 \\
React+Short & GPT & T-Ret & 100 & 100 & 1 \\
React+Short & GPT & T-Tel & 100 & 100 & 1 \\
Smol & GPT & App & 98 & 100 & 1 \\
Smol & GPT & BC+ & 0 & 100 & 1 \\
Smol & GPT & SWE & 99 & 99 & 1 \\
Smol & GPT & T-Air & 50 & 50 & 1 \\
Smol & GPT & T-Ret & 100 & 100 & 1 \\
Smol & GPT & T-Tel & 100 & 100 & 1 \\
Solo & GPT & App & 0 & 100 & 1 \\
Solo & GPT & BC+ & 100 & 100 & 1 \\
Solo & GPT & SWE & 99 & 99 & 1 \\
Solo & GPT & T-Air & 50 & 50 & 1 \\
Solo & GPT & T-Ret & 100 & 100 & 1 \\
Solo & GPT & T-Tel & 100 & 100 & 1 \\
CC & GPT & App & 0 & 100 & 1 \\
CC & GPT & BC+ & 100 & 100 & 1 \\
CC & GPT & SWE & 100 & 100 & 1 \\
CC & GPT & T-Air & 50 & 50 & 1 \\
CC & GPT & T-Ret & 100 & 100 & 1 \\
CC & GPT & T-Tel & 100 & 100 & 1 \\
React & DeepSeek & App & 99 & 100 & 1 \\
React & DeepSeek & BC+ & 99 & 100 & 1 \\
React & DeepSeek & SWE & 96 & 100 & 1 \\
React & DeepSeek & T-Air & 50 & 50 & 1 \\
React & DeepSeek & T-Ret & 100 & 100 & 1 \\
React & DeepSeek & T-Tel & 100 & 100 & 1 \\
React+Short & DeepSeek & App & 100 & 100 & 1 \\
React+Short & DeepSeek & BC+ & 99 & 100 & 1 \\
React+Short & DeepSeek & SWE & 96 & 100 & 1 \\
React+Short & DeepSeek & T-Air & 50 & 50 & 1 \\
React+Short & DeepSeek & T-Ret & 100 & 100 & 1 \\
React+Short & DeepSeek & T-Tel & 100 & 100 & 1 \\
Smol & DeepSeek & App & 100 & 100 & 1 \\
Smol & DeepSeek & BC+ & 100 & 100 & 1 \\
Smol & DeepSeek & SWE & 100 & 100 & 1 \\
Smol & DeepSeek & T-Air & 50 & 50 & 1 \\
Smol & DeepSeek & T-Ret & 98 & 100 & 1 \\
Smol & DeepSeek & T-Tel & 100 & 100 & 1 \\
Solo & DeepSeek & App & 48 & 100 & 1 \\
Solo & DeepSeek & BC+ & 62 & 100 & 1 \\
Solo & DeepSeek & SWE & 38 & 100 & 1 \\
Solo & DeepSeek & T-Air & 50 & 50 & 1 \\
Solo & DeepSeek & T-Ret & 100 & 100 & 1 \\
Solo & DeepSeek & T-Tel & 100 & 100 & 1 \\
CC & DeepSeek & App & 100 & 100 & 1 \\
CC & DeepSeek & BC+ & 100 & 100 & 1 \\
CC & DeepSeek & SWE & 100 & 100 & 1 \\
CC & DeepSeek & T-Air & 50 & 50 & 1 \\
CC & DeepSeek & T-Ret & 100 & 100 & 1 \\
CC & DeepSeek & T-Tel & 100 & 100 & 1 \\
React & Kimi & App & 99 & 100 & 1 \\
React & Kimi & BC+ & 98 & 100 & 1 \\
React & Kimi & SWE & 98 & 100 & 1 \\
React & Kimi & T-Air & 50 & 50 & 1 \\
React & Kimi & T-Ret & 99 & 100 & 1 \\
React & Kimi & T-Tel & 100 & 100 & 1 \\
React+Short & Kimi & App & 96 & 100 & 1 \\
React+Short & Kimi & BC+ & 98 & 100 & 1 \\
React+Short & Kimi & SWE & 98 & 100 & 1 \\
React+Short & Kimi & T-Air & 50 & 50 & 1 \\
React+Short & Kimi & T-Ret & 99 & 100 & 1 \\
React+Short & Kimi & T-Tel & 100 & 100 & 1 \\
Smol & Kimi & App & 100 & 100 & 1 \\
Smol & Kimi & BC+ & 100 & 100 & 1 \\
Smol & Kimi & SWE & 92 & 100 & 1 \\
Smol & Kimi & T-Air & 50 & 50 & 1 \\
Smol & Kimi & T-Ret & 98 & 100 & 1 \\
Smol & Kimi & T-Tel & 99 & 100 & 1 \\
Solo & Kimi & App & 52 & 100 & 1 \\
Solo & Kimi & BC+ & 62 & 100 & 1 \\
Solo & Kimi & SWE & 97 & 100 & 1 \\
Solo & Kimi & T-Air & 50 & 50 & 1 \\
Solo & Kimi & T-Ret & 100 & 100 & 1 \\
Solo & Kimi & T-Tel & 100 & 100 & 1 \\
CC & Kimi & App & 100 & 100 & 1 \\
CC & Kimi & BC+ & 100 & 100 & 1 \\
CC & Kimi & SWE & 98 & 100 & 1 \\
CC & Kimi & T-Air & 50 & 50 & 1 \\
CC & Kimi & T-Ret & 100 & 100 & 1 \\
CC & Kimi & T-Tel & 100 & 100 & 1 \\
\end{longtable}
\endgroup

%% file: appendices/reproducibility.tex
Appendix~\ref{appendix:reproducibility} provides everything required to
reproduce the leaderboard, the statistical analyses, and the auto-generated
appendix tables: the shared run configuration (\S\ref{app:repro-config}),
provider-default inference settings (Table~\ref{tab:repro-inference-args}),
the audit scripts that recompute every numeric claim from the raw session
log (\S\ref{app:repro-scripts}), the data layout and key methodology
conventions (\S\ref{app:repro-data}), and the exact \texttt{task\_id} pool
evaluated in every cell (\S\ref{app:task-ids}). The framework code, the
benchmark adaptors, the per-session results file (\texttt{sessions.csv}),
and the per-run aggregated file (\texttt{runs.csv}) are released alongside
the paper.

\subsection{Shared Run Configuration}
\label{app:repro-config}

Table~\ref{tab:reproducibility} summarizes the configuration shared by all runs. The full per-run configuration (agent version strings, system prompts, tool description templates, benchmark adaptor code) is released with the framework; agent versions and prompts are documented in Appendix~\ref{appendix:agent-adaptation}, per-benchmark adaptations in Appendix~\ref{sec:benchmark-examples}.

\begin{table}[H]
\centering
\small
\begin{tabular}{@{}l l@{}}
\toprule
\textbf{Parameter} & \textbf{Value} \\ \midrule
LLMs & \gpt{}, \opus{}, \gemini{} Pro, \deepseek{}, \kimi{} \\
LLM sampling & Provider API defaults (temperature, top-$p$) \\
Reasoning mode & Provider default for each model (not manually toggled) \\
Max turns per task & 100 \\
Agent architectures & \react{}, \short{}, \smol{}, \solo{}, \cc{} \\
Benchmarks & \AppWorld, \BrowseComp, \SWEBench, $\tau^2$-Bench (3 subdomains) \\
Instances per benchmark & 100 ($\tau^2$-Airline: 50); 550 total per agent-model \\
Total configurations & 150 (5 agents $\times$ 5 models $\times$ 6 benchmarks) \\
Total evaluation cost & $\sim$\$20K \\ \bottomrule
\end{tabular}
\caption{Configuration shared across all runs. Per-benchmark prompts and tool description templates follow the protocol in Appendix~\ref{sec:benchmark-examples}.}
\label{tab:reproducibility}
\end{table}

\subsection{Inference Arguments}

\paragraph{Provider defaults.} For every $(\text{agent}, \text{model}, \text{benchmark})$ cell we use the provider's documented default for every sampling and reasoning parameter. Table~\ref{tab:repro-inference-args} reports those defaults verbatim from each provider's official documentation. We do not re-tune sampling per agent, model, or benchmark, to avoid confounding the model-versus-agent comparison with hyperparameter search; the cost is that absolute scores under-estimate what a tuned configuration could achieve.

\begin{table}[H]
\centering
\small
\setlength{\tabcolsep}{6pt}
\begin{tabular}{@{}l l c c c c@{}}
\toprule
\textbf{Model} & \textbf{Endpoint} & \textbf{temp.} & \textbf{top-$p$} & \textbf{top-$k$} & \textbf{thinking} \\ \midrule
\opus{}        & AWS Bedrock     & $1.0$    & undisclosed  & undisclosed  & off \\
\gpt{}         & Azure OpenAI    & unsupported & unsupported & unsupported & off \\
\gemini{} Pro  & GCP Vertex AI   & $1.0$    & $0.95$       & $64$         & high \\
\deepseek{}    & Azure / Foundry & $1.0$    & $1.0$        & unsupported  & enabled \\
\kimi{}        & Azure / Foundry & $1.0^{*}$ & $1.0$       & unsupported  & enabled \\
\bottomrule
\end{tabular}
\caption{Provider-documented default inference settings, used verbatim
in every cell. ``undisclosed'' = the provider's documentation does not
state a default; ``unsupported'' = the parameter is not exposed by the
endpoint. \textsuperscript{*}For \kimi{}, $1.0$ is the documented
default for the Thinking mode that Azure serves; the generic
request-schema default across modes is $0.6$.}
\label{tab:repro-inference-args}
\end{table}

\subsection{Code Release and Audit Scripts}
\label{app:repro-scripts}

Every numeric claim in the paper is backed by a self-contained audit script
under \texttt{tools/} in the released code repository. Each script imports
the canonical task-level loader from \texttt{tools/\_data.py} (see
\S\ref{app:repro-data}) so all aggregations operate on the same filtered,
normalized data. Running a script prints \texttt{paper\_value} vs.\
\texttt{computed\_value} side-by-side along with an
OK\,/\,DRIFT\,/\,CONTRADICTED verdict, so any drift between the
manuscript and the released data can be detected by re-executing the
relevant audit. Table~\ref{tab:repro-scripts} maps each script to the
section, claim, or table it reproduces.

\begin{table}[H]
\resizebox{\textwidth}{!}{%
\centering
\small
\setlength{\tabcolsep}{4pt}
\begin{tabular}{@{}l l l@{}}
\toprule
\textbf{Script} & \textbf{Reproduces} & \textbf{Invocation} \\ \midrule
\texttt{audit\_failure\_steps.py}        & \S\ref{sec:failure_patterns}, App.~\ref{appendix:statistical}       & \texttt{python3 tools/audit\_failure\_steps.py} \\
\texttt{audit\_variance.py}              & \S\ref{sec:variance}, Tab.~\ref{tab:auto_variance}                  & \texttt{python3 tools/audit\_variance.py} \\
\texttt{audit\_within\_model\_spread.py} & \S\ref{sec:sensitivity}, Tab.~\ref{tab:auto_within_model}            & \texttt{python3 tools/audit\_within\_model\_spread.py} \\
\texttt{audit\_pairwise\_significance.py} & App.~\ref{appendix:statistical}: pairwise tables                    & \texttt{python3 tools/audit\_pairwise\_significance.py} \\
\texttt{audit\_spearman.py}              & \S\ref{sec:generality_sinks}, Tab.~\ref{tab:auto_spearman}          & \texttt{python3 tools/audit\_spearman.py} \\
\texttt{audit\_shortlisting.py}          & \S\ref{sec:sensitivity}: shortlisting deltas                        & \texttt{python3 tools/audit\_shortlisting.py} \\
\texttt{audit\_cost.py}                  & \S\ref{sec:failure_patterns}: cost-efficiency claims                & \texttt{python3 tools/audit\_cost.py} \\
\texttt{audit\_means.py}                 & \S\ref{sec:leaderboard}: per-config weighted means                  & \texttt{python3 tools/audit\_means.py} \\
\texttt{audit\_best\_per\_benchmark.py}  & \S\ref{sec:leaderboard}: per-benchmark winners                      & \texttt{python3 tools/audit\_best\_per\_benchmark.py} \\
\texttt{audit\_stability\_std.py}        & \S\ref{sec:leaderboard}: cross-config STD claim                     & \texttt{python3 tools/audit\_stability\_std.py} \\
\texttt{generate\_appendix\_tables.py}   & batch-regen all \texttt{tables/auto\_*.tex}                         & \texttt{python3 tools/generate\_appendix\_tables.py} \\
\texttt{run\_exgentic\_cli.py}           & end-to-end re-run via the \method{} \texttt{analyse} CLI             & \texttt{python3 tools/run\_exgentic\_cli.py} \\
\bottomrule
\end{tabular}}
\caption{Audit scripts released with the paper. Cross-references in the
``Reproduces'' column point to the corresponding section, claim cluster,
or table in this manuscript; some references resolve to the closest
available label if no granular label exists. \texttt{generate\_appendix\_tables.py}
regenerates every \texttt{tables/auto\_*.tex} file
(\texttt{auto\_full\_leaderboard}, \texttt{auto\_within\_model},
\texttt{auto\_pairwise\_<m>}, \texttt{auto\_variance}, \texttt{auto\_spearman},
\texttt{auto\_data\_quality}) in a single call.}
\label{tab:repro-scripts}
\end{table}

\subsection{Data and Environment}
\label{app:repro-data}

\paragraph{Data layout.} All audits read from a single per-session file
\path{${EXGENTIC_EXPERIMENTS_DIR}/experiments/results/sessions.csv}
and a per-run aggregate
file \texttt{runs.csv} in the same directory. The released paper
artifact bundles both files. The canonical loader is:
\begin{lstlisting}[basicstyle=\small\ttfamily,frame=single,framesep=4pt,breaklines=true,xleftmargin=4pt,xrightmargin=4pt]
from tools._data import load_tasks
tasks = load_tasks(include_open_models=True)
\end{lstlisting}
\texttt{load\_tasks} drops empty rows, the deprecated
\texttt{appworld\_test\_normal\_old} subset, the \texttt{gemini\_cli}
scaffold (no completed runs), and any directories with \texttt{\_old} or
\texttt{\_copy} suffixes; remaining rows are normalized to the schema
\{\texttt{model}, \texttt{agent}, \texttt{bench}, \texttt{task\_id},
\texttt{score}, \texttt{success}, \texttt{cost}, \texttt{steps}\}.

\paragraph{Score field convention.} Two fields are exposed per task:
\texttt{score} (continuous in $[0,1]$, partial credit) and
\texttt{success} (binary). The \method{} CLI's \texttt{analyse} subcommand
explicitly aggregates \texttt{benchmark\_score} (per-config binary
success rate), so paper claims should be checked against the binary
\texttt{success} field; the variance decomposition, pairwise
significance tests, and within-model spread all use \texttt{success}.

\paragraph{SWE-Bench is the exception.} For SWE-Bench specifically, the
session-completion field (\texttt{success}) sits near 99\%, while the
patch-pass criterion (\texttt{score} $\geq 1$) sits near 37\%; using
\texttt{success} would misclassify real failures as successes. The
failure-step gap analysis (\S\ref{sec:failure_patterns},
\texttt{audit\_failure\_steps.py}) therefore uses \texttt{score} $\geq 1$
as the success criterion on SWE-Bench and \texttt{success} elsewhere.
This is the single most important methodology note for any re-run of
the failure-step gap.

\paragraph{Bench-weighted average.} The headline weighted mean is
$\tfrac{1}{4}$ each for \AppWorld, \BrowseComp, and \SWEBench, and
$\tfrac{1}{12}$ each for the three $\tau^2$-Bench subdomains, so the
$\tau^2$ family contributes $\tfrac{1}{4}$ in total (matching paper
\S\ref{sec:leaderboard}). \texttt{tools/\_data.py} exports
\texttt{BENCH\_WEIGHTS} and \texttt{weighted\_mean\_over\_benches} helpers
so every script applies this convention identically.

\paragraph{Closed vs.\ open-weight subsets.} The paper's primary
configurations use the closed-only subset (\opus{}, \gemini{}, \gpt{};
$3 \times 5 \times 6 = 90$ cells). \texttt{include\_open\_models=True}
opts in to \deepseek{} and \kimi{} for the open-weight extension.

\paragraph{Compute footprint.} Total inference spend across the 150
$(\text{agent}, \text{model}, \text{benchmark})$ cells, paid against
provider APIs (AWS Bedrock, Azure OpenAI, GCP Vertex AI, Azure AI
Foundry), came to $\sim$\$20K. Per-task cost varies by roughly
30$\times$ across the 15 closed configurations (\S\ref{sec:failure_patterns},
App.~\ref{app:cost_efficiency}, Tab.~\ref{tab:cost_efficiency}): the
efficiency frontier sits at $\sim$\$2.43\,score/\$ (\react{}+\gpt{},
\solo{}+\deepseek{}) and the lowest-efficiency cells at $\sim$\$0.08\,score/\$
(\cc{}+\opus{}). \texttt{audit\_cost.py} prints the full per-cell
score-and-cost breakdown that backs Tab.~\ref{tab:cost_efficiency}, so a
practitioner can read off the expected API spend for re-running any
single cell.

\subsection{Task IDs Used per Benchmark}
\label{app:task-ids}

We report below the exact \texttt{task\_id} values evaluated in every
agent--model cell ($550$ in total: $100$ per benchmark; $50$ for
\Tbench-Airline). These IDs match the \texttt{task\_id} column of the
released per-session results file and uniquely pin which subset of
each benchmark was run, independent of any sampling seed or upstream
benchmark revision.

\paragraph{\SWEBench{} ($n{=}100$).}
\begin{lstlisting}[basicstyle=\scriptsize\ttfamily,frame=single,framesep=4pt,breaklines=true,breakindent=0pt,xleftmargin=4pt,xrightmargin=4pt,rulecolor=\color{black!30},backgroundcolor=\color{gray!5}]
astropy__astropy-13453
astropy__astropy-13977
astropy__astropy-14096
astropy__astropy-14369
astropy__astropy-8872
django__django-10973
django__django-11163
django__django-11239
django__django-11299
django__django-11477
django__django-11551
django__django-11749
django__django-11999
django__django-12143
django__django-12193
django__django-12273
django__django-12754
django__django-12774
django__django-13012
django__django-13023
django__django-13028
django__django-13033
django__django-13089
django__django-13128
django__django-13568
django__django-13794
django__django-13809
django__django-13820
django__django-14017
django__django-14053
django__django-14351
django__django-14404
django__django-14434
django__django-14493
django__django-14559
django__django-14608
django__django-14631
django__django-14672
django__django-14792
django__django-15037
django__django-15103
django__django-15128
django__django-15252
django__django-15277
django__django-15315
django__django-15368
django__django-15569
django__django-15851
django__django-15916
django__django-15930
django__django-15957
django__django-15987
django__django-16082
django__django-16116
django__django-16263
django__django-16454
django__django-16485
django__django-16901
django__django-16938
django__django-17084
django__django-7530
matplotlib__matplotlib-22871
matplotlib__matplotlib-25332
pallets__flask-5014
psf__requests-1766
psf__requests-6028
pydata__xarray-3993
pydata__xarray-4075
pydata__xarray-6938
pylint-dev__pylint-4551
pylint-dev__pylint-6903
pytest-dev__pytest-5262
pytest-dev__pytest-5631
pytest-dev__pytest-7205
pytest-dev__pytest-7521
scikit-learn__scikit-learn-13135
scikit-learn__scikit-learn-14629
scikit-learn__scikit-learn-14710
scikit-learn__scikit-learn-14894
scikit-learn__scikit-learn-25232
sphinx-doc__sphinx-10449
sphinx-doc__sphinx-7440
sphinx-doc__sphinx-7757
sphinx-doc__sphinx-7910
sphinx-doc__sphinx-7985
sphinx-doc__sphinx-8459
sphinx-doc__sphinx-8593
sphinx-doc__sphinx-8595
sphinx-doc__sphinx-9281
sphinx-doc__sphinx-9591
sympy__sympy-11618
sympy__sympy-12096
sympy__sympy-13551
sympy__sympy-13852
sympy__sympy-13974
sympy__sympy-14248
sympy__sympy-15976
sympy__sympy-17655
sympy__sympy-19040
sympy__sympy-20916
\end{lstlisting}

\paragraph{\AppWorld{} ($n{=}100$).}
\begin{lstlisting}[basicstyle=\scriptsize\ttfamily,frame=single,framesep=4pt,breaklines=true,breakindent=0pt,xleftmargin=4pt,xrightmargin=4pt,rulecolor=\color{black!30},backgroundcolor=\color{gray!5}]
024c982_1
024c982_3
042a9fc_2
09b0ee6_2
09b0ee6_3
0a9d82a_1
0d01c76_3
0de03ea_2
1150ed6_3
166f4ff_1
166f4ff_2
21abae1_2
21abae1_3
270f1ff_1
270f1ff_3
29a7b7e_1
29a7b7e_2
29a7b7e_3
2d9f728_1
2d9f728_3
31dc501_1
31dc501_2
31dc501_3
325d6ec_2
32616b5_2
3aa1a22_2
3aa1a22_3
3b8fb7a_1
3b8fb7a_2
3d9a636_2
3d9a636_3
425a494_1
425a494_3
522e5e5_1
552869a_1
552869a_2
59fae45_1
59fae45_2
5a83b05_1
5a83b05_2
634f342_1
634f342_3
652485c_2
6b6ca61_1
6b6ca61_2
6b6ca61_3
6f4b9a5_3
7847649_1
7847649_3
83a7951_2
83a7951_3
8749218_2
8749218_3
8ce6779_1
8ce6779_3
9016950_2
9016950_3
90adc3f_1
90adc3f_2
90adc3f_3
986aa4e_2
986aa4e_3
9dabbc9_2
9dabbc9_3
9ef798c_2
9ef798c_3
a30375d_1
a30375d_2
a30375d_3
afc4005_1
afc4005_3
b6d1104_2
b9c5c9a_1
b9c5c9a_2
bde252e_2
bde252e_3
c77c005_2
c77c005_3
ccf4b82_1
ccf4b82_3
cef9191_1
cef9191_2
d18139b_1
d18139b_2
d18139b_3
d194965_2
d194965_3
d6ac34d_1
d6ac34d_2
d6ac34d_3
dac78d9_1
dac78d9_2
dac78d9_3
f323bae_3
f3f60f0_2
f3f60f0_3
f861c32_1
fd1f8fa_1
fd1f8fa_3
ff58e36_1
\end{lstlisting}

\paragraph{\BrowseComp{} ($n{=}100$).}
\begin{lstlisting}[basicstyle=\scriptsize\ttfamily,frame=single,framesep=4pt,breaklines=true,breakindent=0pt,xleftmargin=4pt,xrightmargin=4pt,rulecolor=\color{black!30},backgroundcolor=\color{gray!5}]
6, 8, 23, 85, 97, 127, 134, 152, 169, 177,
184, 200, 201, 209, 237, 242, 251, 270, 315, 327,
337, 347, 362, 372, 395, 413, 418, 438, 443, 469,
491, 498, 519, 524, 553, 555, 558, 562, 571, 581,
582, 587, 591, 592, 598, 601, 618, 620, 637, 650,
661, 666, 672, 675, 686, 707, 714, 717, 722, 730,
756, 771, 772, 785, 801, 810, 828, 830, 834, 873,
876, 882, 898, 930, 947, 960, 969, 984, 991, 992,
1021, 1058, 1079, 1096, 1121, 1147, 1161, 1162, 1176, 1191,
1198, 1203, 1208, 1211, 1212, 1220, 1227, 1237, 1239, 1264
\end{lstlisting}

\paragraph{\Tbench-Airline ($n{=}50$).}
IDs \texttt{0} through \texttt{49} (the full benchmark task pool, in order).

\paragraph{\Tbench-Retail ($n{=}100$).}
\begin{lstlisting}[basicstyle=\scriptsize\ttfamily,frame=single,framesep=4pt,breaklines=true,breakindent=0pt,xleftmargin=4pt,xrightmargin=4pt,rulecolor=\color{black!30},backgroundcolor=\color{gray!5}]
1, 2, 3, 4, 5, 6, 7, 8, 9, 10,
11, 12, 13, 14, 16, 17, 18, 19, 22, 23,
24, 25, 26, 27, 28, 29, 30, 31, 32, 33,
34, 35, 36, 37, 38, 39, 40, 41, 42, 43,
46, 47, 48, 49, 50, 51, 52, 53, 55, 56,
57, 58, 59, 60, 61, 63, 64, 65, 66, 68,
69, 70, 71, 72, 75, 77, 78, 79, 80, 81,
82, 83, 84, 85, 86, 87, 88, 89, 90, 91,
92, 93, 94, 95, 96, 97, 98, 99, 100, 101,
102, 103, 104, 106, 107, 108, 109, 110, 112, 113
\end{lstlisting}

\paragraph{\Tbench-Telecom ($n{=}100$).}
\begin{lstlisting}[basicstyle=\tiny\ttfamily,frame=single,framesep=4pt,breaklines=true,breakindent=0pt,xleftmargin=4pt,xrightmargin=4pt,rulecolor=\color{black!30},backgroundcolor=\color{gray!5}]
[mms_issue]airplane_mode_on|bad_network_preference|bad_wifi_calling|break_apn_mms_setting|break_app_both_permissions|data_mode_off|data_usage_exceeded|unseat_sim_card|user_abroad_roaming_disabled_off[PERSONA:Hard]
[mms_issue]airplane_mode_on|bad_network_preference|bad_wifi_calling|break_apn_mms_setting|break_app_both_permissions|data_mode_off|data_usage_exceeded|unseat_sim_card|user_abroad_roaming_disabled_on[PERSONA:Hard]
[mms_issue]airplane_mode_on|bad_network_preference|bad_wifi_calling|break_apn_mms_setting|break_app_both_permissions|unseat_sim_card|user_abroad_roaming_enabled_off[PERSONA:Hard]
[mms_issue]airplane_mode_on|bad_network_preference|bad_wifi_calling|break_apn_mms_setting|break_app_sms_permission|data_mode_off|data_usage_exceeded|unseat_sim_card|user_abroad_roaming_disabled_on[PERSONA:None]
[mms_issue]airplane_mode_on|bad_network_preference|bad_wifi_calling|break_apn_mms_setting|break_app_sms_permission|data_mode_off|data_usage_exceeded|unseat_sim_card|user_abroad_roaming_enabled_off[PERSONA:None]
[mms_issue]airplane_mode_on|bad_network_preference|bad_wifi_calling|break_apn_mms_setting|break_app_storage_permission|data_mode_off|data_usage_exceeded|unseat_sim_card[PERSONA:Easy]
[mms_issue]airplane_mode_on|bad_network_preference|bad_wifi_calling|break_apn_mms_setting|break_app_storage_permission|data_mode_off|data_usage_exceeded|unseat_sim_card|user_abroad_roaming_disabled_off[PERSONA:Easy]
[mms_issue]airplane_mode_on|bad_network_preference|bad_wifi_calling|break_app_sms_permission|data_mode_off|data_usage_exceeded|unseat_sim_card|user_abroad_roaming_disabled_off[PERSONA:None]
[mms_issue]airplane_mode_on|bad_network_preference|bad_wifi_calling|break_app_storage_permission|data_mode_off|data_usage_exceeded|unseat_sim_card|user_abroad_roaming_enabled_off[PERSONA:Hard]
[mms_issue]airplane_mode_on|bad_network_preference|bad_wifi_calling|break_app_storage_permission|unseat_sim_card[PERSONA:Hard]
[mms_issue]airplane_mode_on|bad_network_preference|bad_wifi_calling|data_usage_exceeded|unseat_sim_card|user_abroad_roaming_disabled_on[PERSONA:Easy]
[mms_issue]airplane_mode_on|bad_network_preference|break_apn_mms_setting|break_app_both_permissions|data_mode_off|data_usage_exceeded|user_abroad_roaming_enabled_off[PERSONA:None]
[mms_issue]airplane_mode_on|bad_network_preference|break_apn_mms_setting|break_app_storage_permission[PERSONA:None]
[mms_issue]airplane_mode_on|bad_network_preference|break_apn_mms_setting|data_mode_off|unseat_sim_card|user_abroad_roaming_disabled_off[PERSONA:Easy]
[mms_issue]airplane_mode_on|bad_network_preference|break_apn_mms_setting|data_usage_exceeded[PERSONA:Hard]
[mms_issue]airplane_mode_on|bad_network_preference|break_app_both_permissions|data_usage_exceeded|unseat_sim_card|user_abroad_roaming_disabled_on[PERSONA:Hard]
[mms_issue]airplane_mode_on|bad_network_preference|break_app_storage_permission|data_mode_off|user_abroad_roaming_enabled_off[PERSONA:Easy]
[mms_issue]airplane_mode_on|bad_wifi_calling|break_app_both_permissions|data_mode_off|data_usage_exceeded|unseat_sim_card|user_abroad_roaming_enabled_off[PERSONA:None]
[mms_issue]airplane_mode_on|bad_wifi_calling|user_abroad_roaming_enabled_off[PERSONA:Easy]
[mms_issue]airplane_mode_on|break_app_both_permissions[PERSONA:Hard]
[mms_issue]airplane_mode_on|break_app_both_permissions|data_usage_exceeded|user_abroad_roaming_disabled_off[PERSONA:None]
[mms_issue]bad_network_preference|bad_wifi_calling|break_apn_mms_setting|break_app_storage_permission|data_mode_off|data_usage_exceeded|unseat_sim_card[PERSONA:Easy]
[mms_issue]bad_network_preference|bad_wifi_calling|break_app_both_permissions|data_usage_exceeded|unseat_sim_card|user_abroad_roaming_disabled_off[PERSONA:None]
[mms_issue]bad_network_preference|bad_wifi_calling|break_app_both_permissions|data_usage_exceeded|user_abroad_roaming_disabled_off[PERSONA:Hard]
[mms_issue]bad_network_preference|bad_wifi_calling|break_app_sms_permission|data_mode_off|data_usage_exceeded|unseat_sim_card|user_abroad_roaming_enabled_off[PERSONA:Easy]
[mms_issue]bad_network_preference|bad_wifi_calling|break_app_sms_permission|data_mode_off|unseat_sim_card|user_abroad_roaming_enabled_off[PERSONA:Hard]
[mms_issue]bad_network_preference|break_app_both_permissions[PERSONA:Easy]
[mms_issue]bad_network_preference|break_app_sms_permission|data_mode_off|data_usage_exceeded|user_abroad_roaming_enabled_off[PERSONA:None]
[mms_issue]bad_network_preference|break_app_sms_permission|user_abroad_roaming_disabled_on[PERSONA:Hard]
[mms_issue]bad_network_preference|user_abroad_roaming_disabled_off[PERSONA:None]
[mms_issue]bad_wifi_calling|break_apn_mms_setting|break_app_both_permissions|data_mode_off|data_usage_exceeded|user_abroad_roaming_disabled_off[PERSONA:None]
[mms_issue]bad_wifi_calling|break_apn_mms_setting|break_app_sms_permission|data_usage_exceeded[PERSONA:Hard]
[mms_issue]bad_wifi_calling|break_apn_mms_setting|data_mode_off|data_usage_exceeded|unseat_sim_card[PERSONA:None]
[mms_issue]bad_wifi_calling|break_apn_mms_setting|unseat_sim_card|user_abroad_roaming_enabled_off[PERSONA:Easy]
[mms_issue]break_apn_mms_setting|data_mode_off|data_usage_exceeded|user_abroad_roaming_disabled_on[PERSONA:Hard]
[mms_issue]break_apn_mms_setting|data_mode_off|user_abroad_roaming_disabled_on[PERSONA:Hard]
[mms_issue]break_apn_mms_setting|user_abroad_roaming_enabled_off[PERSONA:Hard]
[mms_issue]break_app_both_permissions|data_usage_exceeded[PERSONA:Hard]
[mms_issue]break_app_both_permissions|unseat_sim_card|user_abroad_roaming_disabled_on[PERSONA:Hard]
[mms_issue]break_app_sms_permission|data_mode_off[PERSONA:None]
[mms_issue]break_app_sms_permission|data_usage_exceeded|user_abroad_roaming_disabled_on[PERSONA:None]
[mms_issue]break_app_storage_permission|data_usage_exceeded[PERSONA:Easy]
[mms_issue]break_app_storage_permission|unseat_sim_card|user_abroad_roaming_disabled_on[PERSONA:Easy]
[mobile_data_issue]airplane_mode_on|bad_network_preference[PERSONA:Hard]
[mobile_data_issue]airplane_mode_on|bad_network_preference|bad_vpn|data_mode_off|data_saver_mode_on|data_usage_exceeded|user_abroad_roaming_disabled_off[PERSONA:Hard]
[mobile_data_issue]airplane_mode_on|bad_network_preference|bad_vpn|data_mode_off|data_saver_mode_on|data_usage_exceeded|user_abroad_roaming_disabled_on[PERSONA:None]
[mobile_data_issue]airplane_mode_on|bad_network_preference|bad_vpn|data_mode_off|data_saver_mode_on|data_usage_exceeded|user_abroad_roaming_enabled_off[PERSONA:Easy]
[mobile_data_issue]airplane_mode_on|bad_network_preference|bad_vpn|data_mode_off|data_saver_mode_on|user_abroad_roaming_disabled_off[PERSONA:Hard]
[mobile_data_issue]airplane_mode_on|bad_network_preference|bad_vpn|data_mode_off|data_usage_exceeded|user_abroad_roaming_enabled_off[PERSONA:Easy]
[mobile_data_issue]airplane_mode_on|bad_network_preference|bad_vpn|data_saver_mode_on|data_usage_exceeded|user_abroad_roaming_enabled_off[PERSONA:None]
[mobile_data_issue]airplane_mode_on|bad_network_preference|bad_vpn|data_saver_mode_on|user_abroad_roaming_disabled_on[PERSONA:Easy]
[mobile_data_issue]airplane_mode_on|bad_network_preference|data_mode_off|data_saver_mode_on[PERSONA:Hard]
[mobile_data_issue]airplane_mode_on|bad_network_preference|data_mode_off|data_saver_mode_on|data_usage_exceeded|user_abroad_roaming_disabled_on[PERSONA:None]
[mobile_data_issue]airplane_mode_on|bad_network_preference|data_mode_off|data_usage_exceeded|user_abroad_roaming_disabled_on[PERSONA:None]
[mobile_data_issue]airplane_mode_on|bad_network_preference|data_mode_off|user_abroad_roaming_disabled_on[PERSONA:Hard]
[mobile_data_issue]airplane_mode_on|bad_network_preference|data_saver_mode_on|data_usage_exceeded|user_abroad_roaming_disabled_off[PERSONA:None]
[mobile_data_issue]airplane_mode_on|data_mode_off[PERSONA:None]
[mobile_data_issue]airplane_mode_on|data_mode_off|data_saver_mode_on|data_usage_exceeded|user_abroad_roaming_enabled_off[PERSONA:Hard]
[mobile_data_issue]airplane_mode_on|data_saver_mode_on|user_abroad_roaming_disabled_on[PERSONA:None]
[mobile_data_issue]airplane_mode_on|user_abroad_roaming_enabled_off[PERSONA:None]
[mobile_data_issue]bad_network_preference|bad_vpn|data_mode_off|data_saver_mode_on[PERSONA:None]
[mobile_data_issue]bad_network_preference|bad_vpn|data_mode_off|data_saver_mode_on|data_usage_exceeded|user_abroad_roaming_enabled_off[PERSONA:Easy]
[mobile_data_issue]bad_network_preference|bad_vpn|data_saver_mode_on|data_usage_exceeded|user_abroad_roaming_disabled_off[PERSONA:Easy]
[mobile_data_issue]bad_network_preference|bad_vpn|user_abroad_roaming_disabled_off[PERSONA:Hard]
[mobile_data_issue]bad_network_preference|bad_vpn|user_abroad_roaming_disabled_on[PERSONA:None]
[mobile_data_issue]bad_network_preference|bad_vpn|user_abroad_roaming_enabled_off[PERSONA:Easy]
[mobile_data_issue]bad_network_preference|data_mode_off|data_saver_mode_on|data_usage_exceeded|user_abroad_roaming_disabled_off[PERSONA:Easy]
[mobile_data_issue]bad_network_preference|data_saver_mode_on|data_usage_exceeded[PERSONA:Hard]
[mobile_data_issue]bad_network_preference|user_abroad_roaming_enabled_off[PERSONA:Hard]
[mobile_data_issue]bad_vpn|data_mode_off|data_usage_exceeded|user_abroad_roaming_disabled_off[PERSONA:None]
[mobile_data_issue]bad_vpn|data_saver_mode_on|user_abroad_roaming_disabled_on[PERSONA:None]
[mobile_data_issue]data_mode_off|data_usage_exceeded|user_abroad_roaming_disabled_off[PERSONA:Hard]
[mobile_data_issue]data_saver_mode_on|data_usage_exceeded[PERSONA:Easy]
[mobile_data_issue]data_saver_mode_on|user_abroad_roaming_enabled_off[PERSONA:Easy]
[mobile_data_issue]data_usage_exceeded|user_abroad_roaming_enabled_off[PERSONA:Easy]
[service_issue]airplane_mode_on|break_apn_settings|contract_end_suspension|lock_sim_card_pin[PERSONA:None]
[service_issue]airplane_mode_on|break_apn_settings|contract_end_suspension|lock_sim_card_pin|unseat_sim_card[PERSONA:Easy]
[service_issue]airplane_mode_on|break_apn_settings|contract_end_suspension|unseat_sim_card[PERSONA:Easy]
[service_issue]airplane_mode_on|break_apn_settings|lock_sim_card_pin[PERSONA:None]
[service_issue]airplane_mode_on|break_apn_settings|lock_sim_card_pin|overdue_bill_suspension[PERSONA:Hard]
[service_issue]airplane_mode_on|break_apn_settings|lock_sim_card_pin|overdue_bill_suspension|unseat_sim_card[PERSONA:None]
[service_issue]airplane_mode_on|break_apn_settings|lock_sim_card_pin|unseat_sim_card[PERSONA:None]
[service_issue]airplane_mode_on|break_apn_settings|overdue_bill_suspension[PERSONA:None]
[service_issue]airplane_mode_on|break_apn_settings|unseat_sim_card[PERSONA:None]
[service_issue]airplane_mode_on|contract_end_suspension|lock_sim_card_pin|unseat_sim_card[PERSONA:Hard]
[service_issue]airplane_mode_on|lock_sim_card_pin[PERSONA:Easy]
[service_issue]airplane_mode_on|lock_sim_card_pin|overdue_bill_suspension[PERSONA:Easy]
[service_issue]airplane_mode_on|lock_sim_card_pin|overdue_bill_suspension|unseat_sim_card[PERSONA:Easy]
[service_issue]airplane_mode_on|lock_sim_card_pin|unseat_sim_card[PERSONA:Hard]
[service_issue]airplane_mode_on|overdue_bill_suspension[PERSONA:None]
[service_issue]airplane_mode_on|overdue_bill_suspension|unseat_sim_card[PERSONA:Easy]
[service_issue]airplane_mode_on|unseat_sim_card[PERSONA:None]
[service_issue]break_apn_settings|lock_sim_card_pin[PERSONA:None]
[service_issue]break_apn_settings|lock_sim_card_pin|overdue_bill_suspension[PERSONA:Easy]
[service_issue]break_apn_settings|lock_sim_card_pin|overdue_bill_suspension|unseat_sim_card[PERSONA:Easy]
[service_issue]break_apn_settings|overdue_bill_suspension|unseat_sim_card[PERSONA:Hard]
[service_issue]contract_end_suspension|lock_sim_card_pin[PERSONA:Hard]
[service_issue]contract_end_suspension|unseat_sim_card[PERSONA:Hard]
[service_issue]lock_sim_card_pin|overdue_bill_suspension[PERSONA:Easy]
[service_issue]overdue_bill_suspension|unseat_sim_card[PERSONA:Easy]
\end{lstlisting}

%% file: appendices/agentic_errormap.tex
This appendix characterizes how general agents fail. We adapt \textsc{ErrorMap}~\citep{ashury2026errormap} to long-horizon trajectories and produce 27 behavioral failure categories over all failed sessions with full trajectory data, validated on a 100-record sample. The analysis is descriptive: it documents the shape of the failure surface in our corpus and is not powered for causal claims about model, architecture, or benchmark effects.

\subsection{Setup}

We adapt \textsc{ErrorMap} to agent trajectories produced by the
\method{}, using the five agent architectures (\cc{},
\react{}, \short{}, \smol{}, \solo{}), five LM backbones (\opus{},
\deepseek{}, \gemini{} Pro, \gpt{}, \kimi{}), and six benchmarks
(\AppWorld{}, \BrowseComp{}, \SWEBench{}, and three \Tbench{}
variants) as the leaderboard. We retain all failed sessions with trajectory data. Zero-step sessions
(run-level orchestrator failures, including the cases that
dominate the open-weight architecture sink on \Tbench{} in
\S\ref{sec:generality_sinks}) lack trajectories and are excluded
by construction. Of the retained sessions, 76.7\% have a
peer-success ``gold'' trajectory from another architecture or model on
the same task, used as the judge's reference.

The port required only a trajectory-to-CSV adapter and a
QA$\rightarrow$trajectory vocabulary refresh of the prompt
templates; \textsc{ErrorMap}'s clustering pipeline, schema, and
output format are unchanged. We use \texttt{gpt-5.5} as the
judge with the upstream default \texttt{max\_num\_clusters=25},
and run end-to-end in 42~minutes for \$246. The first run
produced 26 top-level categories with \emph{Other}~$=$~25.8\%; we
manually linked synonyms (released as a deterministic mapping) to
reach the final 27 categories with \emph{Other}~$=$~2.1\%
(\textsc{ErrorMap} reports 4.8\%). The adapted code and all
analysis artifacts are released.

\subsection{Distribution of Failure Categories}

\begin{figure}[h]
\centering
\includegraphics[width=0.65\textwidth]{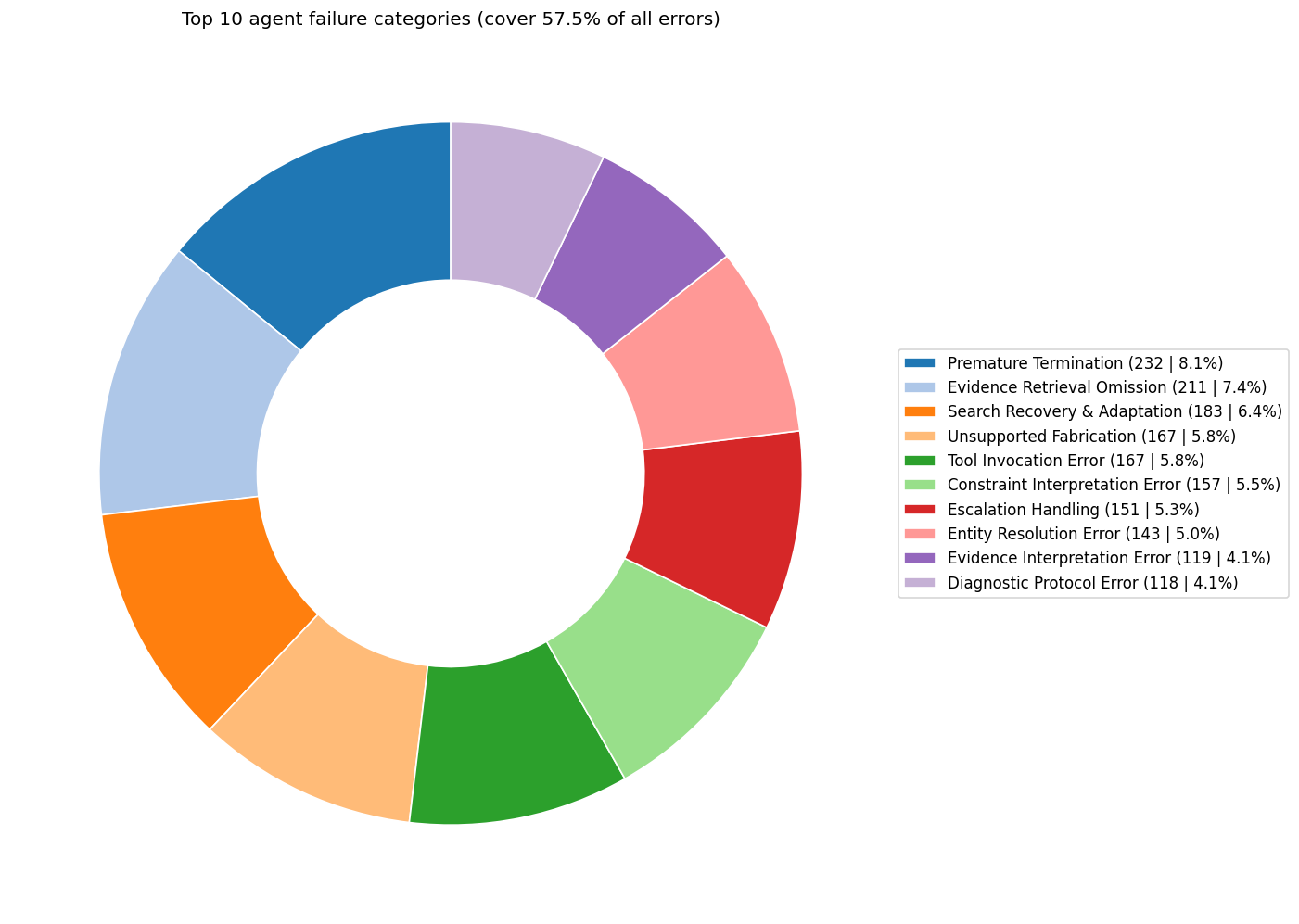}
\caption{Top-10 categories by share of categorized records ($n=2{,}868$).}
\label{fig:agentic-top10}
\end{figure}

The largest category, \emph{Premature Termination}, accounts for
8.1\% of categorized records ($n=2{,}868$); the top-10 cumulatively
cover 57.5\% (Table~\ref{tab:agentic-prevalence},
Figure~\ref{fig:agentic-top10}). Of the 27 top-level categories,
22 cross 1\% prevalence (Table~\ref{tab:agentic-prevalence});
the remaining five plus the \emph{Other} bucket cover the long
tail. Several categories are heavily concentrated in a single
benchmark (e.g., \emph{Diagnostic Protocol Error} is $97\%$
\Tbench{}-Telecom; \emph{Search Recovery \& Adaptation} is $92\%$
\BrowseComp{}), so the prevalence figures reflect benchmark mix
in our corpus.

\subsection{Per-Model Profiles}

\begin{figure}[h]
\centering
\includegraphics[width=\textwidth]{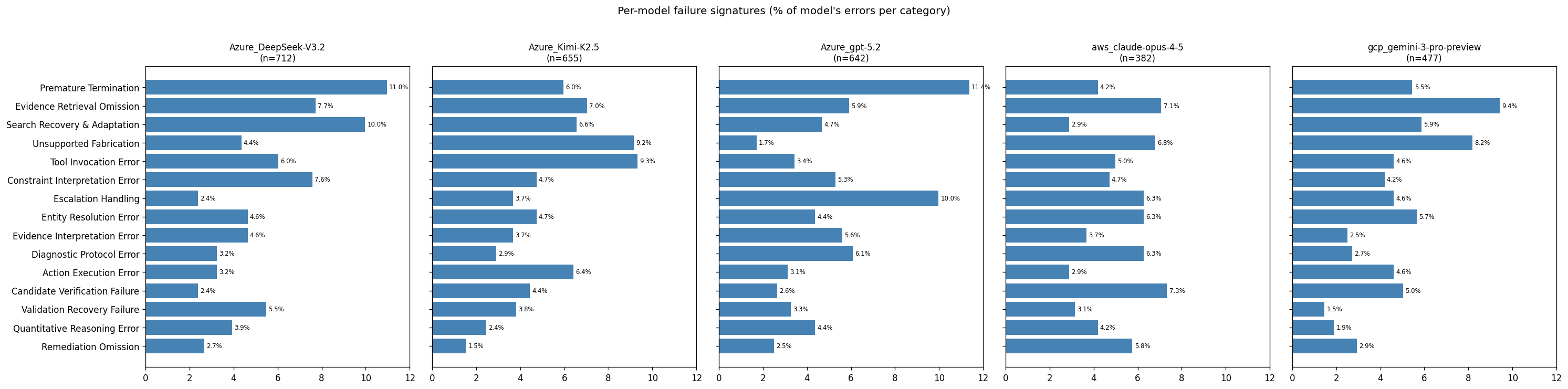}
\caption{Per-model failure-category shares (top-15 categories,
column-normalized).}
\label{fig:agentic-per-model}
\end{figure}

Failure-category shares vary across the five backbone models
(Figure~\ref{fig:agentic-per-model}). Reporting categories where a
model's share differs from the 5-model grand mean by at least 1.5
percentage points: \deepseek{} over-represents \emph{Search
Recovery \& Adaptation} ($+4.0$pp), \emph{Premature Termination}
($+3.4$pp), and \emph{Constraint Interpretation Error}
($+2.3$pp). \kimi{} over-represents \emph{Tool Invocation Error}
($+3.6$pp), \emph{Unsupported Fabrication} ($+3.1$pp), and
\emph{Action Execution Error} ($+2.4$pp). \gpt{} shows the largest
single deviation in this analysis on \emph{Escalation Handling}
($+4.6$pp), followed by \emph{Premature Termination}
($+3.8$pp); it also has the lowest \emph{Unsupported Fabrication}
share of any model ($-4.3$pp from mean). \opus{} over-represents
\emph{Candidate Verification Failure} ($+3.0$pp),
\emph{Remediation Omission} ($+2.7$pp), and \emph{Diagnostic
Protocol Error} ($+2.0$pp). \gemini{} Pro over-represents
\emph{Unsupported Fabrication} ($+2.1$pp), \emph{Evidence
Retrieval Omission} ($+2.0$pp), and \emph{Confirmation Handling
Error} ($+1.8$pp).

\subsection{Per-Architecture Profiles}

\begin{figure}[h]
\centering
\includegraphics[width=\textwidth]{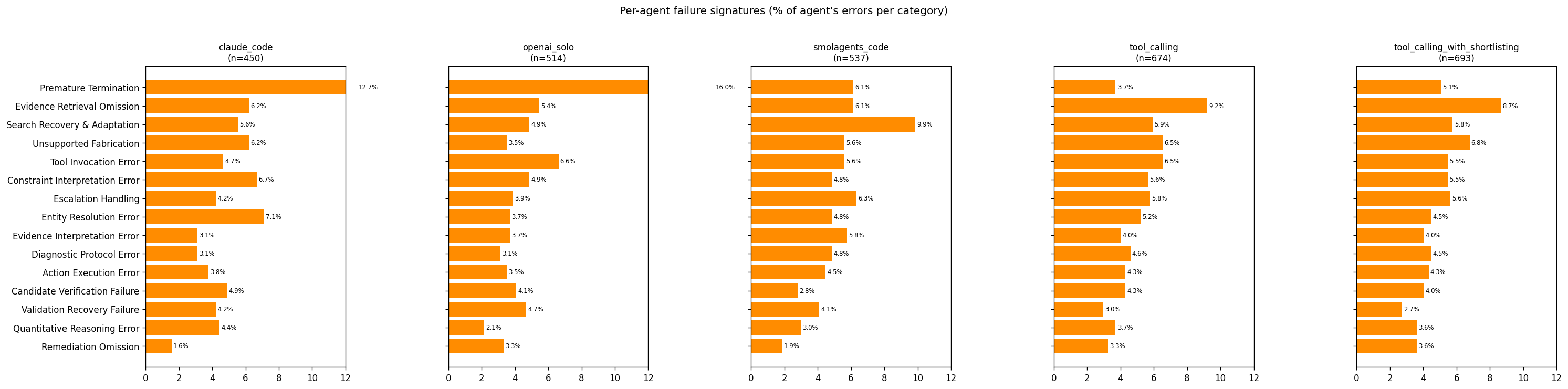}
\caption{Per-architecture failure-category shares (top-15 categories,
column-normalized).}
\label{fig:agentic-per-agent}
\end{figure}

Architecture-level shares vary along a different axis than the
model axis (Figure~\ref{fig:agentic-per-agent}). The autonomous
architectures \cc{} and \solo{} over-represent \emph{Premature
Termination} relative to the 5-architecture mean of 8.7\%
($+4.0$pp and $+7.3$pp respectively); \solo{} also
over-represents \emph{Finalization Protocol Error} ($+1.8$pp).
The tool-calling architectures \react{} and \short{} sit at the
opposite end on \emph{Premature Termination} ($-5.0$pp and
$-3.7$pp); both over-represent \emph{Evidence Retrieval Omission}
($+2.1$pp and $+1.5$pp). \smol{}, the code-execution architecture,
has a distinct profile with elevated \emph{Search Recovery \&
Adaptation} ($+3.5$pp), \emph{Authentication Handling Error}
($+1.9$pp), and \emph{Evidence Interpretation Error} ($+1.7$pp).
On this dependent variable, the architecture axis produces
non-trivial within-axis spread, in contrast to its small
contribution to success-rate variance ($\eta^2=0.5\%$,
\S\ref{sec:variance}).

\subsection{Per-Benchmark Profiles}

\begin{figure}[h]
\centering
\includegraphics[width=\textwidth]{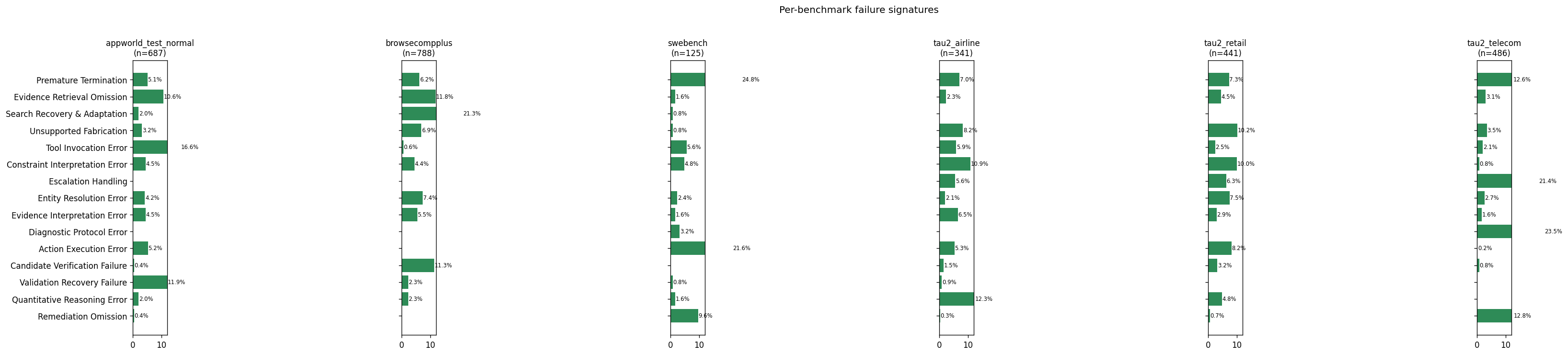}
\caption{Per-benchmark failure-category shares (top-15 categories,
column-normalized).}
\label{fig:agentic-per-benchmark}
\end{figure}

Benchmark-level shares show the largest within-axis spread of the
three axes (Figure~\ref{fig:agentic-per-benchmark}).
\BrowseComp{} over-represents \emph{Search Recovery \& Adaptation}
($+17.3$pp, the largest single deviation in this analysis),
\emph{Candidate Verification Failure} ($+8.4$pp), and
\emph{Evidence Retrieval Omission} ($+6.1$pp).
\SWEBench{} over-represents \emph{Action Execution Error}
($+14.9$pp) and \emph{Premature Termination} ($+14.3$pp).
\Tbench{}-Telecom over-represents \emph{Diagnostic Protocol
Error} ($+19.0$pp), \emph{Escalation Handling} ($+15.9$pp), and
\emph{Remediation Omission} ($+8.8$pp). \Tbench{}-Airline
over-represents \emph{Policy Eligibility Error} ($+11.9$pp),
\emph{Quantitative Reasoning Error} ($+8.5$pp), and
\emph{Constraint Interpretation Error} ($+5.0$pp).
\Tbench{}-Retail over-represents \emph{Confirmation Handling
Error} ($+6.6$pp), \emph{Unsupported Fabrication} ($+4.7$pp), and
\emph{Constraint Interpretation Error} ($+4.1$pp).
\AppWorld{} over-represents \emph{Tool Invocation Error}
($+11.1$pp), \emph{Validation Recovery Failure} ($+9.3$pp), and
\emph{Coverage Enumeration Omission} ($+6.2$pp), and shows the
broadest spread (seven categories at $\geq{+}2$pp). Per-benchmark
sample sizes are unequal: $n=125$ for \SWEBench{}, $n=341$ for
\Tbench{}-Airline, $n=441$ for \Tbench{}-Retail, $n=486$ for
\Tbench{}-Telecom, $n=687$ for \AppWorld{}, $n=788$ for
\BrowseComp{}; \SWEBench{} carries the largest uncertainty.

\begin{table}[h]
\centering\small
\caption{Top-level agent failure categories with $\geq$1\%
prevalence ($n=2{,}868$ categorized records).}
\label{tab:agentic-prevalence}
\begin{tabular}{rlrr}
\toprule
\# & Category & n & \% \\
\midrule
1 & Premature Termination        & 232 & 8.1 \\
2 & Evidence Retrieval Omission  & 211 & 7.4 \\
3 & Search Recovery \& Adaptation & 183 & 6.4 \\
4 & Unsupported Fabrication      & 167 & 5.8 \\
5 & Tool Invocation Error        & 167 & 5.8 \\
6 & Constraint Interpretation Error & 157 & 5.5 \\
7 & Escalation Handling          & 151 & 5.3 \\
8 & Entity Resolution Error      & 143 & 5.0 \\
9 & Evidence Interpretation Error & 119 & 4.1 \\
10 & Diagnostic Protocol Error   & 118 & 4.1 \\
11 & Action Execution Error      & 118 & 4.1 \\
12 & Candidate Verification Failure & 115 & 4.0 \\
13 & Validation Recovery Failure & 104 & 3.6 \\
14 & Quantitative Reasoning Error & 97  & 3.4 \\
15 & Remediation Omission        & 81  & 2.8 \\
16 & Coverage Enumeration Omission & 75  & 2.6 \\
17 & Finalization Protocol Error & 73  & 2.5 \\
18 & Unauthorized Action Execution & 69  & 2.4 \\
19 & Policy Eligibility Error    & 67  & 2.3 \\
20 & Query Formulation Error     & 60  & 2.1 \\
21 & Confirmation Handling Error & 43  & 1.5 \\
22 & Authentication Handling Error & 36  & 1.3 \\
\midrule
\multicolumn{2}{r}{Other (uncategorized)}  & 61  & 2.1 \\
\bottomrule
\end{tabular}
\end{table}

\subsection{Validation}

Following \textsc{ErrorMap}'s protocol, we sampled 100 labelled
records and presented the judge with the assigned category vs.\ a
random alternative; the sample over-represents \emph{Other}
($27/100$) to stress-test the residual bucket. The judge prefers
the assigned label in $85/100$ cases overall. Broken down,
agreement is $66/73$ ($90\%$) for named categories and $19/27$
($70\%$) for \emph{Other}; misclassification concentrates in the
residual bucket.

\subsection{Category Definitions}
\label{app:category-definitions}

Table~\ref{tab:agentic-definitions} lists the natural-language definition of each top-level failure category, as produced by the LLM judge during the recursive-categorization stage and used as the classification reference.
\input{appendices/category_definitions}

%% file: appendices/category_definitions.tex
\begingroup
\small
\begin{longtable}{@{}p{3.5cm}p{9cm}@{}}
\caption{Definitions of the 27 top-level failure categories produced by the recursive categorization stage. Definitions are the LLM judge's own outputs (saved at category-construction time); the analysis records released alongside the paper include the raw responses.}
\label{tab:agentic-definitions} \\
\toprule
\textbf{Category} & \textbf{Definition} \\
\midrule
\endfirsthead
\multicolumn{2}{l}{\emph{(Table~\ref{tab:agentic-definitions} continued)}} \\
\toprule
\textbf{Category} & \textbf{Definition} \\
\midrule
\endhead
\midrule
\multicolumn{2}{r}{\emph{Continued on next page}} \\
\endfoot
\bottomrule
\endlastfoot
\emph{Action Execution Error} & Agent has authorization and correct intent but carries out required mutations, payments, bookings, patches, cleanup, batching, or edits incorrectly or incompletely. \\
\emph{Authentication Handling Error} & Agent omits, misreads, invents, corrupts, extracts incorrectly, or misuses credentials, tokens, login state, authentication order, or authenticated context. \\
\emph{Candidate Verification Failure} & Agent commits to or rejects a candidate without checking discriminating constraints, alternatives, contradictions, disqualifiers, or uncertainty against available evidence. \\
\emph{Confirmation Handling Error} & Agent omits, mistimes, misreads, repeats, or malforms required consent, clarification, preference check, disclosure, approval, or post-action confirmation communication. \\
\emph{Constraint Interpretation Error} & Agent misreads explicit user or task constraints, including scope, dates, filters, preferences, privacy requirements, answer form, variants, or replacement boundaries. \\
\emph{Coverage Enumeration Omission} & Agent fails to exhaust required pages, lists, histories, variants, batches, statuses, transactions, recommendations, options, contacts, or libraries for complete coverage. \\
\emph{Diagnostic Protocol Error} & Agent skips, misorders, detours from, or misinterprets required troubleshooting checks, reproductions, account tests, connectivity tests, configuration checks, or diagnostic verification. \\
\emph{Entity Resolution Error} & Agent identifies or acts on the wrong entity, such as a person, account, order, item, address, path, role, or counterparty. \\
\emph{Escalation Handling} & Agent fails to escalate when required, or performs required escalation with missing tools, handoff details, notifications, routing, or protocol steps. Combined with: Agent escalates or transfers too early, before available in-scope troubleshooting, diagnostics, modification, cancellation, policy decision, or self-service resolution is attempted. \\
\emph{Evidence Interpretation Error} & Agent misunderstands retrieved evidence, including source details, filters, tool outputs, rankings, attributions, or implications, after the evidence is available. \\
\emph{Evidence Retrieval Omission} & Agent fails to open, inspect, expand, download, extract, or verify a necessary source, file, message, prior state, or record. \\
\emph{Finalization Protocol Error} & Agent completes work but omits or malforms required final answer, submission, completion signal, exact-answer field, alias handling, or answer format. \\
\emph{Hard to Analyze} & Failure label is ambiguous, subjective, null, artifact-like, or lacks enough context to infer a specific failed skill confidently. \\
\emph{Instruction Following Error} & Agent ignores, misapplies, or contradicts explicit user, system, or task instructions. \\
\emph{Looping Behavior} & Agent repeats actions, reasoning, or tool calls without making progress toward completion. \\
\emph{Planning Error} & Agent lacks a viable step sequence, decomposes poorly, or chooses a strategy that prevents completion. \\
\emph{Policy Eligibility Error} & Agent misapplies policy rules governing eligibility, approval, refusal, compensation, cancellation, exchange, refund, return, insurance, payments, or in-scope service. \\
\emph{Premature Termination} & Agent stops or responds as finished while required investigation, recovery, processing, implementation, troubleshooting, cleanup, or resolution work remains incomplete. \\
\emph{Quantitative Reasoning Error} & Agent incorrectly extracts, counts, ranks, compares, allocates, or calculates numeric values such as prices, refunds, fares, balances, durations, quantities, or availability. \\
\emph{Query Formulation Error} & Agent constructs missing, malformed, overbroad, overconstrained, literal, or poorly targeted queries, causing relevant information to be excluded from retrieval. \\
\emph{Remediation Omission} & Agent diagnoses or identifies a required fix but fails to perform necessary remediation, such as reset, update, enablement, permission, configuration, or cleanup. \\
\emph{Search Recovery \& Adaptation} & Agent does not handle retrieval failures, unavailable sources, invalid queries, timeouts, or empty results through retries, repair, or alternate sources. Combined with: Agent persists with unproductive search strategy instead of pivoting terms, decomposing clues, changing scope, switching sources, or following promising leads. \\
\emph{State/Goal Tracking Error} & Agent loses the user's objective, switches tasks, or optimizes for the wrong success condition. / Agent loses track of completed actions, environment state, or unresolved subtasks during execution. / Agent forgets or misuses prior conversation details needed for later decisions. \\
\emph{Tool Invocation Error} & Agent chooses an inappropriate tool or violates tool schema, arguments, parameters, capability limits, batching rules, placeholders, call order, or one-tool-per-turn constraints. \\
\emph{Unauthorized Action Execution} & Agent performs an action that is unapproved, ineligible, irreversible, unnecessary, overbroad, duplicate, out-of-scope, or not requested by the user. \\
\emph{Unsupported Fabrication} & Agent states or relies on facts, source content, credentials, diagnostics, procedures, restrictions, or causal explanations not supported by evidence. \\
\emph{User Communication Error} & Agent gives unclear, incomplete, misleading, or poorly scoped communication to the user. Also includes: Agent proceeds despite missing essential user input that should have been requested first. \\
\emph{Validation Recovery Failure} & Agent receives tool validation feedback but repeats invalid calls, makes ineffective repairs, or fails to choose a valid call or alternative path. \\
\end{longtable}
\endgroup

%% file: sections/08-limitations.tex
While \method{} provides a clear methodology and reusable building blocks for adaptation, familiarity with these capabilities and additional development work are still required when integrating new agents or benchmarks.

The agent protocols we evaluate (tool-calling, \MCP{}, Python code generation) all accept multimodal inputs, but the benchmarks in this study (widely adopted by frontier-model developers) are selected to stress cross-domain capability rather than multimodal processing. Extending the evaluation to environments with continuous action spaces, such as pixel-level computer-use or robotic control, is a natural next step and will require additions to the \UP{}'s current typed-action model.

Agent evaluation is expensive, especially for general-purpose agents that must be tested across many benchmarks. Due to cost constraints, our selection of agents and models is limited and does not cover the full range of open-source models or existing general-purpose agents. Three specific scope caveats follow from this. First, the open-weight tier consists of two checkpoints (\deepseek{}, \kimi{}); claims about ``open-weight generality sinks'' should be read as observations about these two models, not about open-weight checkpoints in general. Second, each (architecture, model, benchmark) cell is run once with the provider's documented default sampling settings; the resulting per-cell scores reflect a single-sample point estimate of an inherently stochastic process, and per-cell variability across reruns is not measured here. Third, success is judged by each benchmark's native automated evaluator; we do not conduct human evaluation of partial-credit outputs, which is particularly relevant on \BrowseComp{} and \SWEBench{} where benchmark-native scoring may diverge from human judgment.
To enable further progress in the field, future work should therefore explore techniques such as intelligent sampling and early stopping to reduce evaluation costs when it is clear that certain agent--model combinations underperform.